\documentclass[10pt,twocolumn,letterpaper]{article}

\usepackage[pagenumbers]{wacv}               

\usepackage{graphicx}
\usepackage{amsmath}
\usepackage{amssymb}
\usepackage{booktabs}

\newcommand{\comment}[1]{\ignorespaces}

\usepackage{subfiles}
\usepackage{gensymb}
\usepackage{amsthm}
\usepackage{exscale}
\usepackage{textcomp}		       
\usepackage{caption}               
\usepackage{subcaption}            
\usepackage{threeparttable}        
\usepackage{booktabs}
\usepackage{bbding}                
\usepackage{amsmath}               
\usepackage{accents}               
\usepackage{multirow}              
\usepackage{mathtools}             
\usepackage{footnote}
\usepackage{tablefootnote}         
\usepackage{colortbl}              
\usepackage{amssymb}               
\usepackage{pifont}
\usepackage{titletoc}              

\usepackage{algorithm}
\usepackage{algorithmicx}

\usepackage{algcompatible}
\usepackage{algpseudocode}

\usepackage{adjustbox}

\usepackage{overpic}

\usepackage{lipsum}                
\usepackage{kotex}                 

\newcommand{\cmark}{\textcolor{ForestGreen}{\ding{51}}}%
\newcommand{\xmark}{\textcolor{BrickRed}{\ding{55}}}%
\usepackage{xspace}

\newcommand{\MODELNAME}[1]{{ScribbleDiff}}

\usepackage[dvipsnames]{xcolor}

\usepackage[pagebackref,breaklinks,colorlinks]{hyperref}

\usepackage[capitalize]{cleveref}
\crefname{section}{Sec.}{Secs.}
\Crefname{section}{Section}{Sections}
\Crefname{table}{Table}{Tables}
\crefname{table}{Tab.}{Tabs.}

\def\wacvPaperID{2370} 
\def\confName{WACV}
\def\confYear{2025}

\begin{document}

\title{Scribble-Guided Diffusion for Training-free Text-to-Image Generation}



\author{
    Seonho Lee\thanks{Equal contribution} \quad Jiho Choi\footnotemark[1] \quad Seohyun Lim \quad Jiwook Kim \quad Hyunjung Shim\thanks{Corresponding author} \\
    Graduate School of Artificial Intelligence, KAIST \\
    Seoul, Republic of Korea \\
    {\tt\small \{glanceyes, jihochoi, seohyunlim, tom919, kateshim\}@kaist.ac.kr}
}

\maketitle


\begin{abstract}
{


{\color{lightgray}


}

Recent advancements in text-to-image diffusion models have demonstrated remarkable success, yet they often struggle to fully capture the user's intent.
Existing approaches using textual inputs combined with bounding boxes or region masks fall short in providing precise spatial guidance, often leading to misaligned or unintended object orientation.
To address these limitations, we propose Scribble-Guided Diffusion (ScribbleDiff), a training-free approach that utilizes simple user-provided scribbles as visual prompts to guide image generation.
However, incorporating scribbles into diffusion models presents challenges due to their sparse and thin nature, making it difficult to ensure accurate orientation alignment.
To overcome these challenges, we introduce moment alignment and scribble propagation, which allow for more effective and flexible alignment between generated images and scribble inputs.
Experimental results on the PASCAL-Scribble dataset demonstrate significant improvements in spatial control and consistency, showcasing the effectiveness of scribble-based guidance in diffusion models.
Our code is available at \url{https://github.com/kaist-cvml-lab/scribble-diffusion}.
}
\end{abstract}

\section{Introduction}
\label{sec:introduction}


\comment{
* https://docs.google.com/document/d/1PcHVrWfteZsW0eM5AepIkwJ_s7_iBo_OkWFGTGbBjxs

1. T2I diffusion 모델과 diffusion model
2. 사용자 의도를 반영하기 위한 다양한 기존 conditional diffusion model
3. 기존 방법의 한계 (box, dense) 과 Scribble Guidance 의 필요성
    * Scribble
        * 많이 사용 PixelLLM, scribble segmentation
        * 사용성 (쉬움) box > scribble > mask (어려움)
        * 정보량 (적음) box < scribble < mask (많음)
4. 우리의 해법
    * ill-posed problem
    * User Intention 정의
        * location & shape (3.1 preliminary + focal loss)
        * orientation (3.2 image moments)
        * overlap (3.3 scribble propagation)
5. Contribution 요약
}


\comment{
    1.1. T2I diffusion 모델과 diffusion model
    
    다양한 Text-to-image diffusion models (DALL-E, Imagen, SD) 의 발전은 텍스트를 활용한 이미지 생성 분야에서 높은 성능의 결과를 보이고 있다.
    그러나 이러한 확산 모델들은 텍스트에 가반하기 때문에 사용자의 의도를 온전히 반영한 이미지를 생성하는 데 어려움이 있다.
    이러한 문제는 텍스트 프롬프트에서의 공간 정보 부족과 텍스트와 이미지 modality 간 모호성에서 비롯된다.
    그 결과 이미지 생성에서 사용자의 의도를 더 잘 반영하기 위한 box 와 mask 등의 visual prompt 를 활용하는 조건부 확산 모델의 필요성을 증진 시켰다.
}

\noindent
Text-to-image diffusion models~\cite{ramesh2021zero_dalle,saharia2022photorealistic_imagen,rombach2022high_latent} have achieved great success in text-based image generation, producing high-quality visuals that align closely with textual descriptions.
However, these models often struggle to fully capture the \textit{user's intent} due to their reliance on textual input, which inherently lacks spatial information.
This reliance introduces ambiguity in aligning the generated image with the user’s intent, as textual descriptions can be open to multiple interpretations~\cite{liu2024training_free_composite,gafni2022make_a_scene}, particularly regarding object location, shape, and orientation.

\begin{figure}[tb]
    \centering
    \begin{subfigure}[b]{0.325\linewidth}
        \centering
        \includegraphics[width=\linewidth, trim={0cm 0cm 0cm 0cm}, clip]{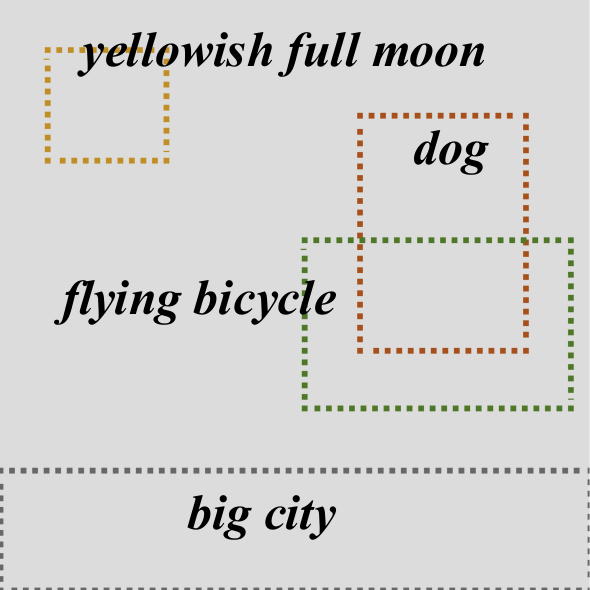}
        \includegraphics[width=\linewidth, trim={0cm 0cm 0cm 0cm}, clip]{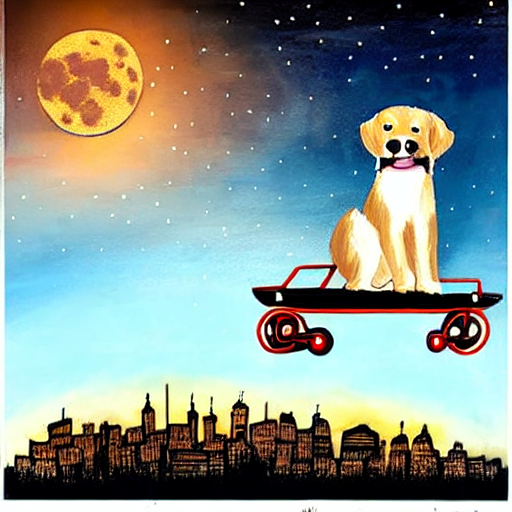}
        \captionsetup{justification=centering}
        \caption{
            Boxes\\
            (BoxDiff~\cite{xie2023boxdiff})
        }
    \end{subfigure}
    \begin{subfigure}[b]{0.325\linewidth}
        \centering
        \includegraphics[width=\linewidth, trim={0cm 0cm 0cm 0cm}, clip]{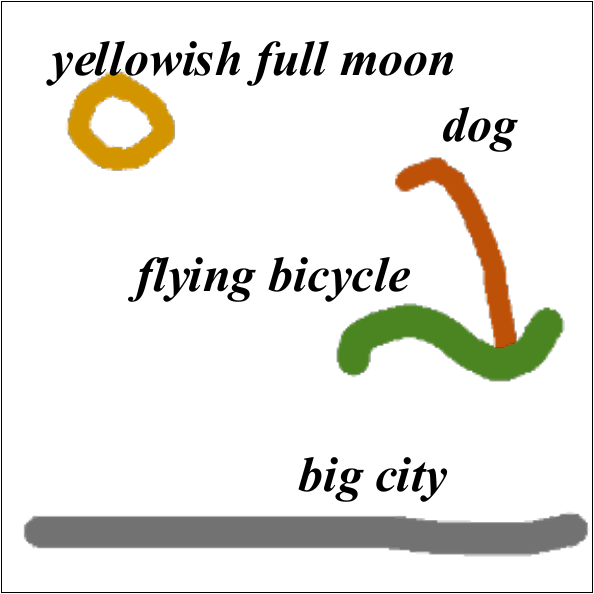}
        \includegraphics[width=\linewidth, trim={0cm 0cm 0cm 0cm}, clip]{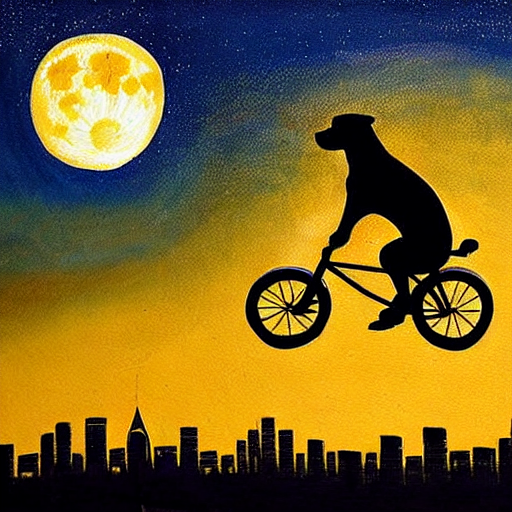}
        \captionsetup{justification=centering}
        \caption{
            Scribbles\\
            (Ours)
        }
    \end{subfigure}
    \begin{subfigure}[b]{0.325\linewidth}
        \centering
        \includegraphics[width=\linewidth, trim={0cm 0cm 0cm 0cm}, clip]{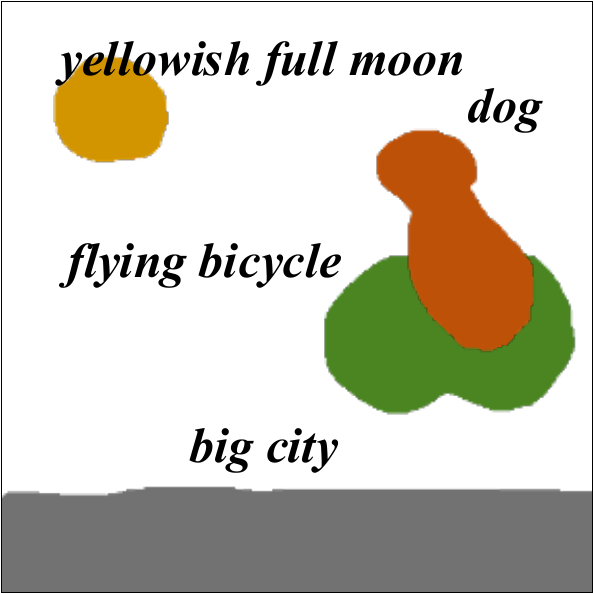}
        \includegraphics[width=\linewidth]{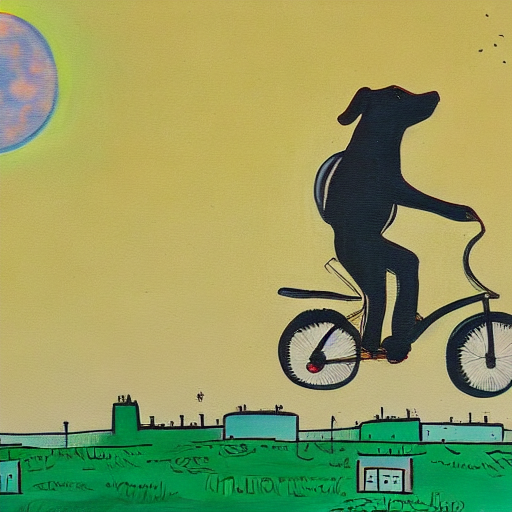}
        \captionsetup{justification=centering}
        \caption{
            Region Masks\\
            (DenseDiffusion~\cite{densediffusion})
        }
    \end{subfigure}
    \caption{
        Comparison of User Visual Prompts: Box, Scribble, and Mask in terms of usability, information amount, and directionality. \\
        {
            \footnotesize
            $\bullet\quad$ \textbf{Usability} (Easy to Difficult): Box {\scriptsize $>$} Scribble {\scriptsize $>$} Mask \\
            $\bullet\quad$ \textbf{Directionality} (Low to High): Box {\scriptsize $<$} Mask {\scriptsize $<$} Scribble \\
            (Text Prompt: \textit{A painting of a \textbf{dog} riding a \textbf{flying bicycle}, over a \textbf{big city} with a \textbf{yellowish full moon} in the night sky.})
        }
    }
    \label{fig:motivation}
    \vspace{-1em}
\end{figure}

To address these challenges, there has been a growing need for conditional diffusion models~\cite{xie2023boxdiff,densediffusion,wang2023compositional, liu2024training_free_composite, avrahami2023spatext, meng2021sdedit} that incorporate visual prompts offering greater control over the generation process.
Techniques like IP-Adapter and ControlNet~\cite{zhang2023adding_controlnet, ye2023ip_adapter} extend the approaches by accommodating diverse grounding inputs, including key points, depth maps, and normal maps.
Although these methods facilitate conditional generation into pre-trained large-scale diffusion models, they still require fine-tuning.
In contrast, some training-free approaches~\cite{bansal2023universal, ma2023directed, park2022shape_guidied} guide the diffusion model's reverse process with additional inputs like bounding boxes and region masks.
These methods define new loss functions to optimize the noisy latent code during the denoising process, eliminating the need for fine-tuning.


\comment{
    1.3. 기존 방법의 한계 (box, dense) 과 Scribble Guidance 의 필요성
        * Scribble
            * 많이 사용 PixelLLM, scribble segmentation
            * 사용성 (쉬움) box > scribble > mask (어려움)
            * 정보량 (적음) box < scribble < mask (많음)
}


While the conditioning inputs discussed above~\cite{densediffusion,xie2023boxdiff,bansal2023universal,ma2023directed,park2022shape_guidied} are essential for guiding generation, they have notable limitations.
Bounding boxes often fail to accurately convey spatial attributes such as the abstract shape or orientation of objects inside the boxes, leading to generated images where objects may face unintended directions, as shown in~\cref{fig:motivation} (a).
Region masks, although more precise, involve higher annotation costs and may not effectively convey the orientation of the object as~\cref{fig:motivation} (c).
As a compromise between boxes and region masks, we employ \textit{scribbles}~\footnote{
    We refer to \textit{scribble} as Bezier Scribble, following the terminology in ScribbleSeg~\cite{chen2023scribbleseg}. While the term 'scribble diffusion' exists, it aligns more closely with sketch-guided diffusion~\cite{voynov2023sketch_diff,ding2024training_free_sketch_diff}, which is particularly sensitive to user-defined boundaries and edges.
}, a visual prompt closely related to its use in weakly supervised semantic learning~\cite{boykov2001interactive_scribble,lin2016scribblesup_data,chen2023scribbleseg,chen2022scribble2d5,dorent2020scribble_weakly_seg,valvano2021learning_scribble_seg,wong2023scribbleprompt} and interactive segmentation~\cite{wang2005interactive_scribble_video,cheng2021modular_scribble_video}, as visual prompts to capture the user's intent with strokes, as illustrated in~\cref{fig:motivation} (b).

While scribbles are simple annotations, they effectively convey spatial information, such as object location and abstract shapes, similar to region masks, but with lower annotation costs~\cite{lin2016scribblesup_data,wong2023scribbleprompt}. Additionally, scribbles are particularly well-suited for expressing directionality, offering spatial cues that are often lacking in traditional inputs like bounding boxes and region masks.
Given the success of diffusion models in conditional image generation, a compelling question arises: Can a single scribble (or stroke) serve as an effective spatial guiding input for diffusion models? 
Although BoxDiff~\cite{xie2023boxdiff} provide examples of using scribbles, it propose a method that do not account for its distinctive properties.
As a result, features like the thinness and directional nature of scribbles were not adequately reflected and remained understudied.

In this study, we propose a novel training-free method for text-to-image generation using \textit{\textbf{scribble}} prompts to overcome the limitations of traditional spatial inputs, such as bounding boxes and region masks, which often fail to capture object orientation and abstract shape. To address this, we introduce a moment loss that refines the cross-attention activation distribution, aligning the generated object's orientation with the scribble’s direction. Additionally, to handle the sparse and thin nature of scribbles, which can make precise control challenging, we propose scribble propagation.  This method allows for fine-grained control of object orientation and spatial arrangement using scribbles, effectively balancing simplicity and precision in guiding diffusion models. Our experimental results demonstrate that this approach not only improves positional and shape accuracy but also significantly enhances orientation alignment with the scribble prompts across various baselines.

\section{Background}
\label{sec:background}


\noindent
\textbf{Diffusion Models.}
Diffusion models~\cite{sohl2015deep,ho2020denoising_ddpm,song2020denoising_ddim} have gained significant attention for their ability to generate high-quality images. 
The diffusion U-Net $\epsilon_{\theta}$, parameterized by $\theta$, predicts the noise $\epsilon$ with respect to each timestep $t \in \left\{ 1, \dots, T -1, T \right\}$ to denoise the noisy sample in reverse process. DDPM~\cite{ho2020denoising_ddpm} samples new images from a noise distribution $\mathcal{N}(0, I)$, using $\epsilon_{\theta}$ and its sampling algorithm. The forward process sampling distribution $q(\mathbf{x_t} | \mathbf{x_{t-1}})$ is described as a first-order Markov process, where $\mathbf{x_t}$ is a noisy sample in image space perturbed by timestep $t$, characterized by the variance scheduling hyperparameter $\beta_t$.
An intermediate noisy sample $\mathbf{x_t}$ derived from the input image $\mathbf{x_0}$ can be computed using the following distribution $q(\mathbf{x_t} | \mathbf{x_0}) = \mathcal{N}(\sqrt{\alpha_t} \mathbf{x_0}, (1 - \alpha_t)I)$, where $\alpha_t = \prod_{s=1}^t (1 - \beta_s)$. 

Building upon this, DDIM~\cite{song2020denoising_ddim} introduced a reparameterization of the forward process as a non-Markovian approach. Specifically, the backward process can be formulated as follows:
\begin{equation}
\begin{aligned}
    \mathbf{x_{t-1}} &= \sqrt{\alpha_{t-1}} \underbrace{\left( \frac{\mathbf{x_t} - \sqrt{1 - \alpha_t}\epsilon_{\theta}(\mathbf{x_t}, t)}{\sqrt{\alpha_t}} \right)}_{\text{predicted } \mathbf{x_0}} \\
    &+ \underbrace{\sqrt{1 - \alpha_{t-1} - \sigma_t^2}\cdot \epsilon_\theta(\mathbf{x_t}, t)}_{\text{direction pointing to } \mathbf{x_t}} + \underbrace{\sigma_t \mathbf{z_t}}_{\text{random noise}},
    \label{eq:ddim_reverse}
\end{aligned}
\end{equation}
where $\sigma_t = \eta \sqrt{\frac{(1 - \alpha_{t-1})}{\alpha_t}}$. When $\sigma_t = 0$, then the backward process becomes deterministic.

\begin{figure*}[ht]
    \centering
    \includegraphics[width=0.95\linewidth]{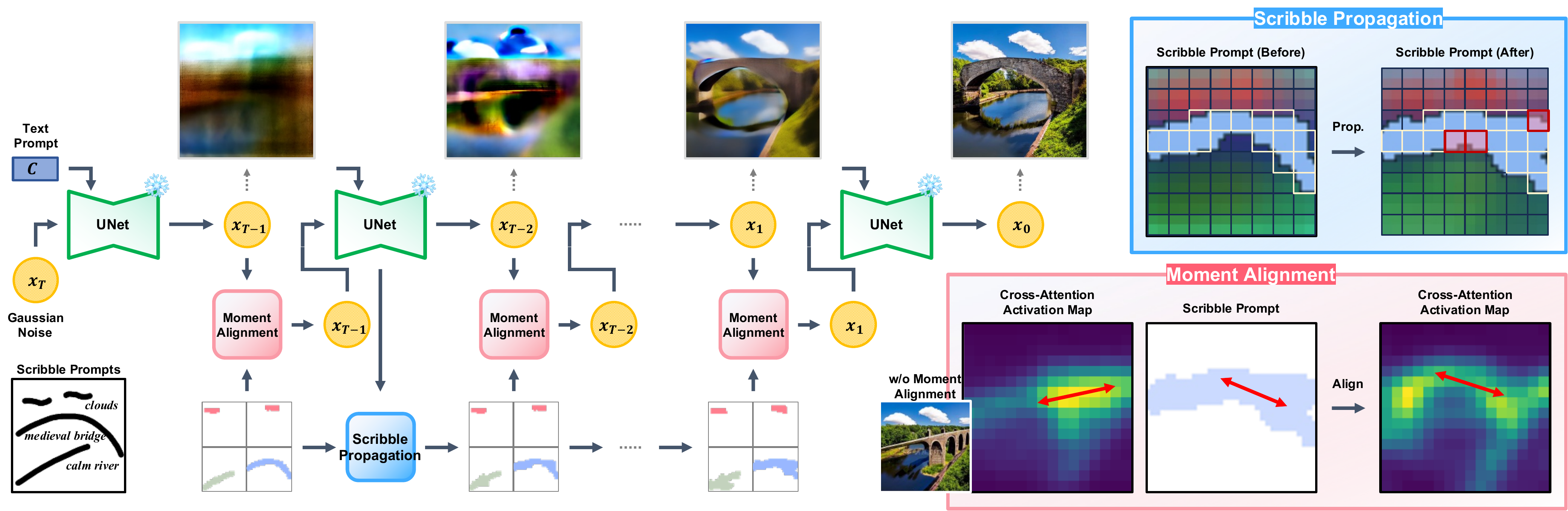}
    \caption{
        \textbf{The overall architecture.}
        Training-free Scribble-Guided Diffusion (\MODELNAME \.) consists of two main components: Moment alignment and scribble propagation.
        The red arrows represent the main orientations of the distributions.
        and the anchors with high similarity (red rectangles) are gathered based on the scribble's anchors (yellow rectangles).
        (Text Prompt: \textit{The \textbf{clouds} drift high in the sky, casting soft, shifting shadows on the \textbf{calm river} below. A \textbf{medieval bridge} spans the width of the waterway.})
    }
    \label{fig:overview}
\end{figure*}


\noindent
\textbf{Guidance with Energy Function}. According to the score-based perspectives from previous studies~\cite{sohl2015deep, song2019generative, song2020score}, diffusion models can be viewed as a denoising network $\epsilon_{\theta}$ that estimate a score function $\nabla_{\mathbf{x}_t} \log p_t(\mathbf{x}_t) \propto -\epsilon_{\theta}(\mathbf{x}_t, t)$. For conditional image generation with additional inputs $y$, the conditional score function can be decomposed with the Bayes' rule as follows:
\begin{equation}
\begin{aligned}
    \nabla_{\mathbf{x}_t} \log p_t(\mathbf{x}_t | y) = \nabla_{\mathbf{x}_t} \log p_t(\mathbf{x}_t) + \nabla_{\mathbf{x}_t} \log p_t(y | \mathbf{x}_t),
    \label{eq:conditional_score_function}
\end{aligned}
\end{equation}
where $\nabla_{\mathbf{x}_t} \log p_t(\mathbf{x}_t)$ is the unconditional score from the diffusion models, and $\nabla_{\mathbf{x}_t} \log p_t(y | \mathbf{x}_t)$ is the conditional gradient, which adjusts the results of denoising process to align more closely with some functions or auxiliary models such as classifier guidance~\cite{dhariwal2021diffusion} dependent on the noisy sample $\mathbf{x}_t$. 
From the perspective of energy-based generative models~\cite{zhao2016energy,yu2023freedom}, this conditional gradient can be interpreted as deriving from an energy function $\mathcal{E}(\mathbf{x_t}, t, y)$, which encodes the discrepancy between the current state of $\mathbf{x}_t$ and the conditioning input $y$. 
Consequently, the estimated noise $ \hat{\epsilon}_{\theta}$ with classifier-free guidance~\cite{ho2022classifier} using the energy function $\mathcal{E}$ can be reformulated as:
\begin{equation}
\begin{aligned}
    \hat{\epsilon}_{\theta}(\mathbf{x}_t, t, y) &= (1 + \omega){\epsilon}_{\theta}(\mathbf{x}_t, t, y) \\
    &- \omega {\epsilon}_{\theta}(\mathbf{x}_t, t) 
    + \eta \nabla_{\mathbf{x}_t}\mathcal{E}(\mathbf{x}_t, t, y),
    \label{eq:energy_function}
\end{aligned}
\end{equation}
where $\omega$ is a classifier-free guidance scale and $\eta$ is a coefficient. The energy function $\mathcal{E}$ can be flexibly defined based on the user's intent, allowing the generated output to more closely align with the conditioning input $y$. 

Consequently, the noisy latent code $\mathbf{x}_t$ can be optimized using $\hat{\epsilon}_{\theta}$ at each denoising step during inference as follows:



\begin{equation}
\begin{aligned}
\mathbf{x}_{t-1}' = \mathbf{x}_{t} - \hat{\epsilon}_{\theta}(\mathbf{x}_{t}, t, y), 
\end{aligned}
\end{equation}
where $\mathbf{x}_{t-1}'$ represents the optimized latent code at $t-1$.




\noindent
\textbf{Controllable Diffusion Models.}
There have been several approaches aimed at providing users with fine-grained spatial control over the generation process in diffusion models. Some methods introduce diverse spatial conditions by incorporating additional trainable modules, such as zero convolution layers~\cite{zhang2023adding_controlnet} or adapters~\cite{ye2023ip_adapter}. However, these models often incur higher computational costs due to the need for fine-tuning with each type of conditioning input. Furthermore, they do not fully capture the nuances of certain forms of guidance, particularly scribbles, which are inherently ambiguous and sparse. As a result, scribbles are frequently overlooked or underutilized as effective visual prompts.
Although FreeControl~\cite{mo2024freecontrol} proposes a training-free method to controllable diffusion that accommodates various spatial conditions, it similarly fails to fully account for the characteristics of scribbles.

\noindent
\textbf{Attention Control in Diffusion Models.} 
Recent studies~\cite{tang2023emergent, kwon2022diffusion} have shown that intermediate results from the U-Net architecture in diffusion models provide valuable information for image synthesis.
In particular, cross-attention maps show the correspondence between input prompts and the reconstructed content~\cite{hertz2022prompt_cross_attention_control}.
Building on these observations, several methods~\cite{epstein2023diffusion_self_guidance, chefer2023attend, bao2024separate} have been proposed to manipulate attention maps to improve the quality and controllability of diffusion models.

Some approaches~\cite{phung2023grounded_Refocusing, wang2023compositional} use visual prompts, such as bounding box layouts, to better control spatial information and object placement by manipulating cross-attention maps. 
However, few works have explored using scribbles as a guiding input for conveying structural information. For instance, BoxDiff~\cite{xie2023boxdiff} introduces a training-free method with scribble constraints, but it primarily focuses on box-based spatial conditions and lacks a comprehensive understanding of scribbles as an input.
Similarly, DenseDiffusion~\cite{densediffusion} uses attention modulation to synthesize images using region masks, but it relies on masks rather than scribbles for spatial guidance and struggles with fine-grained, thin structures. While sketched-based conditional T2I generation models~\cite{voynov2023sketch_diff, ding2024training_free_sketch_diff} address the text-to-image generation with sketches, they differ from our approach, as sketches are more sensitive to edges or boundaries compared to scribbles. 

Inspired by these visual prompts and attention control techniques, we propose a method that allows the scribble, commonly used in weakly supervised learning, to better guide the generation process through newly defined energy functions. Our method effectively captures both the directional features and the abstract shape encoded in the scribble prompt.

\section{Method}
\label{sec:method}


\noindent
We propose a novel, training-free Text-to-Image (T2I) diffusion method, named Scribble-Guided Diffusion (\MODELNAME\.), which efficiently incorporates user-provided scribble prompts.
To enhance alignment with the input scribbles, we utilize attention control (\cref{Attention_Control_with_Scribble}), moment alignment (\cref{sec:Guidance_for_Moment_Alignment}), and scribble propagation techniques (\cref{sec:Scribble_Propagation}).
The overall architecture of \MODELNAME\. is shown in~\cref{fig:overview}.


We define the effective incorporation of scribbles as two main objectives: (1) alignment between the direction of the scribble and the generated object, and (2) transforming the sparse scribble into a dense annotation, ensuring that that the generated object fully encompasses the scribble.
To achieve these goals, the \MODELNAME \. consists of two key components: cross-attention control with moment alignment and scribble propagation.
In this section, we will explore these components in detail.

\subsection{Attention Control with Scribble}
\label{Attention_Control_with_Scribble}

\noindent
The proposed approach begins with cross-attention control~\cite{chefer2023attend,epstein2023diffusion_self_guidance,xie2023boxdiff,densediffusion,agarwal2023star}, which is commonly adopted in diffusion models.
Given a set of scribbles $\mathcal{S}$, where each scribble $s \in \mathcal{S}$ is associated with one or more text tokens $\mathcal{C}(s) = \{c_1, c_2, \dots, c_n\}$, the cross-attention activation maps $\mathcal{A}^{c}_{\texttt{cross}}$ represent the relationship between visual patches and each text token $c \in \mathcal{C}$.

To align the cross-attention activation map $\mathcal{A}^{c}_{\texttt{cross}}$ with the binary mask of corresponding scribble region $\mathcal{M}_{s}$, we define a focal loss for the cross-attention as follows:
\begin{equation}
\begin{aligned}
    \mathcal{L}_{\texttt{focal}} = \frac{1}{|\mathcal{S}|} \frac{1}{|\mathcal{C}(s)|} & \sum_{s \in 
    \mathcal{S}} \sum_{c \in \mathcal{C}(s)} (1 -     
    \sigma(\mathcal{A}^{c}_{\texttt{cross}}))^{\beta} \\
    \cdot & \left(\alpha \mathcal{M}_{s} \right. 
    \left. + (1 - \alpha) (1 - \mathcal{M}_{s} ) \right) \cdot \mathcal{L}_{\texttt{BCE}}
    \label{eq:cross_attention_focal_loss}
\end{aligned}
\end{equation}
where $\mathcal{L}_{\texttt{BCE}}$ is a binary cross entropy loss between $\mathcal{M}_{s}$ and $\sigma(\mathcal{A}^{c}_{\texttt{cross}})$, $\sigma$ is a sigmoid function, and $\alpha$ and $\beta$ are hyperparameters. This loss helps minimize cross-attention activations outside the scribble region and maximize them inside the scribble region, aligning the cross-attention activation with the valid regions defined by the abstract shape of the scribbles.
We set $\alpha=0.25$ since a lower $\alpha$ reduces the penalty on false predictions related to scribbles, considering that most scribbles are thin and should not be neglected.

\subsection{Guidance for Moment Alignment}
\label{sec:Guidance_for_Moment_Alignment}



\begin{figure}[t!]
    \includegraphics[width=\linewidth]{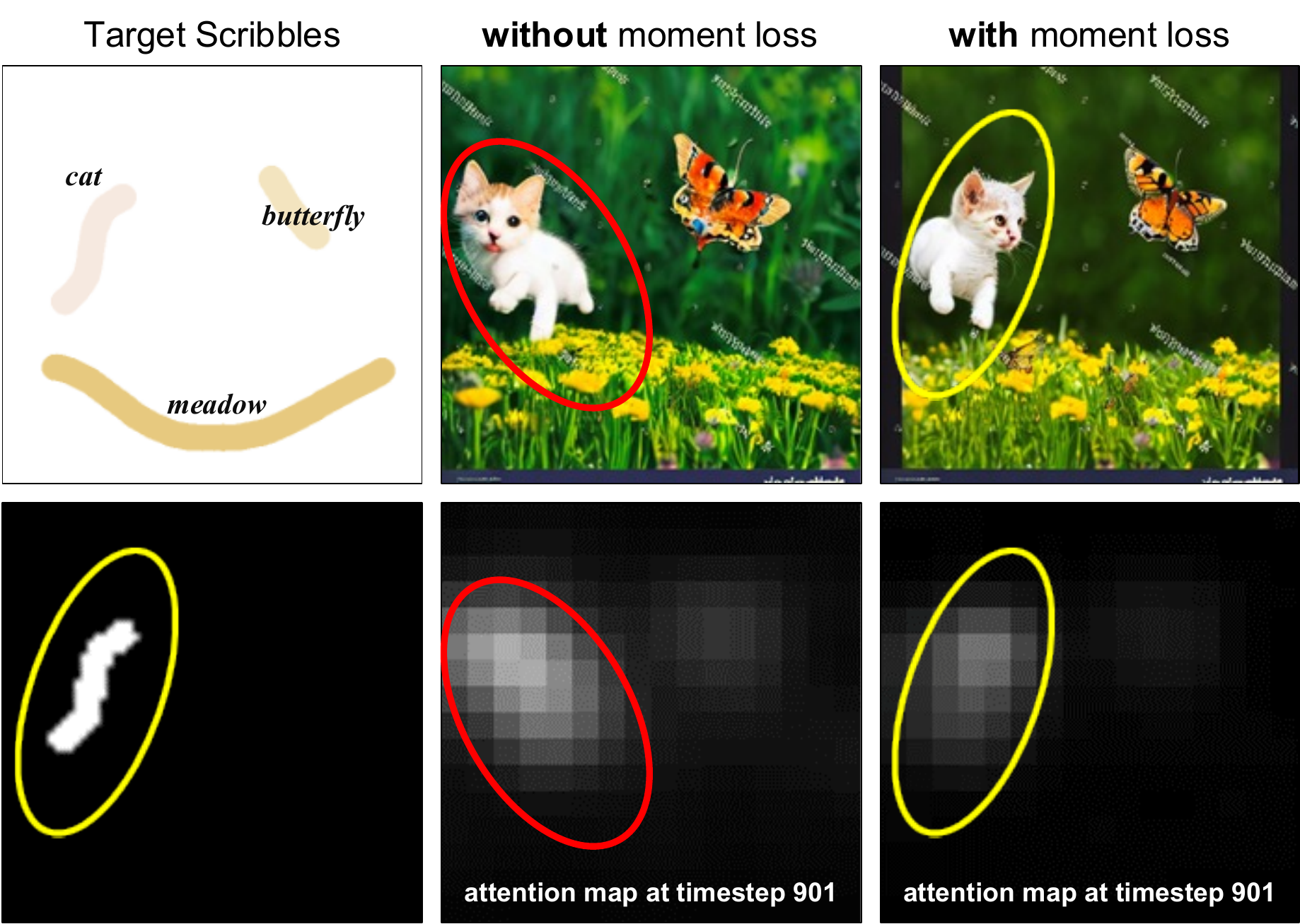}
    \caption{
        \textbf{Impact of moment loss on object orientation.}
        Moment loss improves alignment between the object’s orientation and the direction of the scribble.
        Without moment loss, the cat faces opposite to the scribble’s direction.
    }
    \vspace{-1.0em}
    \label{fig:moment_loss}
\end{figure}


\noindent
To achieve a higher degree of correspondence between the user-provided scribbles $\mathcal{S}$, and the cross-attention activation map $\mathcal{A}_{\texttt{cross}}$, we utilize the concept of image moments~\cite{mukundan1998moment,flusser2006moment}.
Image moments are statistical measures that capture the spatial distribution of an image or region within the image.

We propose that the spatial distribution of the cross-attention activations can be interpreted as an image moment, where each patch in the attention map corresponds semantically to a token with varying degrees of strength, ranging between 0 and 1. The first-order moment (or \textit{centroid moment}), represented as $\left(\frac{m_{10}}{m_{00}}, \frac{m_{01}}{m_{00}}\right)$, indicates the centroid or center of mass of a given region. The general moment is defined as:
\begin{equation}
m_{pq} = \sum_{x,y} x^p y^q I(x,y),
\end{equation}
where $I(x,y)$ denotes the image intensity at the point $(x,y)$. Diffusion Self-Guidance~\cite{epstein2023diffusion_self_guidance} introduces a method to align an object's position by adjusting the centroid of the cross-attention map to the target position. Similarly, our method leverages centroid loss to better align the generated content with the position specified by the scribble prompt.
The discrepancy between the centroids $(\bar{x}^{c}, \bar{y}^{c})$ and $(\bar{x}^{s}, \bar{y}^{s})$ of the cross-attention map and the scribble, respectively, defined by the first-order moments, can be minimized as:
\begin{equation}
\begin{aligned}
    {\mathcal{L}}^{s}_{\texttt{centroid}}
    &= \frac{1}{|\mathcal{C}(s)|} \sum_{c\in \mathcal{C}(s)}
    \left\{
        \left( \bar{x}^{c} - \bar{x}^{s} \right)^2
        +
        \left( \bar{y}^{c} - \bar{y}^{s} \right)^2
    \right\}, \\
    & \quad \text{where} \quad \bar{x} = \frac{{m}_{1,0}}{{m}_{0,0}}, \bar{y} = \frac{{m}_{0,1}}{{m}_{0,0}}, \\
    \mathcal{L}_{\texttt{centroid}} &= \frac{1}{|S|} \sum_{s \in S} \mathcal{L}^{s}_{\text{centroid}}.
    \label{eq:first_order_image_moment}
\end{aligned}
\end{equation}

Furthermore, we introduce a generalization of the centroid loss function by incorporating second-order moments to align the orientation of the generated object $\theta^{c}$ with the direction of scribble $\theta^{s}$. The second-order moments (or \textit{central moment}), such as $m_{20}$, $m_{02}$, and $m_{11}$, describe the objects' orientation and dispersion in the image, capturing its spread and shape.
The difference in the second-order moment between the scribble and the cross-attention activation map can be reduced as:
\begin{equation}
\begin{aligned}
    \mathcal{L}^{s}_{\texttt{central}} &= \frac{1}{\mathcal{C}(s)} \sum_{c \in \mathcal{C}(s)} |\theta^{c} - \theta^{s}| \\
    & \quad \text{where} \quad \theta = \frac{1}{2} \cdot \tan^{-1} \left( \frac{2\mu'_{1,1}}{\mu'_{2,0} - \mu'_{0,2}} \right), \\
    \mathcal{L}_{\texttt{central}} &= \frac{1}{2\pi * |S|} \sum_{s \in S} \mathcal{L}^{s}_{\texttt{central}},
    \label{eq:second_order_image_moment}
\end{aligned}
\end{equation}
where $\mu'_{1,1}=\frac{m_{1,1}}{m_{0,0}}-\bar{x}\bar{y}$, $\mu'_{0,2}=\frac{m_{0,2}}{m_{0,0}}-\bar{y}^2$, and $\mu'_{2,0}=\frac{m_{2,0}}{m_{0,0}}-\bar{x}^2$.
Finally, the method aligns the scribble itself along with the \textit{first} and \textit{second moments} of each scribble component with the moment loss $\mathcal{L}_{\texttt{moment}} = \lambda_{\texttt{1}} \mathcal{L}_{\texttt{centroid}} + \lambda_{\texttt{2}} \mathcal{L}_{\texttt{central}}$.
The corresponding cross-attention loss $\mathcal{L}_{\texttt{cross}}$ is a combination of focal and moment loss as follows:
\begin{equation}
    \mathcal{L}_{\texttt{cross}} = \mathcal{L}_{\texttt{focal}} + \mathcal{L}_{\texttt{moment}},
    \label{eq:cross_attention_loss}
\end{equation}
where $\lambda_{1}$ and $\lambda_{2}$ are hyperparameters that weight the centroid and central moment losses, respectively.
This approach not only enhances direct alignment but also better captures the orientation and positional information of the scribbles.


\subsection{Scribble Propagation}
\label{sec:Scribble_Propagation}

\begin{figure}[t!]
    \includegraphics[width=\linewidth]{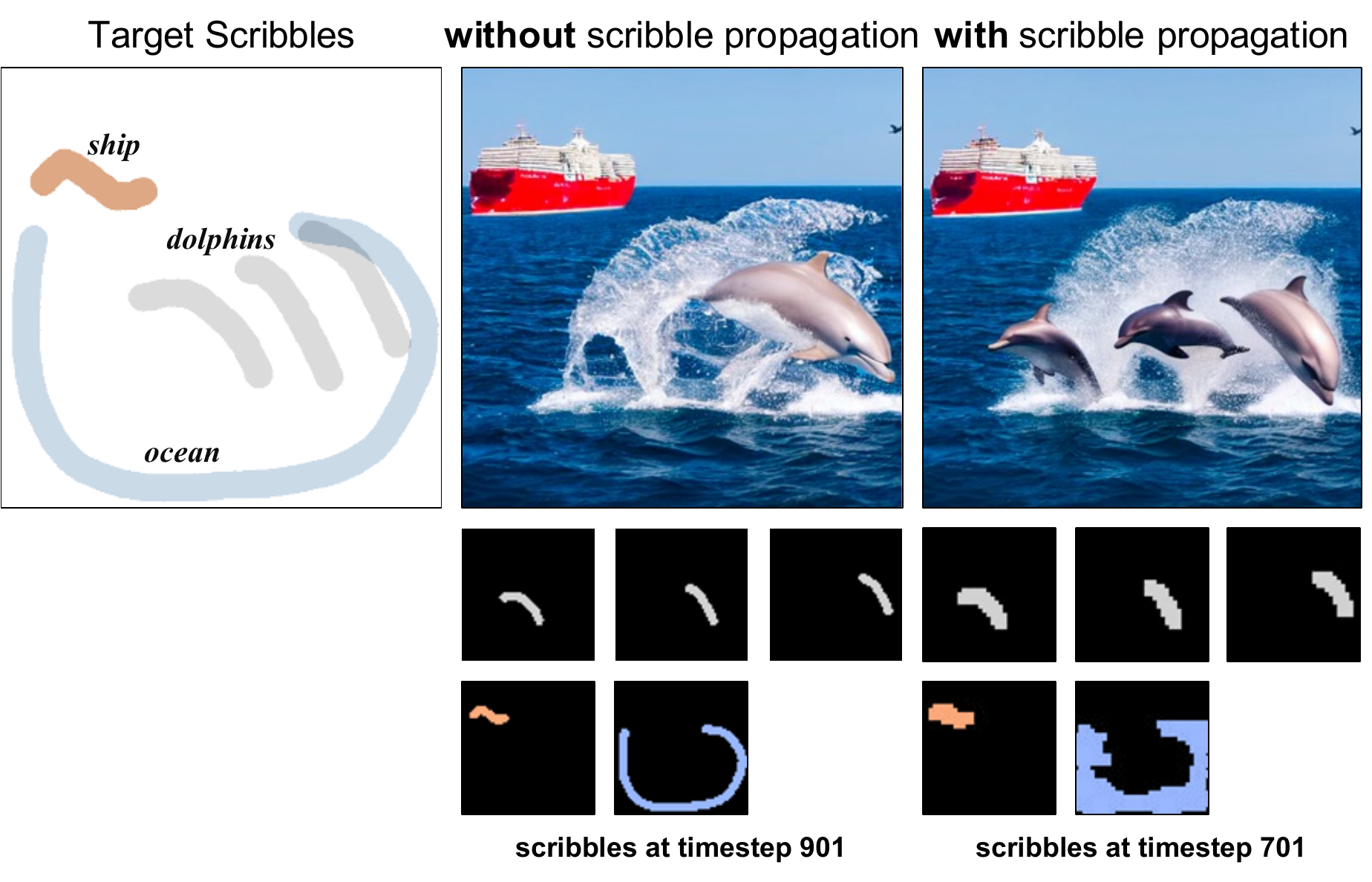}
    \caption{
        \textbf{Effect of scribble propagation.} 
        With scribble propagation in Stable Diffusion, the scribble expands significantly by timestep, improving object shape and enhancing visual coherence. 
    }
    \vspace{-1em}
    \label{fig:scribble_propagation}
\end{figure}

\noindent
While reducing $\alpha$ in \cref{eq:cross_attention_focal_loss} in \ref{Attention_Control_with_Scribble} helps mitigate penalties on false predictions related to thin scribbles,  this adjustment alone does not fully resolve the inherent sparsity of scribbles. To address this limitation, we propose a method to modify the input scribble prompt for more effective guidance without requiring additional training or modules.
One key challenge is that scribbles may initially be too narrow, leading to imprecise cross-attention with the target object, resulting in degraded quality or missing objects, as seen in \cref{fig:scribble_propagation}.
To overcome this, we introduce an iterative scribble expansion based on the reverse process's timestep.
This approach is inspired by the denoising stages in P2 weighting~\cite{choi2022perception_p2_weighting}, which identifies the reverse process in diffusion models as consisting of coarse, content, and clean-up phases.
In the early denoising stage, a general image is generated, followed by more detailed refinement.
By expanding the scribble prompt during the early stages of denoising, a coarse outline is created, which is progressively refined, leading to improved alignment with the target regions and more effective guidance.


DiffSeg~\cite{shuai2024diffseg} proposes zero-shot semantic segmentation by aggregating self-attention maps during the denoising process to reconstruct images, as the self-attention from U-Net layers highlights patches that are semantically similar. Inspired by this, we adopt a method proposed in DiffSeg without adding extra modules or training. Specifically, we aggregates $H \times W$ self-attention maps, $\mathcal{A}_{\texttt{self}}$, which integrate the varying resolutions of self-attention maps from different levels of the layers. Through this process, we obtain $\tilde{{\mathcal{A}}^{s}}$ for each scribble $s$, representing the mean distribution of self-attention activations within the scribble region $S$. 
Utilizing these self-attention maps $\tilde{{\mathcal{A}}^{s}}$ and $\mathcal{A}_{\texttt{self}}$, the decision to extend the scribble region $S$ is made by selecting candidate anchors near the boundary $B_{s}$ within a certain distance.
The distance $\mathcal{D}$ between the scribble prompt $s$ and an anchor $(x, y) \notin S$ near $B_{s}$ is computed using the Kullback-Leibler divergence $\mathcal{D}_{KL}$ as:

\begin{equation}
\begin{aligned}
    \mathcal{D}^{s} (x,y)
    &=
    \frac{1}{2} \mathcal{D}_{KL}
    \left(
        \tilde{{\mathcal{A}}^{s}} \middle| \mathcal{A}_{\texttt{self}}[x,y]
    \right) \\
    & +
    \frac{1}{2} \mathcal{D}_{KL}
    \left(
        \mathcal{A}_{\texttt{self}}[x,y] \middle| \tilde{{\mathcal{A}}^{s}}
    \right).
    \label{eq:distance_D}
\end{aligned}
\end{equation}
Finally, anchors adjacent to $\mathcal{B}_s$ with a distance below the threshold $\tau$ are selected as candidates for extension into each scribble region $\mathcal{S}$.
The $k$ anchors with the lowest distance are then collected into the scribble as:
\begin{equation}
\begin{aligned}
    \mathcal{S}' &= \text{argmin}_{(x, y) \in \mathcal{N}(B_s)}
    \left\{
        \mathcal{D}^{s} (x,y) |
        \forall s \in \mathcal{S}
    \right\}_{k},
    \\
    \label{eq:token_selection}
\end{aligned}
\end{equation}
where $\mathcal{N}(B_s)$ represents the neighborhood of $B_s$. This allows clustering in regions where the scribble regions $\mathcal{S}'$, which have high self-attention similarity with the scribble region $\mathcal{S}$, can be identified and merged with the existing $\mathcal{S}$ to update the scribble area.


\section{Experiments}
\label{sec:Experiments}

\noindent
Our method is implemented on the GLIGEN~\cite{li2023gligen} baseline. GLIGEN allows the use of bounding boxes as grounding inputs, so we first generate bounding boxes that encompass the scribbles, adding 5\% padding to both the width and height of each box. These bounding boxes are then used as grounding inputs for GLIGEN.




\subsection{Experimental Setup}

\noindent
\textbf{Dataset.}
The primary goal is to assess how well the generated objects match the scribbles in abstract shape and orientation.
Thus, we conduct our quantitative evaluation on the PASCAL-Scribble dataset~\cite{lin2016scribblesup_data}, a widely used benchmark for scribble-supervised semantic segmentation.
Additionally, each image is paired with a textual prompt based on its class name(s), formatted as ``\textit{a photo of} [\texttt{classname}] \textit{(and $\dots$)}."

For qualitative evaluation, we conducted additional experiments using detailed description-style prompts curated from previous works~\cite{chefer2023attend, xie2023boxdiff, liu2024training_free_composite} or generated by GPT-4~\cite{openai2023gpt}.

\noindent
\textbf{Evaluation Metrics.}
Our quantitative evaluation focuses on how well the generated images align with the scribble inputs while maintaining consistency with the corresponding prompts. To measure different aspects of the generation quality, we use several metrics.
The mean Intersection over Union (mIoU) score evaluates the alignment between the predicted masks of the generated objects using DeepLabV3+\cite{chen2017rethinking_deeplab} and the ground-truth masks. To assess text-to-image similarity, we use the CLIP-Score\cite{hessel2021clipscore}.

However, existing evaluation metrics are often insufficient to fully capture whether the scribble is fully encompassed by the generated object. To address this limitation, we introduce a novel metric, \textit{Scribble Ratio}, which quantifies the overlap between the areas defined by the original scribble and the masks obtained by DeepLabV3+.

\noindent
\textbf{Baselines.}
We compare our training-free Text-to-Image (T2I) generation method with two other approaches: BoxDiff~\cite{xie2023boxdiff} and DenseDiffusion~\cite{densediffusion}, both of which incorporate additional spatial inputs. BoxDiff primarily uses bounding box guidance but also includes scribble constraints in certain cases. DenseDiffusion, on the other hand, leverages region masks for image synthesis. For a fair comparison, we run BoxDiff experiments using the GLIGEN pipeline, while DenseDiffusion experiments are conducted using Stable Diffusion v1.5, as it directly modifies the attention layers in Stable Diffusion. In both cases, we applied scribble conditioning inputs to evaluate how well each method handles generation under scribble constraints.

Additionally, we include a fine-tuning-based comparison by evaluating ControlNet~\cite{zhang2023adding_controlnet} on the PASCAL-Scribble dataset. We fine-tune ControlNet using scribble inputs from the PASCAL-Scribble training set for 100 epochs.


\begin{figure*}[ht]
    \centering
    \parbox{1.02 \textwidth}{
        \centering
        \small
        \textit{
            A \textcolor{orange}{\textbf{lone astronaut}} exploring on a \textcolor{PineGreen}{\textbf{barren alien planet}}, with \textcolor{teal}{\textbf{distant galaxies}} visible in the sky, mysterious, vast, and lonely.
        }
    }
    \begin{subfigure}[t]{0.19\textwidth} \includegraphics[width=\textwidth]{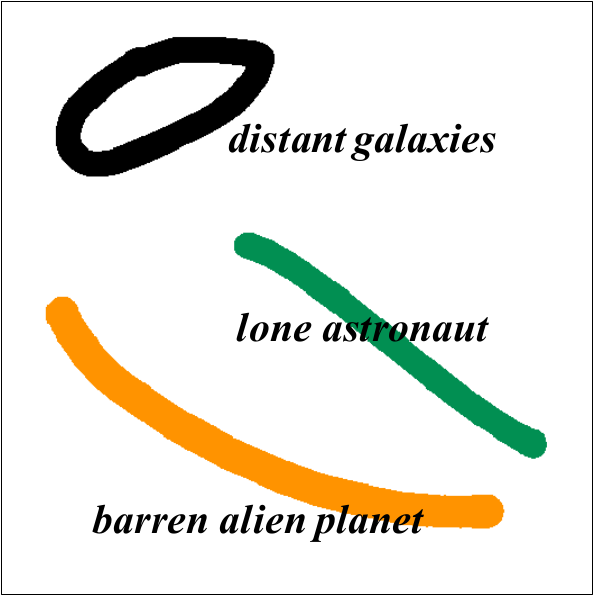} \end{subfigure}
    \begin{subfigure}[t]{0.19\textwidth} \includegraphics[width=\textwidth]{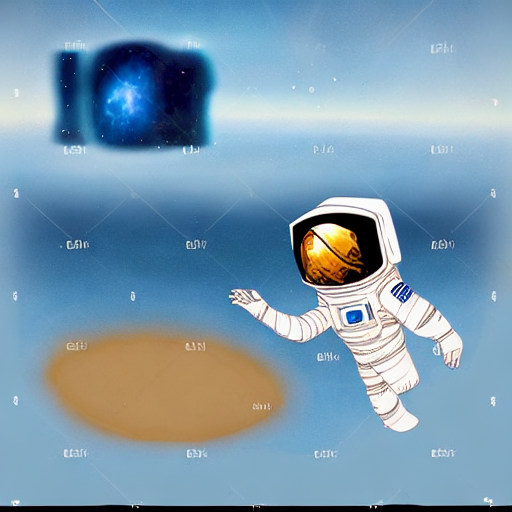} \end{subfigure}
    \begin{subfigure}[t]{0.19\textwidth} \includegraphics[width=\textwidth]{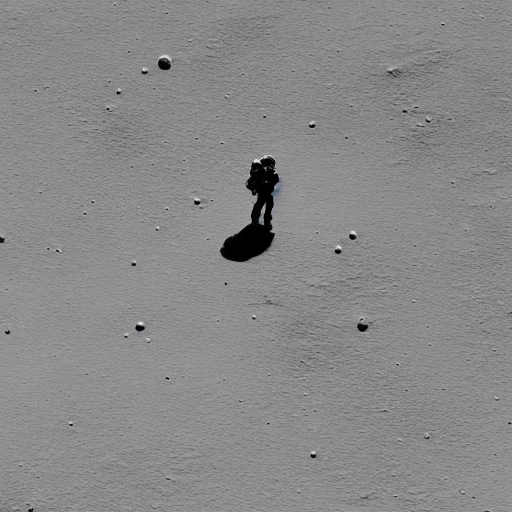} \end{subfigure}
    \begin{subfigure}[t]{0.19\textwidth} \includegraphics[width=\textwidth]{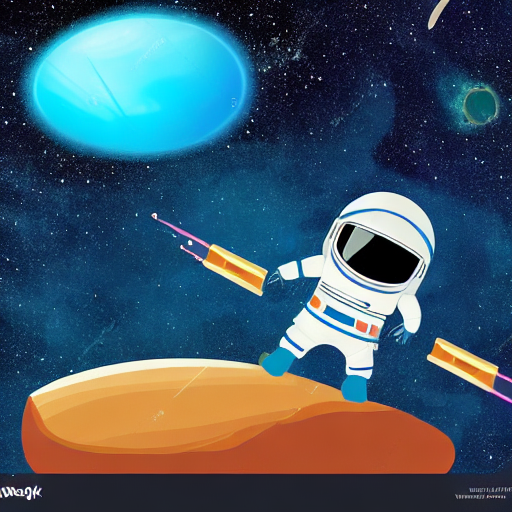} \end{subfigure}
    \begin{subfigure}[t]{0.19\textwidth} \includegraphics[width=\textwidth]{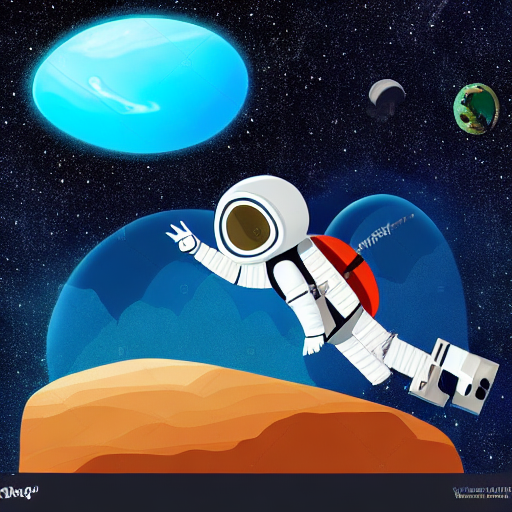} \end{subfigure}

    \vspace{-0.25em}
    \parbox{1.02 \textwidth}{
        \centering
        \small
        \textit{
            A \textcolor{red}{\textbf{Chinese dragon}} flying over a \textcolor{orange}{\textbf{medieval village}} at sunset, glowing embers in the sky, \textcolor{PineGreen}{\textbf{mountains}} in the background, fantasy, warm colors.
        }
    }
    \begin{subfigure}[t]{0.19\textwidth} \includegraphics[width=\textwidth]{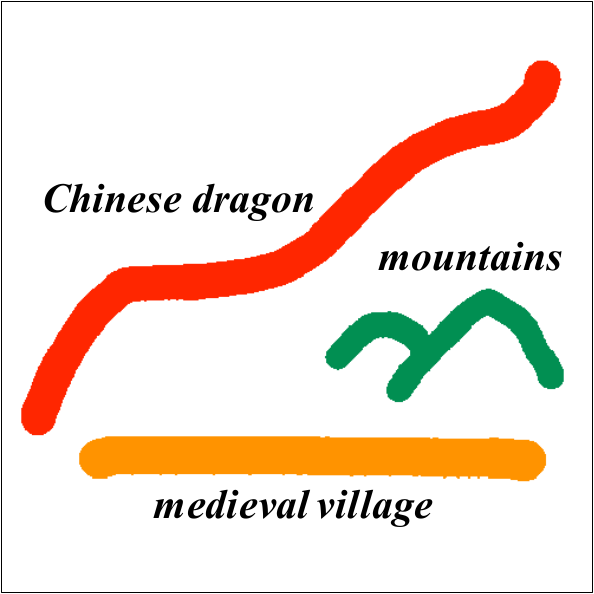} \end{subfigure}
    \begin{subfigure}[t]{0.19\textwidth} \includegraphics[width=\textwidth]{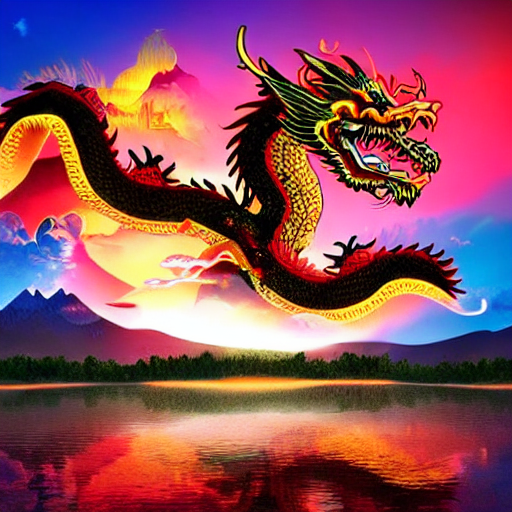} \end{subfigure}
    \begin{subfigure}[t]{0.19\textwidth} \includegraphics[width=\textwidth]{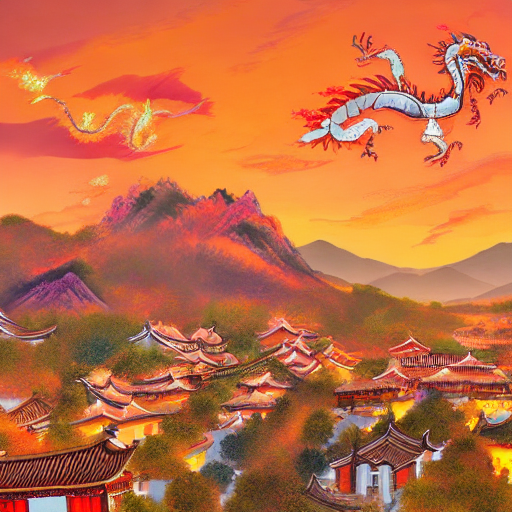} \end{subfigure}
    \begin{subfigure}[t]{0.19\textwidth} \includegraphics[width=\textwidth]{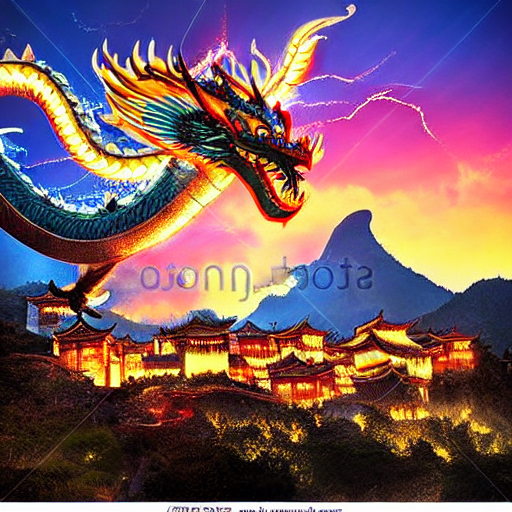} \end{subfigure}
    \begin{subfigure}[t]{0.19\textwidth} \includegraphics[width=\textwidth]{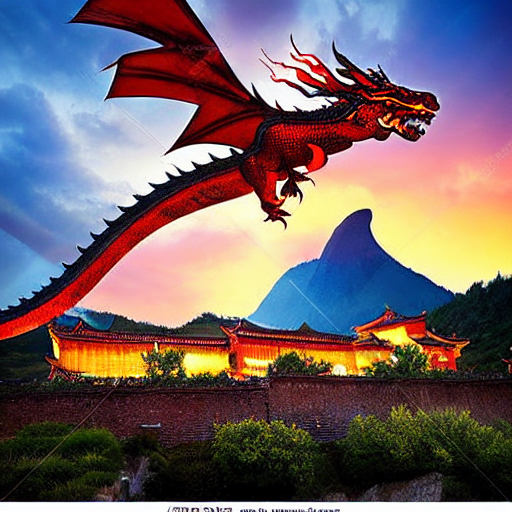} \end{subfigure}


    \vspace{-0.25em}
    \parbox[t]{\dimexpr\textwidth-2cm\relax}{
        \centering
        \small
        \textit{
            Detailed cyberpunk cityscape with a \textcolor{red}{\textbf{sleek car}} on a \textcolor{Mahogany}{\textbf{bustling street}}, surrounded by \textcolor{teal}{\textbf{skyscrapers}}, high-resolution.
        }
    }
    \begin{subfigure}[t]{0.19\textwidth} \includegraphics[width=\textwidth]{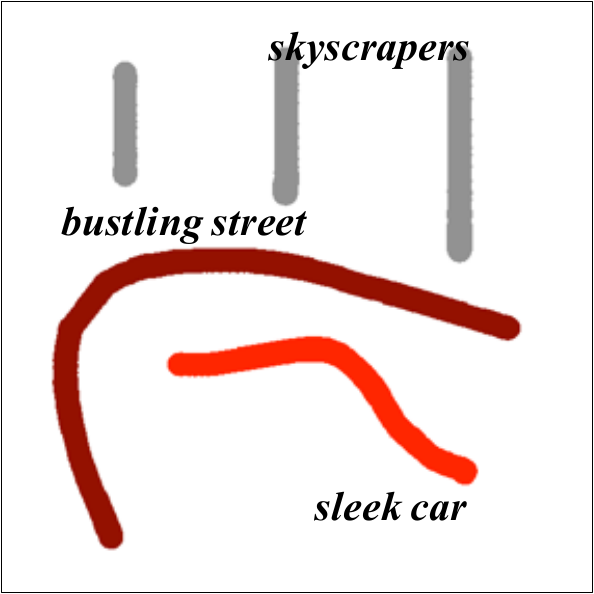} \end{subfigure}
    \begin{subfigure}[t]{0.19\textwidth} \includegraphics[width=\textwidth]{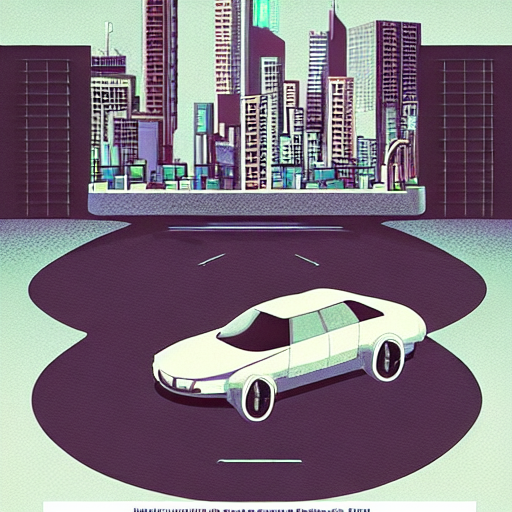} \end{subfigure}
    \begin{subfigure}[t]{0.19\textwidth} \includegraphics[width=\textwidth]{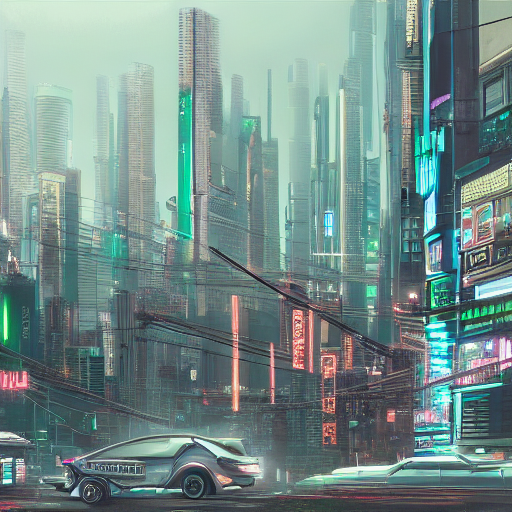} \end{subfigure}
    \begin{subfigure}[t]{0.19\textwidth} \includegraphics[width=\textwidth]{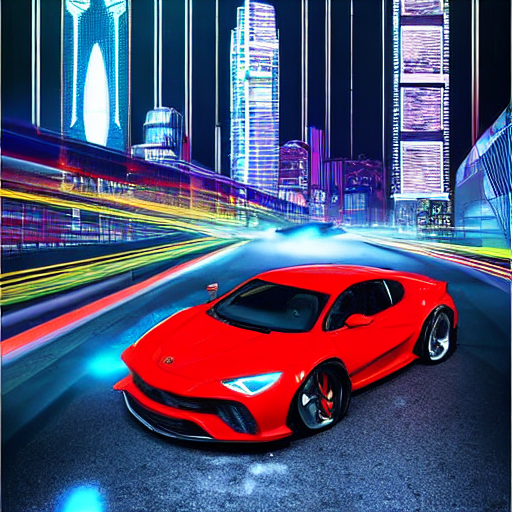} \end{subfigure}
    \begin{subfigure}[t]{0.19\textwidth} \includegraphics[width=\textwidth]{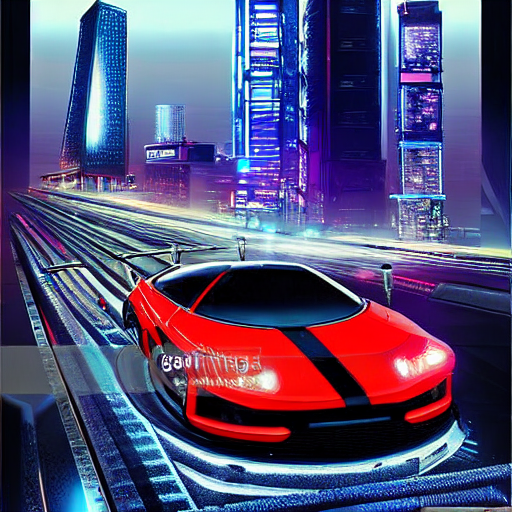} \end{subfigure}

    \vspace{-0.25em}
    \parbox{1.02 \textwidth}{
        \centering
        \small
        \textit{
            A pod of \textcolor{teal}{\textbf{dolphins}} leaping out of the water in an \textcolor{blue}{\textbf{ocean}} with a \textcolor{red}{\textbf{ship}} in the background.
        }
    }
    \begin{subfigure}[t]{0.19\textwidth}
        \includegraphics[width=\textwidth]{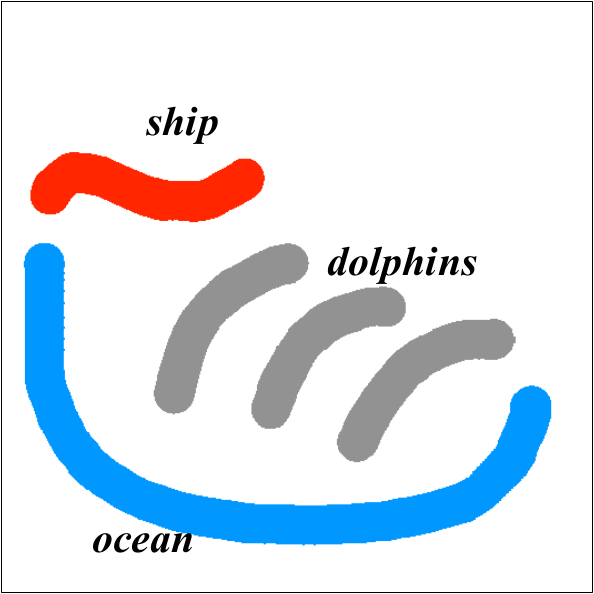}
        \caption{Scribbles}
    \end{subfigure}
    \begin{subfigure}[t]{0.19\textwidth}
        \includegraphics[width=\textwidth]{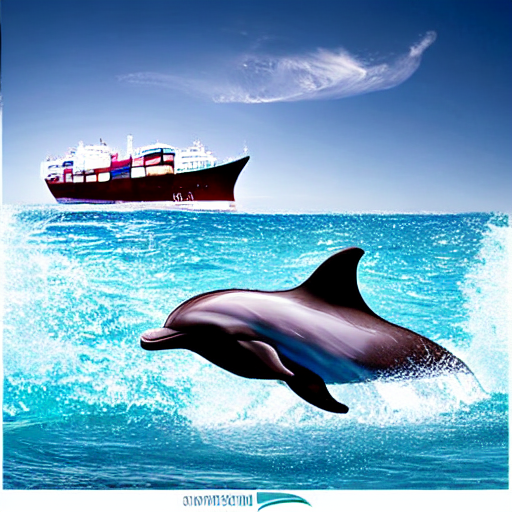}
        \caption{BoxDiff~\cite{xie2023boxdiff}}
    \end{subfigure}
    \begin{subfigure}[t]{0.19\textwidth}
        \includegraphics[width=\textwidth]{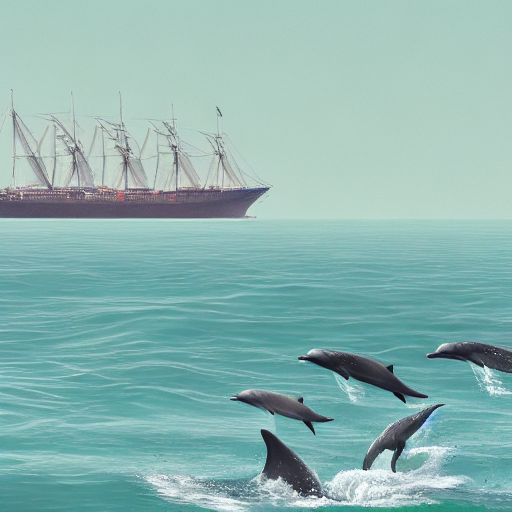}
        \caption{DenseDiff~\cite{densediffusion}}
    \end{subfigure}
    \begin{subfigure}[t]{0.19\textwidth}
        \includegraphics[width=\textwidth]{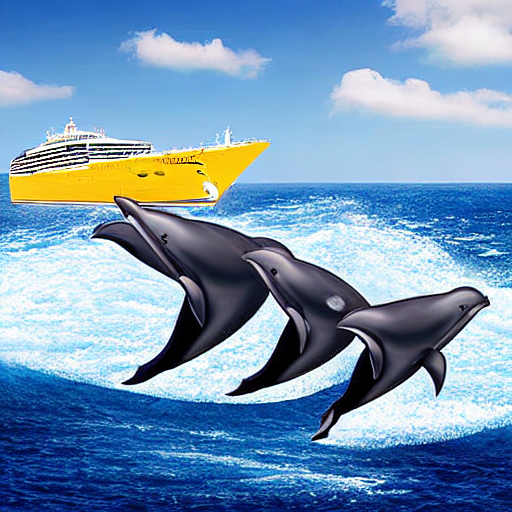}
        \caption{GLIGEN~\cite{li2023gligen}}
    \end{subfigure}
    \begin{subfigure}[t]{0.19\textwidth}
        \includegraphics[width=\textwidth]{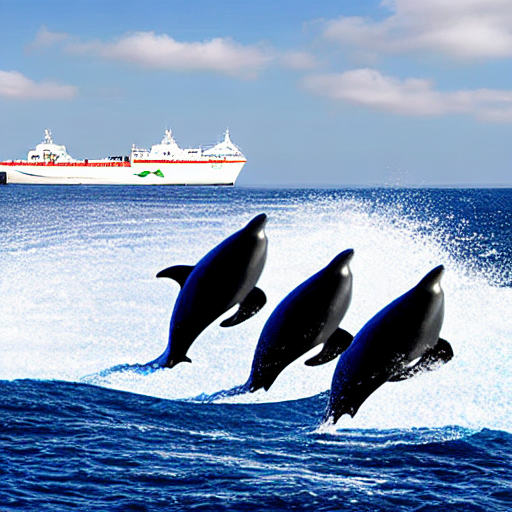}
        \caption{\MODELNAME \. (Ours)}
    \end{subfigure}
    
    \caption{
        \textbf{Qualitative comparison of Text-to-Image generation methods using scribble prompts.} \MODELNAME \. produces results that better align with the scribble inputs, particularly in orientations and abstract shapes of the objects.
    }
    \label{fig:qualitative_comparison}
    \vspace{-1.0em}
\end{figure*}

\begin{table}[!t]
    \centering
    \resizebox{\linewidth}{!}{\begin{tabular}{@{}lrrr@{}}
        \toprule
        Method   & mIoU ($\uparrow$)  & T2I Similarity ($\uparrow$) & Scribble Ratio ($\uparrow$) \\  
        \midrule

        BoxDiff~\cite{xie2023boxdiff} & 0.228 & \textbf{0.188}  & 0.406 \\
        DenseDiffusion~\cite{densediffusion} & 0.238 & 0.187 & 0.418 \\ 
        \MODELNAME\. (Ours) & \textbf{0.406} & 0.184 & \textbf{0.717} \\
        
        \bottomrule
    \end{tabular}}
    \caption{
        \textbf{Comparison of BoxDiff, DenseDiffusion, and \MODELNAME\. on the PASCAL-Scribble dataset~\cite{lin2016scribblesup_data}.}
         \MODELNAME\. demonstrates superior precision and consistency in interpreting thin scribble inputs.
    }
    \vspace{-0.5em}
    \label{tab:main_quantitative_01}
\end{table}

\begin{table}[!t]
    \small
    \centering
    \resizebox{\linewidth}{!}{\begin{tabular}{@{}l c rrr@{}}
        \toprule
        Method   & Fine-tuned & mIoU ($\uparrow$)  & Scribble Ratio ($\uparrow$) \\  
        \midrule

        ControlNet~\cite{zhang2023adding_controlnet} & \cmark & 0.165  & 0.229  \\ 
        \MODELNAME\. (Ours) & \xmark & \textbf{0.394} & \textbf{0.687} \\
        
        \bottomrule
    \end{tabular}}
    \vspace{-1em}
    \caption{
        \textbf{Comparison of \MODELNAME\. and fine-tuned ControlNet on the PASCAL-Scribble validation set.} 
        \MODELNAME\. significantly outperforms in both mIoU and Scribble Ratio.
    }
    \vspace{-0.5em}
    \label{tab:main_quantitative_02}
\end{table}

\subsection{Qualitative Results}
\noindent
\cref{fig:qualitative_comparison} compares the proposed \MODELNAME\. with other training-free text-to-image models.
Other methods generally exhibit poor alignment with the input scribbles. For example, in the case of the first row, with \emph{the astronaut} on a \emph{alien planet}, traditional methods often misinterpret the astronaut's spatial orientation, placing it incorrectly. In contrast, the \MODELNAME\, correctly positions the astronaut, aligning with the specified direction from the top-left to the bottom-right of the image.
This consistent preservation of scribble orientation is observed across all rows.
This highlights our central loss effectively captures the object direction and aligns it with the input scribble.

\begin{figure*}[ht]
    \centering
    \parbox{1.02 \textwidth}{
        \centering
        \small
        \textit{
            A photo of an \textcolor{PineGreen}{\textbf{airplane}}
        }
    }
    \begin{subfigure}[t]{0.16\textwidth} \includegraphics[width=\textwidth]{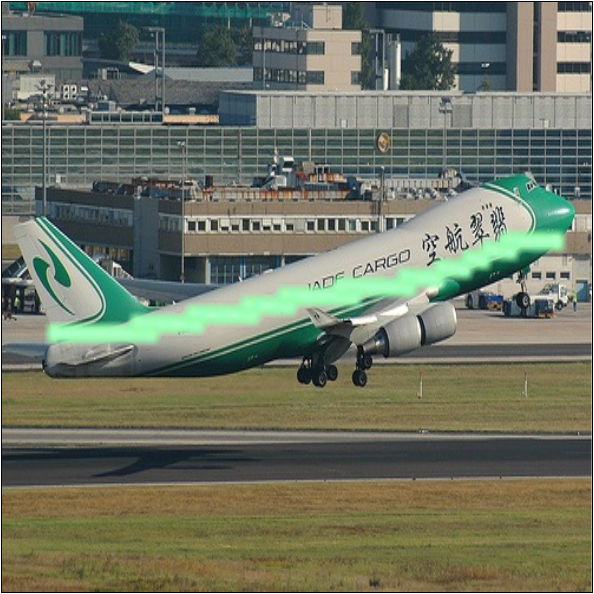} \end{subfigure}
    \begin{subfigure}[t]{0.16\textwidth} \includegraphics[width=\textwidth]{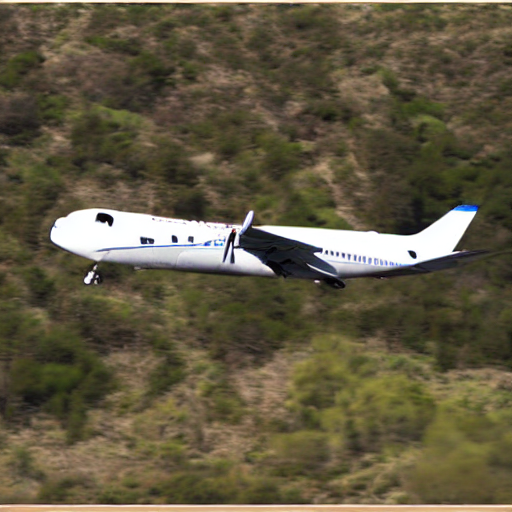} \end{subfigure}
    \begin{subfigure}[t]{0.16\textwidth} \includegraphics[width=\textwidth]{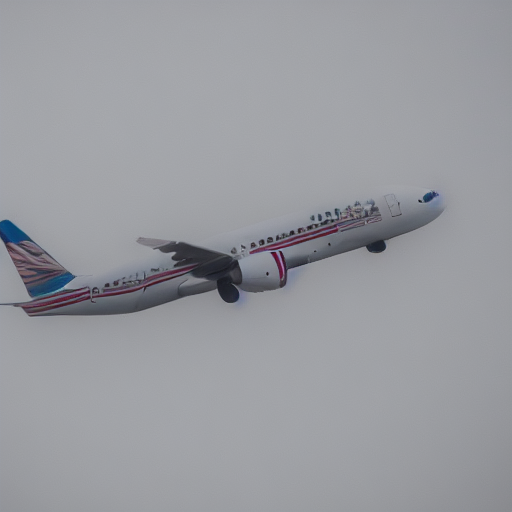} \end{subfigure}
    \begin{subfigure}[t]{0.16\textwidth} \includegraphics[width=\textwidth]{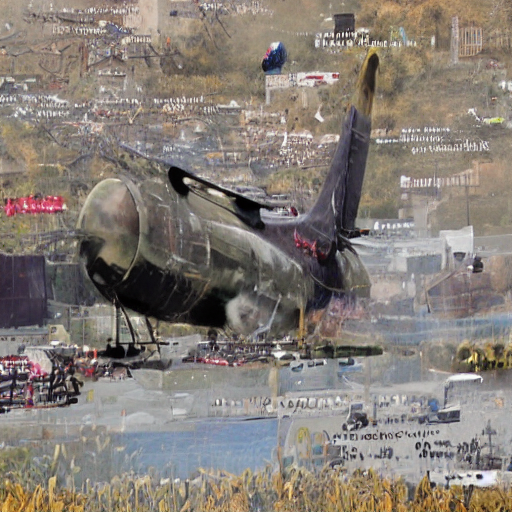} \end{subfigure}
    \begin{subfigure}[t]{0.16\textwidth} \includegraphics[width=\textwidth]{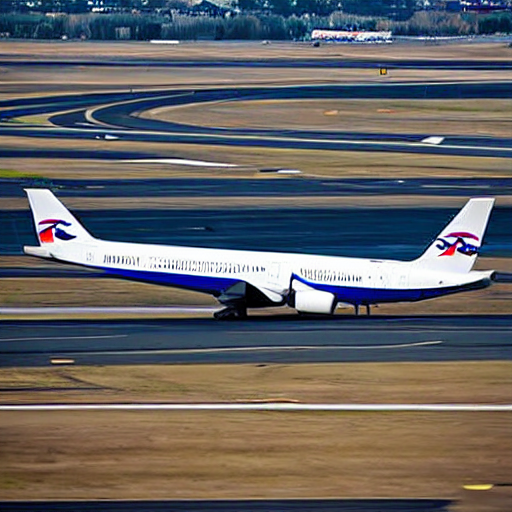} \end{subfigure}
    \begin{subfigure}[t]{0.16\textwidth} \includegraphics[width=\textwidth]{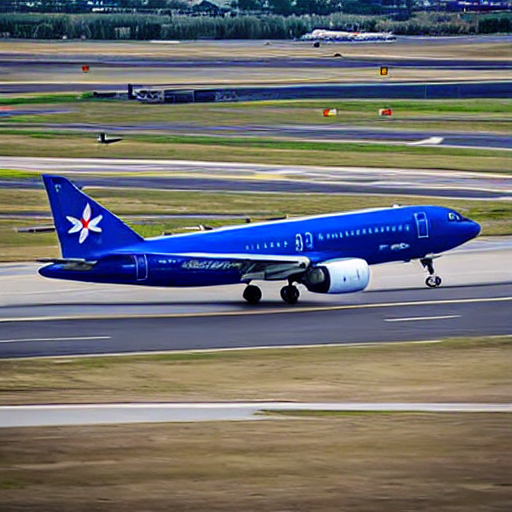} \end{subfigure}

    \vspace{-0.25em}
    \parbox{1.02 \textwidth}{
        \centering
        \small
        \textit{
            A photo of a \textcolor{PineGreen}{\textbf{horse}}
        }
    }
    \begin{subfigure}[t]{0.16\textwidth} \includegraphics[width=\textwidth]{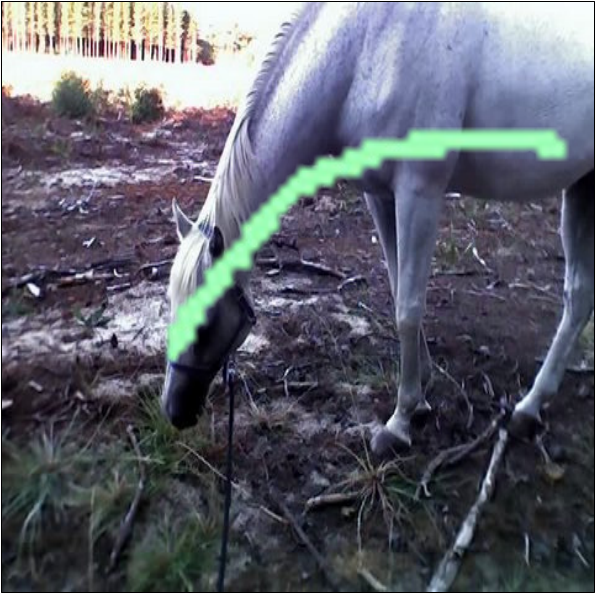} \end{subfigure}
    \begin{subfigure}[t]{0.16\textwidth} \includegraphics[width=\textwidth]{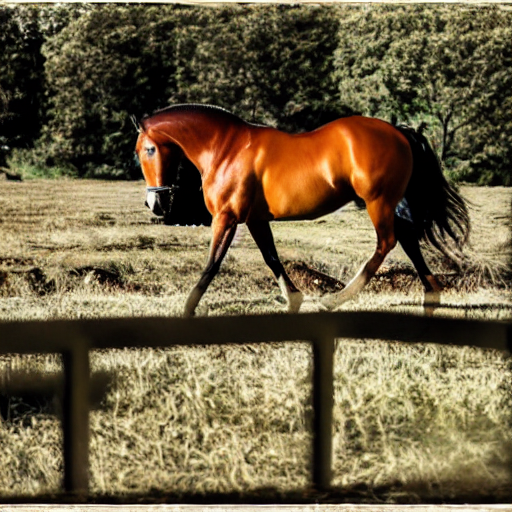} \end{subfigure}
    \begin{subfigure}[t]{0.16\textwidth} \includegraphics[width=\textwidth]{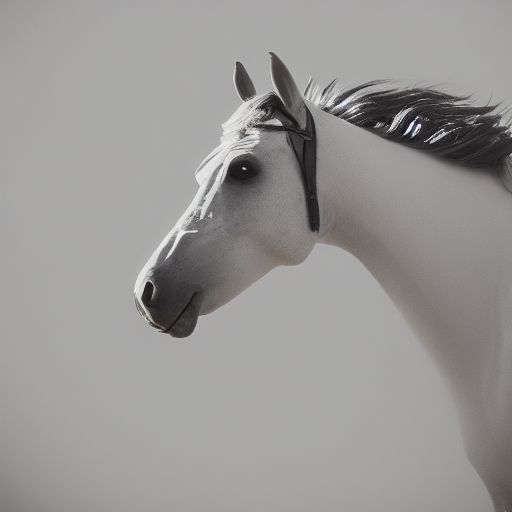} \end{subfigure}
    \begin{subfigure}[t]{0.16\textwidth} \includegraphics[width=\textwidth]{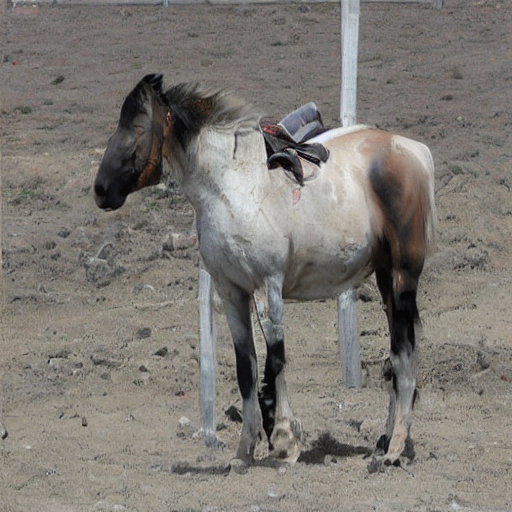} \end{subfigure}
    \begin{subfigure}[t]{0.16\textwidth} \includegraphics[width=\textwidth]{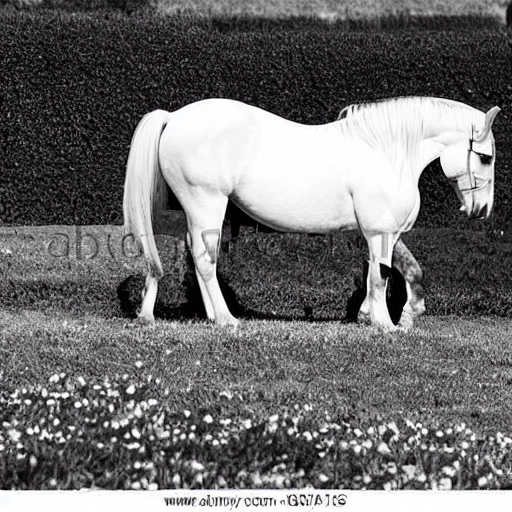} \end{subfigure}
    \begin{subfigure}[t]{0.16\textwidth} \includegraphics[width=\textwidth]{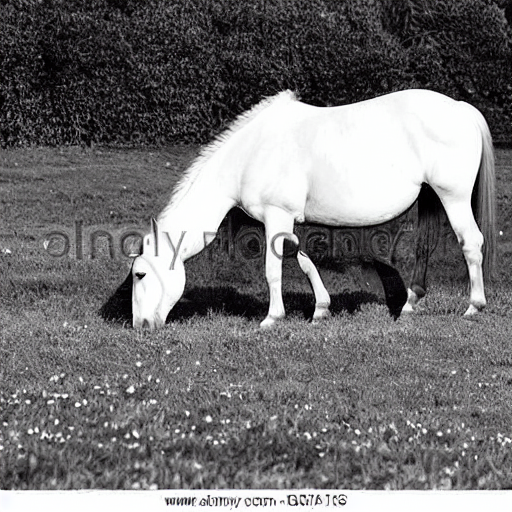} \end{subfigure}

    \vspace{-0.25em}
    \parbox{1.02 \textwidth}{
        \centering
        \small
        \textit{
            A photo of a \textcolor{PineGreen}{\textbf{monitor}}
        }
    }

    \begin{subfigure}[t]{0.16\textwidth} \includegraphics[width=\textwidth]{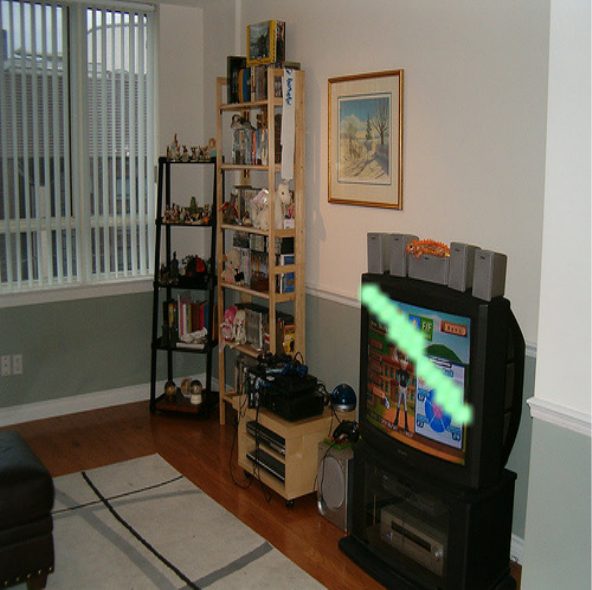} \end{subfigure}
    \begin{subfigure}[t]{0.16\textwidth} \includegraphics[width=\textwidth]{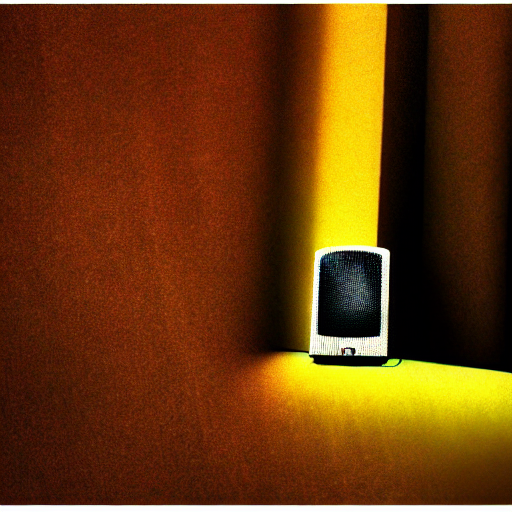} \end{subfigure}
    \begin{subfigure}[t]{0.16\textwidth} \includegraphics[width=\textwidth]{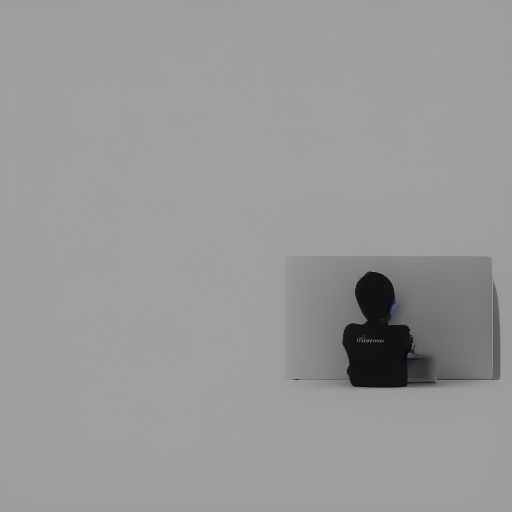} \end{subfigure}
    \begin{subfigure}[t]{0.16\textwidth} \includegraphics[width=\textwidth]{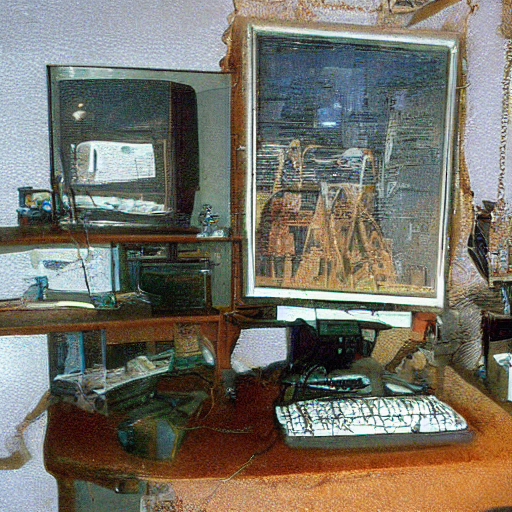} \end{subfigure}
    \begin{subfigure}[t]{0.16\textwidth} \includegraphics[width=\textwidth]{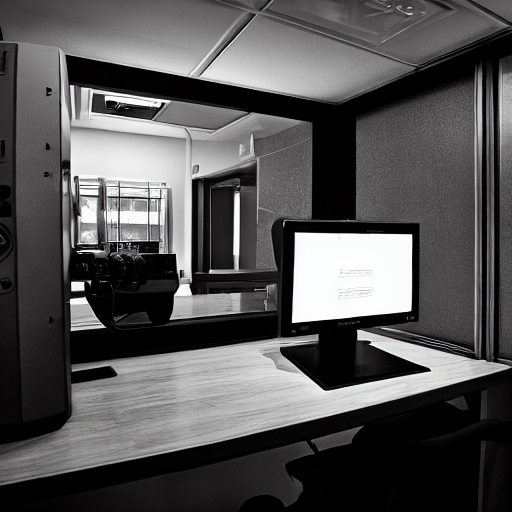} \end{subfigure}
    \begin{subfigure}[t]{0.16\textwidth} \includegraphics[width=\textwidth]{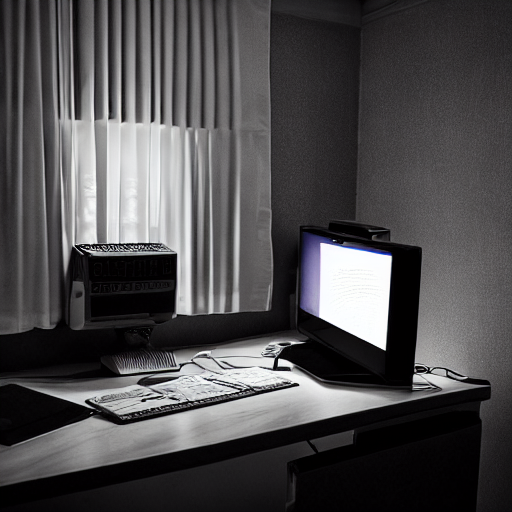} \end{subfigure}

    \vspace{-0.25em}
    \parbox{1.02 \textwidth}{
        \centering
        \small
        \textit{
            A photo of a \textcolor{PineGreen}{\textbf{cat}}
        }
    }
    \begin{subfigure}[t]{0.16\textwidth}
        \includegraphics[width=\textwidth]{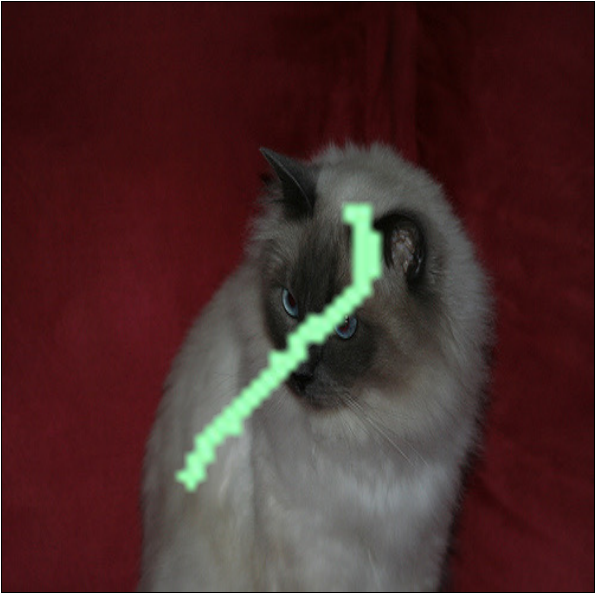}
        \caption{Scribbles~\cite{lin2016scribblesup_data}}
    \end{subfigure}
    \begin{subfigure}[t]{0.16\textwidth}
        \includegraphics[width=\textwidth]{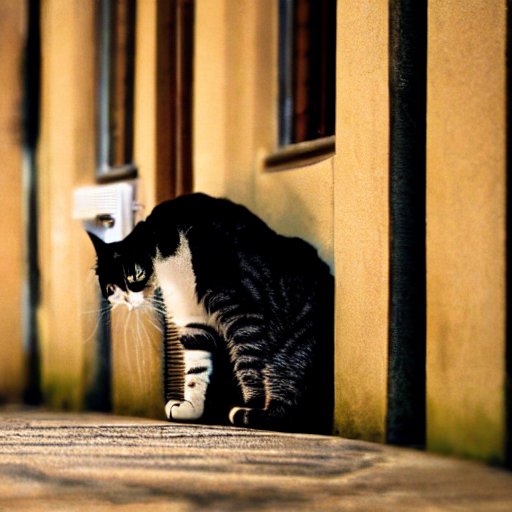}
        \caption{BoxDiff~\cite{xie2023boxdiff}}
    \end{subfigure}
    \begin{subfigure}[t]{0.16\textwidth}
        \includegraphics[width=\textwidth]{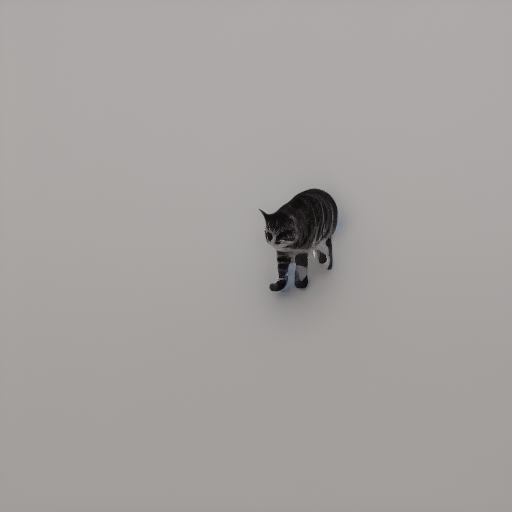}
        \caption{DenseDiff~\cite{densediffusion}}
    \end{subfigure}
    \begin{subfigure}[t]{0.16\textwidth}
        \includegraphics[width=\textwidth]{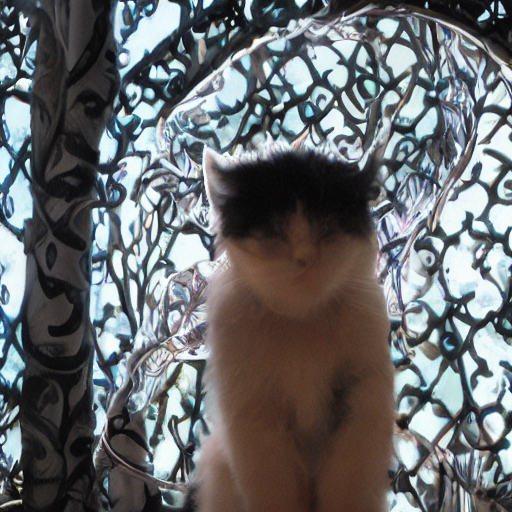}
        \caption{ControlNet~\cite{mo2024freecontrol}}
    \end{subfigure}
    \begin{subfigure}[t]{0.16\textwidth}
        \includegraphics[width=\textwidth]{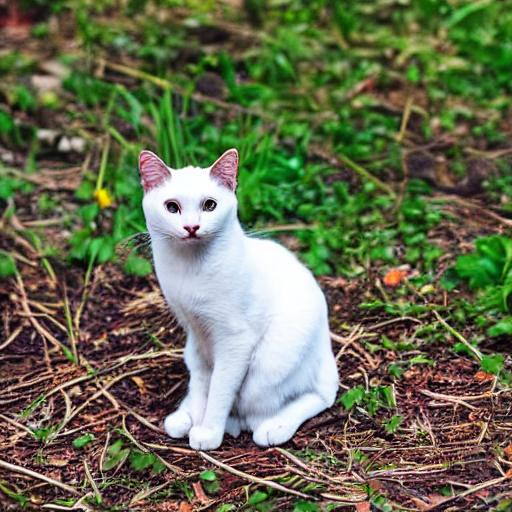}
        \caption{GLIGEN~\cite{li2023gligen}}
    \end{subfigure}
    \begin{subfigure}[t]{0.16\textwidth}
        \includegraphics[width=\textwidth]{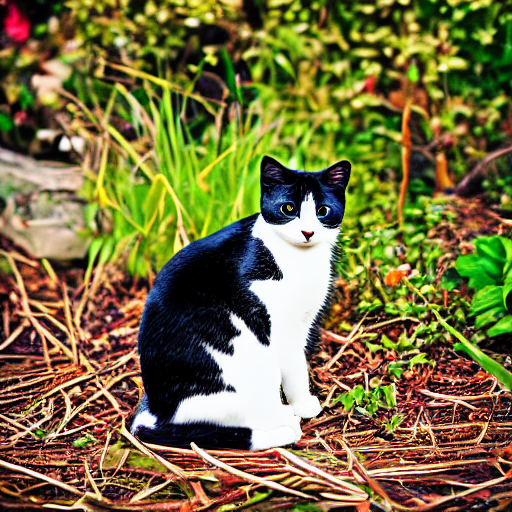}
        \caption{\MODELNAME \. (Ours)}
    \end{subfigure}
    
    \caption{
        \textbf{Qualitative results on the PASCAL-Scribble dataset~\cite{lin2016scribblesup_data}.} Comparison of various Text-to-Image generation methods, including the ControlNet fine-tuned on the training dataset. \MODELNAME \. demonstrates superior alignment with the input scribbles, particularly in handling abstract shapes and object orientations.
    }
    \vspace{-1.em}
    \label{fig:qualitative_comparison_pascal}
\end{figure*}

\cref{fig:qualitative_comparison_pascal} presents a qualitative comparison of existing text-to-image diffusion models on the PASCAL-Scribble dataset, including a comparison between our \MODELNAME\. and the fine-tuned ControlNet.
Despite not requiring additional training, \MODELNAME\. shows superior performance in reflecting the scribble prompts.
ControlNet, by contrast, lacks explicit learning of the scribble's direction, leading to suboptimal alignment. By leveraging moment alignment, \MODELNAME\. better captures the intended scribble prompt, surpassing both training-free and fine-tuned methods in handling scribble inputs.

\subsection{Quantitative Results}
\noindent
\cref{tab:main_quantitative_01} shows that \MODELNAME\. outperforms other methods by a significant margin.
In addition to adhering closely to the target input, it achieves higher consistency, as evidenced by its strong performance in the mIoU score.
While the T2I Similarity score does not show a significant difference across methods, our approach focuses on satisfying the constraints provided by the scribble input rather than enhancing semantic alignment with the textual prompt. \MODELNAME\. maintains a comparable T2I Similarity score while significantly improving performance in terms of mIoU and Scribble Ratio, demonstrating its ability to better adhere to scribble guidance.

In \cref{tab:main_quantitative_02}, we compare \MODELNAME\. with ControlNet finetuned on a validation set of the PASCAL-Scribble dataset.  Compared to the fine-tuned ControlNet with scribbles, our method demonstrates superior performance in alignment with the scribbles. Specifically, it achieves a 0.23 point increase in the mIoU score and a 0.46 gain in the Scribble Ratio score, indicating that our method is effective in the use of scribbles.


\begin{table}[tb!]
    \centering
    \resizebox{\linewidth}{!}{\begin{tabular}{@{}lrrr@{}}
        \toprule
        Method & Scribble Alignment ($\uparrow$)  & Text Prompt Fidelity ($\uparrow$) & Overall Quality ($\uparrow$) \\  
        \midrule

        BoxDiff~\cite{xie2023boxdiff}        &   5.67\% &   5.00\% &  3.00\% \\
        DenseDiffusion~\cite{densediffusion} &   0.67\% &   5.67\% &  1.33\% \\
        GLIGEN~\cite{li2023gligen}           &  18.33\% &  37.67\% & 28.33\% \\
        \MODELNAME\. (Ours)                  &  \textbf{75.33\%} &  \textbf{51.67\%} & \textbf{67.33\%} \\
        \bottomrule
    \end{tabular}}
    \caption{
        \textbf{User study results.} Comparing Text-to-Image generation methods based on Scribble Alignment, Text Prompt Fidelity, and Overall Quality.
    }
    \label{tab:user_study}
\end{table}

\newpage

\subsection{User Study}

\noindent
We further conducted a user study to assess the alignment and fidelity of generated images. 
Using the same seed, we generate images for 10 randomly selected prompts across each method.
30 participants were asked to select the best image that reflects the input scribble.
Each case is evaluated in three aspects: alignment with scribble, text prompt fidelity, and overall quality. 
As shown in \cref{tab:user_study}, \MODELNAME\. achieved the highest percentage of votes against other methods. 
For a detailed setup of the user study, please refer to the supplementary material~\cref{sec:supple_user_study}.


\begin{figure}[ht]
    \centering
    \begin{subfigure}[b]{0.24\linewidth}
        \centering
        \includegraphics[width=\linewidth]{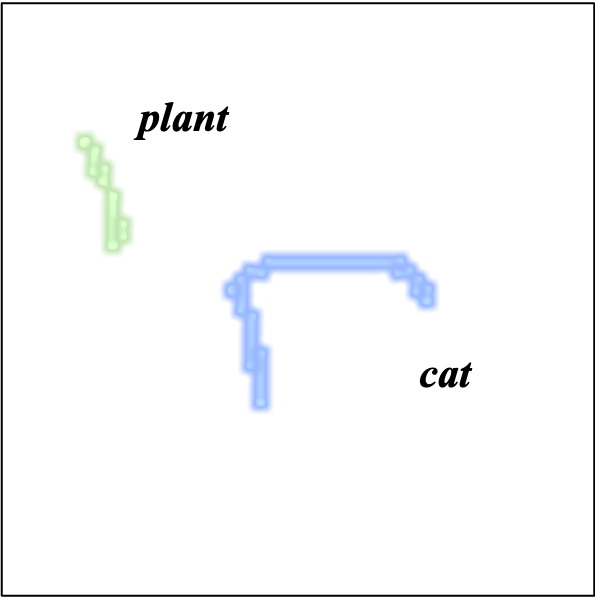}
        \captionsetup{justification=centering}
        \caption{
            Scribble
        }
    \end{subfigure}
    \begin{subfigure}[b]{0.24\linewidth}
        \centering
        \includegraphics[width=\linewidth]{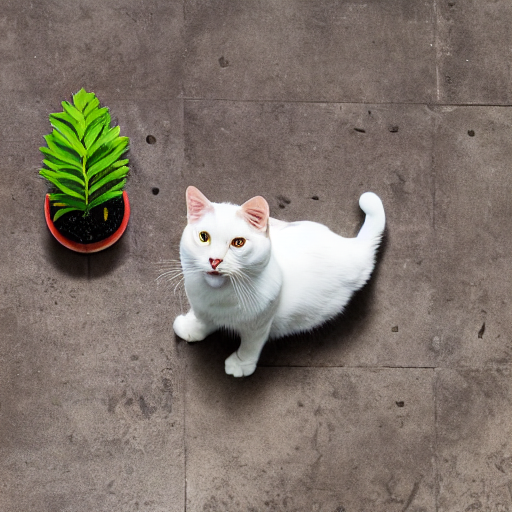}
        \captionsetup{justification=centering}
        \caption{
            w/o moment
        }
    \end{subfigure}
    \begin{subfigure}[b]{0.24\linewidth}
        \centering
        \includegraphics[width=\linewidth]{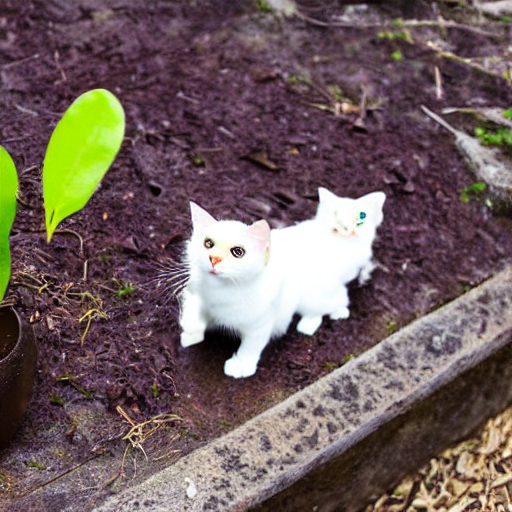}
        \captionsetup{justification=centering}
        \caption{
            w/o prop
        }
    \end{subfigure}
    \begin{subfigure}[b]{0.24\linewidth}
        \centering
        \includegraphics[width=\linewidth]{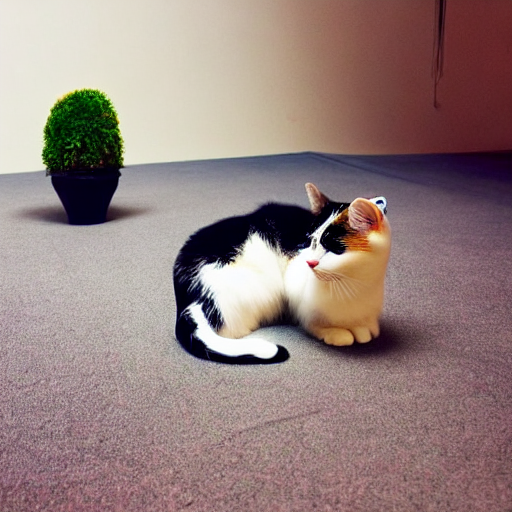}
        \captionsetup{justification=centering}
        \caption{
            \MODELNAME
        }
    \end{subfigure}
    \caption{
        \textbf{Ablation study on the PASCAL-Scribble dataset.} Comparison of qualitative results with and without key components on the same random seed.
    }
    \label{fig:ablation_study}
    \vspace{-0.5em}
\end{figure}

\newpage

\subsection{Ablation Study}


\noindent \textbf{Moment Loss.}
Moment loss $\mathcal{L}_{\texttt{moment}}$ enhances the precision of alignment and orientation with the target scribble. As shown in \cref{fig:moment_loss} and \cref{fig:ablation_study}, without moment loss, the generated object (\textbf{\textit{e.g.,}} \emph{cat}) may appear misaligned or face an incorrect direction relative to the scribble.
By incorporating moment loss, the cross-attention better aligns the object's orientation with the intended direction of the scribble, resulting in a more accurate final output.

\noindent \textbf{Scribble Propagation.}
Scribble propagation is designed to handle the sparse and thin nature of scribble annotations, as discussed in \cref{sec:Scribble_Propagation}. \cref{fig:scribble_propagation} demonstrates that, without propagation, scribbles remain narrow and constrained (e.g., timestep 901), leading to incomplete object representation. With scribble propagation, scribbles expand and improve object coverage by timestep 701. In \cref{fig:ablation_study}, the use of scribble propagation produces more coherent, complete, and higher-quality results compared to models without it.
For a detailed quantitative analysis of the ablation study, please refer to the supplementary material~\cref{sec:supple_ablation_study}.

\section{Conclusion}



\noindent
Our method overcomes the limitations of traditional bounding boxes and region masks, which often fail to capture abstract shapes and object orientations efficiently.
However, the sparse and thin nature of scribbles can hinder precise control, we mitigate this by introducing two key components: (1) moment loss to align object orientation with scribble direction, and (2) scribble propagation to enhance sparse scribble inputs into complete masks.
Experimental results show that \MODELNAME surpasses both training-free and fine-tuning methods across various metrics, including the new Scribble Ratio.
Our approach consistently improves object orientation and spatial alignment while maintaining fidelity to textual prompts.

\noindent \textbf{Acknowledgement.}
We would like to express our gratitude to Jaejin Lee, Minhee Lee, Hannah Park, and Jihoon Lee for their valuable discussions and inspiration.




{
    \small
    \bibliographystyle{ieee_fullname}
    \bibliography{references}

\begin{thebibliography}{10}\itemsep=-1pt

\bibitem{agarwal2023star}
Aishwarya Agarwal, Srikrishna Karanam, KJ Joseph, Apoorv Saxena, Koustava Goswami, and Balaji~Vasan Srinivasan.
\newblock A-star: Test-time attention segregation and retention for text-to-image synthesis.
\newblock In {\em Proceedings of the IEEE/CVF International Conference on Computer Vision}, pages 2283--2293, 2023.

\bibitem{avrahami2023spatext}
Omri Avrahami, Thomas Hayes, Oran Gafni, Sonal Gupta, Yaniv Taigman, Devi Parikh, Dani Lischinski, Ohad Fried, and Xi Yin.
\newblock Spatext: Spatio-textual representation for controllable image generation.
\newblock In {\em Proceedings of the IEEE/CVF Conference on Computer Vision and Pattern Recognition}, pages 18370--18380, 2023.

\bibitem{babaud1986uniqueness_Gaussian_kernel}
Jean Babaud, Andrew~P Witkin, Michel Baudin, and Richard~O Duda.
\newblock Uniqueness of the gaussian kernel for scale-space filtering.
\newblock {\em IEEE transactions on pattern analysis and machine intelligence}, pages 26--33, 1986.

\bibitem{bansal2023universal}
Arpit Bansal, Hong-Min Chu, Avi Schwarzschild, Soumyadip Sengupta, Micah Goldblum, Jonas Geiping, and Tom Goldstein.
\newblock Universal guidance for diffusion models.
\newblock In {\em Proceedings of the IEEE/CVF Conference on Computer Vision and Pattern Recognition}, pages 843--852, 2023.

\bibitem{bao2024separate}
Zhipeng Bao, Yijun Li, Krishna~Kumar Singh, Yu-Xiong Wang, and Martial Hebert.
\newblock Separate-and-enhance: Compositional finetuning for text-to-image diffusion models.
\newblock In {\em ACM SIGGRAPH 2024 Conference Papers}, pages 1--10, 2024.

\bibitem{boykov2001interactive_scribble}
Yuri~Y Boykov and M-P Jolly.
\newblock Interactive graph cuts for optimal boundary \& region segmentation of objects in nd images.
\newblock In {\em Proceedings eighth IEEE international conference on computer vision. ICCV 2001}, volume~1, pages 105--112. IEEE, 2001.

\bibitem{chefer2023attend}
Hila Chefer, Yuval Alaluf, Yael Vinker, Lior Wolf, and Daniel Cohen-Or.
\newblock Attend-and-excite: Attention-based semantic guidance for text-to-image diffusion models.
\newblock {\em ACM Transactions on Graphics (TOG)}, 42(4):1--10, 2023.

\bibitem{chen2017rethinking_deeplab}
Liang-Chieh Chen.
\newblock Rethinking atrous convolution for semantic image segmentation.
\newblock {\em arXiv preprint arXiv:1706.05587}, 2017.

\bibitem{chen2022scribble2d5}
Qiuhui Chen and Yi Hong.
\newblock Scribble2d5: Weakly-supervised volumetric image segmentation via scribble annotations.
\newblock In {\em International Conference on Medical Image Computing and Computer-Assisted Intervention}, pages 234--243. Springer, 2022.

\bibitem{chen2023scribbleseg}
Xi Chen, Yau Shing~Jonathan Cheung, Ser-Nam Lim, and Hengshuang Zhao.
\newblock Scribbleseg: Scribble-based interactive image segmentation.
\newblock {\em arXiv preprint arXiv:2303.11320}, 2023.

\bibitem{cheng2021modular_scribble_video}
Ho~Kei Cheng, Yu-Wing Tai, and Chi-Keung Tang.
\newblock Modular interactive video object segmentation: Interaction-to-mask, propagation and difference-aware fusion.
\newblock In {\em Proceedings of the IEEE/CVF Conference on Computer Vision and Pattern Recognition}, pages 5559--5568, 2021.

\bibitem{choi2022perception_p2_weighting}
Jooyoung Choi, Jungbeom Lee, Chaehun Shin, Sungwon Kim, Hyunwoo Kim, and Sungroh Yoon.
\newblock Perception prioritized training of diffusion models.
\newblock In {\em Proceedings of the IEEE/CVF Conference on Computer Vision and Pattern Recognition}, pages 11472--11481, 2022.

\bibitem{dhariwal2021diffusion}
Prafulla Dhariwal and Alexander Nichol.
\newblock Diffusion models beat gans on image synthesis.
\newblock {\em Advances in neural information processing systems}, 34:8780--8794, 2021.

\bibitem{ding2024training_free_sketch_diff}
Sandra~Zhang Ding, Jiafeng Mao, and Kiyoharu Aizawa.
\newblock Training-free sketch-guided diffusion with latent optimization.
\newblock {\em arXiv preprint arXiv:2409.00313}, 2024.

\bibitem{dorent2020scribble_weakly_seg}
Reuben Dorent, Samuel Joutard, Jonathan Shapey, Sotirios Bisdas, Neil Kitchen, Robert Bradford, Shakeel Saeed, Marc Modat, S{\'e}bastien Ourselin, and Tom Vercauteren.
\newblock Scribble-based domain adaptation via co-segmentation.
\newblock In {\em Medical Image Computing and Computer Assisted Intervention--MICCAI 2020: 23rd International Conference, Lima, Peru, October 4--8, 2020, Proceedings, Part I 23}, pages 479--489. Springer, 2020.

\bibitem{epstein2023diffusion_self_guidance}
Dave Epstein, Allan Jabri, Ben Poole, Alexei~A Efros, and Aleksander Holynski.
\newblock Diffusion self-guidance for controllable image generation.
\newblock {\em arXiv preprint arXiv:2306.00986}, 2023.

\bibitem{flusser2006moment}
Jan Flusser.
\newblock Moment invariants in image analysis.
\newblock In {\em proceedings of world academy of science, engineering and technology}, volume~11, pages 196--201. Citeseer, 2006.

\bibitem{gafni2022make_a_scene}
Oran Gafni, Adam Polyak, Oron Ashual, Shelly Sheynin, Devi Parikh, and Yaniv Taigman.
\newblock Make-a-scene: Scene-based text-to-image generation with human priors.
\newblock In {\em European Conference on Computer Vision}, pages 89--106. Springer, 2022.

\bibitem{hertz2022prompt_cross_attention_control}
Amir Hertz, Ron Mokady, Jay Tenenbaum, Kfir Aberman, Yael Pritch, and Daniel Cohen-Or.
\newblock Prompt-to-prompt image editing with cross attention control.
\newblock {\em arXiv preprint arXiv:2208.01626}, 2022.

\bibitem{hessel2021clipscore}
Jack Hessel, Ari Holtzman, Maxwell Forbes, Ronan~Le Bras, and Yejin Choi.
\newblock Clipscore: A reference-free evaluation metric for image captioning.
\newblock {\em arXiv preprint arXiv:2104.08718}, 2021.

\bibitem{ho2020denoising_ddpm}
Jonathan Ho, Ajay Jain, and Pieter Abbeel.
\newblock Denoising diffusion probabilistic models.
\newblock {\em Advances in neural information processing systems}, 33:6840--6851, 2020.

\bibitem{ho2022classifier}
Jonathan Ho and Tim Salimans.
\newblock Classifier-free diffusion guidance.
\newblock {\em arXiv preprint arXiv:2207.12598}, 2022.

\bibitem{densediffusion}
Yunji Kim, Jiyoung Lee, Jin-Hwa Kim, Jung-Woo Ha, and Jun-Yan Zhu.
\newblock Dense text-to-image generation with attention modulation.
\newblock In {\em ICCV}, 2023.

\bibitem{kwon2022diffusion}
Mingi Kwon, Jaeseok Jeong, and Youngjung Uh.
\newblock Diffusion models already have a semantic latent space.
\newblock {\em arXiv preprint arXiv:2210.10960}, 2022.

\bibitem{li2023gligen}
Yuheng Li, Haotian Liu, Qingyang Wu, Fangzhou Mu, Jianwei Yang, Jianfeng Gao, Chunyuan Li, and Yong~Jae Lee.
\newblock Gligen: Open-set grounded text-to-image generation.
\newblock In {\em Proceedings of the IEEE/CVF Conference on Computer Vision and Pattern Recognition}, pages 22511--22521, 2023.

\bibitem{lin2016scribblesup_data}
Di Lin, Jifeng Dai, Jiaya Jia, Kaiming He, and Jian Sun.
\newblock Scribblesup: Scribble-supervised convolutional networks for semantic segmentation.
\newblock In {\em Proceedings of the IEEE conference on computer vision and pattern recognitio, Micheelsenn}, pages 3159--3167, 2016.

\bibitem{liu2024training_free_composite}
Jiaqi Liu, Tao Huang, and Chang Xu.
\newblock Training-free composite scene generation for layout-to-image synthesis.
\newblock {\em arXiv preprint arXiv:2407.13609}, 2024.

\bibitem{ma2023directed}
Wan-Duo~Kurt Ma, JP Lewis, W~Bastiaan Kleijn, and Thomas Leung.
\newblock Directed diffusion: Direct control of object placement through attention guidance.
\newblock {\em arXiv preprint arXiv:2302.13153}, 2023.

\bibitem{meng2021sdedit}
Chenlin Meng, Yutong He, Yang Song, Jiaming Song, Jiajun Wu, Jun-Yan Zhu, and Stefano Ermon.
\newblock Sdedit: Guided image synthesis and editing with stochastic differential equations.
\newblock {\em arXiv preprint arXiv:2108.01073}, 2021.

\bibitem{mo2024freecontrol}
Sicheng Mo, Fangzhou Mu, Kuan~Heng Lin, Yanli Liu, Bochen Guan, Yin Li, and Bolei Zhou.
\newblock Freecontrol: Training-free spatial control of any text-to-image diffusion model with any condition.
\newblock In {\em Proceedings of the IEEE/CVF Conference on Computer Vision and Pattern Recognition}, pages 7465--7475, 2024.

\bibitem{mukundan1998moment}
Ramakrishnan Mukundan and KR Ramakrishnan.
\newblock {\em Moment functions in image analysis: theory and applications}.
\newblock World scientific, 1998.

\bibitem{openai2023gpt}
R OpenAI.
\newblock Gpt-4 technical report. arxiv 2303.08774.
\newblock {\em View in Article}, 2(5), 2023.

\bibitem{park2022shape_guidied}
Dong~Huk Park, Grace Luo, Clayton Toste, Samaneh Azadi, Xihui Liu, Maka Karalashvili, Anna Rohrbach, and Trevor Darrell.
\newblock Shape-guided diffusion with inside-outside attention.
\newblock {\em arXiv preprint arXiv:2212.00210}, 2022.

\bibitem{phung2023grounded_Refocusing}
Quynh Phung, Songwei Ge, and Jia-Bin Huang.
\newblock Grounded text-to-image synthesis with attention refocusing.
\newblock {\em arXiv preprint arXiv:2306.05427}, 2023.

\bibitem{ramesh2021zero_dalle}
Aditya Ramesh, Mikhail Pavlov, Gabriel Goh, Scott Gray, Chelsea Voss, Alec Radford, Mark Chen, and Ilya Sutskever.
\newblock Zero-shot text-to-image generation.
\newblock In {\em International conference on machine learning}, pages 8821--8831. Pmlr, 2021.

\bibitem{rombach2022high_latent}
Robin Rombach, Andreas Blattmann, Dominik Lorenz, Patrick Esser, and Bj{\"o}rn Ommer.
\newblock High-resolution image synthesis with latent diffusion models.
\newblock In {\em Proceedings of the IEEE/CVF conference on computer vision and pattern recognition}, pages 10684--10695, 2022.

\bibitem{saharia2022photorealistic_imagen}
Chitwan Saharia, William Chan, Saurabh Saxena, Lala Li, Jay Whang, Emily~L Denton, Kamyar Ghasemipour, Raphael Gontijo~Lopes, Burcu Karagol~Ayan, Tim Salimans, et~al.
\newblock Photorealistic text-to-image diffusion models with deep language understanding.
\newblock {\em Advances in neural information processing systems}, 35:36479--36494, 2022.

\bibitem{shuai2024diffseg}
Zhihao Shuai, Yinan Chen, Shunqiang Mao, Yihan Zho, and Xiaohong Zhang.
\newblock Diffseg: A segmentation model for skin lesions based on diffusion difference.
\newblock {\em arXiv preprint arXiv:2404.16474}, 2024.

\bibitem{sohl2015deep}
Jascha Sohl-Dickstein, Eric Weiss, Niru Maheswaranathan, and Surya Ganguli.
\newblock Deep unsupervised learning using nonequilibrium thermodynamics.
\newblock In {\em International conference on machine learning}, pages 2256--2265. PMLR, 2015.

\bibitem{song2020denoising_ddim}
Jiaming Song, Chenlin Meng, and Stefano Ermon.
\newblock Denoising diffusion implicit models.
\newblock {\em arXiv preprint arXiv:2010.02502}, 2020.

\bibitem{song2019generative}
Yang Song and Stefano Ermon.
\newblock Generative modeling by estimating gradients of the data distribution.
\newblock {\em Advances in neural information processing systems}, 32, 2019.

\bibitem{song2020score}
Yang Song, Jascha Sohl-Dickstein, Diederik~P Kingma, Abhishek Kumar, Stefano Ermon, and Ben Poole.
\newblock Score-based generative modeling through stochastic differential equations.
\newblock {\em arXiv preprint arXiv:2011.13456}, 2020.

\bibitem{tang2023emergent}
Luming Tang, Menglin Jia, Qianqian Wang, Cheng~Perng Phoo, and Bharath Hariharan.
\newblock Emergent correspondence from image diffusion.
\newblock {\em Advances in Neural Information Processing Systems}, 36:1363--1389, 2023.

\bibitem{valvano2021learning_scribble_seg}
Gabriele Valvano, Andrea Leo, and Sotirios~A Tsaftaris.
\newblock Learning to segment from scribbles using multi-scale adversarial attention gates.
\newblock {\em IEEE Transactions on Medical Imaging}, 40(8):1990--2001, 2021.

\bibitem{voynov2023sketch_diff}
Andrey Voynov, Kfir Aberman, and Daniel Cohen-Or.
\newblock Sketch-guided text-to-image diffusion models.
\newblock In {\em ACM SIGGRAPH 2023 Conference Proceedings}, pages 1--11, 2023.

\bibitem{wang2005interactive_scribble_video}
Jue Wang, Pravin Bhat, R~Alex Colburn, Maneesh Agrawala, and Michael~F Cohen.
\newblock Interactive video cutout.
\newblock {\em ACM Transactions on Graphics (ToG)}, 24(3):585--594, 2005.

\bibitem{wang2023compositional}
Ruichen Wang, Zekang Chen, Chen Chen, Jian Ma, Haonan Lu, and Xiaodong Lin.
\newblock Compositional text-to-image synthesis with attention map control of diffusion models.
\newblock {\em arXiv preprint arXiv:2305.13921}, 2023.

\bibitem{wong2023scribbleprompt}
Hallee~E Wong, Marianne Rakic, John Guttag, and Adrian~V Dalca.
\newblock Scribbleprompt: Fast and flexible interactive segmentation for any medical image.
\newblock {\em arXiv preprint arXiv:2312.07381}, 2023.

\bibitem{xie2023boxdiff}
Jinheng Xie, Yuexiang Li, Yawen Huang, Haozhe Liu, Wentian Zhang, Yefeng Zheng, and Mike~Zheng Shou.
\newblock Boxdiff: Text-to-image synthesis with training-free box-constrained diffusion.
\newblock In {\em Proceedings of the IEEE/CVF International Conference on Computer Vision}, pages 7452--7461, 2023.

\bibitem{ye2023ip_adapter}
Hu Ye, Jun Zhang, Sibo Liu, Xiao Han, and Wei Yang.
\newblock Ip-adapter: Text compatible image prompt adapter for text-to-image diffusion models.
\newblock {\em arXiv preprint arXiv:2308.06721}, 2023.

\bibitem{yu2023freedom}
Jiwen Yu, Yinhuai Wang, Chen Zhao, Bernard Ghanem, and Jian Zhang.
\newblock Freedom: Training-free energy-guided conditional diffusion model.
\newblock In {\em Proceedings of the IEEE/CVF International Conference on Computer Vision}, pages 23174--23184, 2023.

\bibitem{zhang2023adding_controlnet}
Lvmin Zhang and Maneesh Agrawala.
\newblock Adding conditional control to text-to-image diffusion models.
\newblock {\em arXiv preprint arXiv:2302.05543}, 2023.

\bibitem{zhao2016energy}
Junbo Zhao, Michael Mathieu, and Yann LeCun.
\newblock Energy-based generative adversarial network.
\newblock {\em arXiv preprint arXiv:1609.03126}, 2016.

\end{thebibliography}
}


\newpage

\appendix

\startcontents[supplement] 

\setcounter{section}{0}   
\renewcommand{\thetable}{S\arabic{table}}
\setcounter{figure}{0}  
\renewcommand{\thefigure}{S\arabic{figure}}
\setcounter{table}{0}   
\renewcommand{\theequation}{S\arabic{equation}}
\setcounter{equation}{0}

\noindent\section*{Supplementary Materials}


\noindent
In this supplementary material, we provide detailed descriptions of the algorithm and implementation, additional qualitative comparisons, experimental results, a detailed user study setup, and limitations with discussion.


\noindent\subsubsection*{Table of Contents}


\begin{itemize}
    \item Details of Scribble Diffusion (\Cref{sec:supple_detailed_architecture})
    \item Implementation Details (\Cref{sec:supple_imple_detail})
    \item Overall Algorithm (\Cref{sec:supple_overall_algorithm})
    \item More Qualitative Results (\Cref{sec:supple_more_qualitative_results})
    \item Additional Ablation Studies (\Cref{sec:supple_ablation_study})
    \item User Study Details (\Cref{sec:supple_user_study})
    \item Limitation \& Discussion (\Cref{sec:limitation_and_discussion})
\end{itemize}



\section{Details of Scribble Diffusion}
\label{sec:supple_detailed_architecture}

\begin{figure}[ht]
    \centering
    \includegraphics[width=1.00\linewidth]{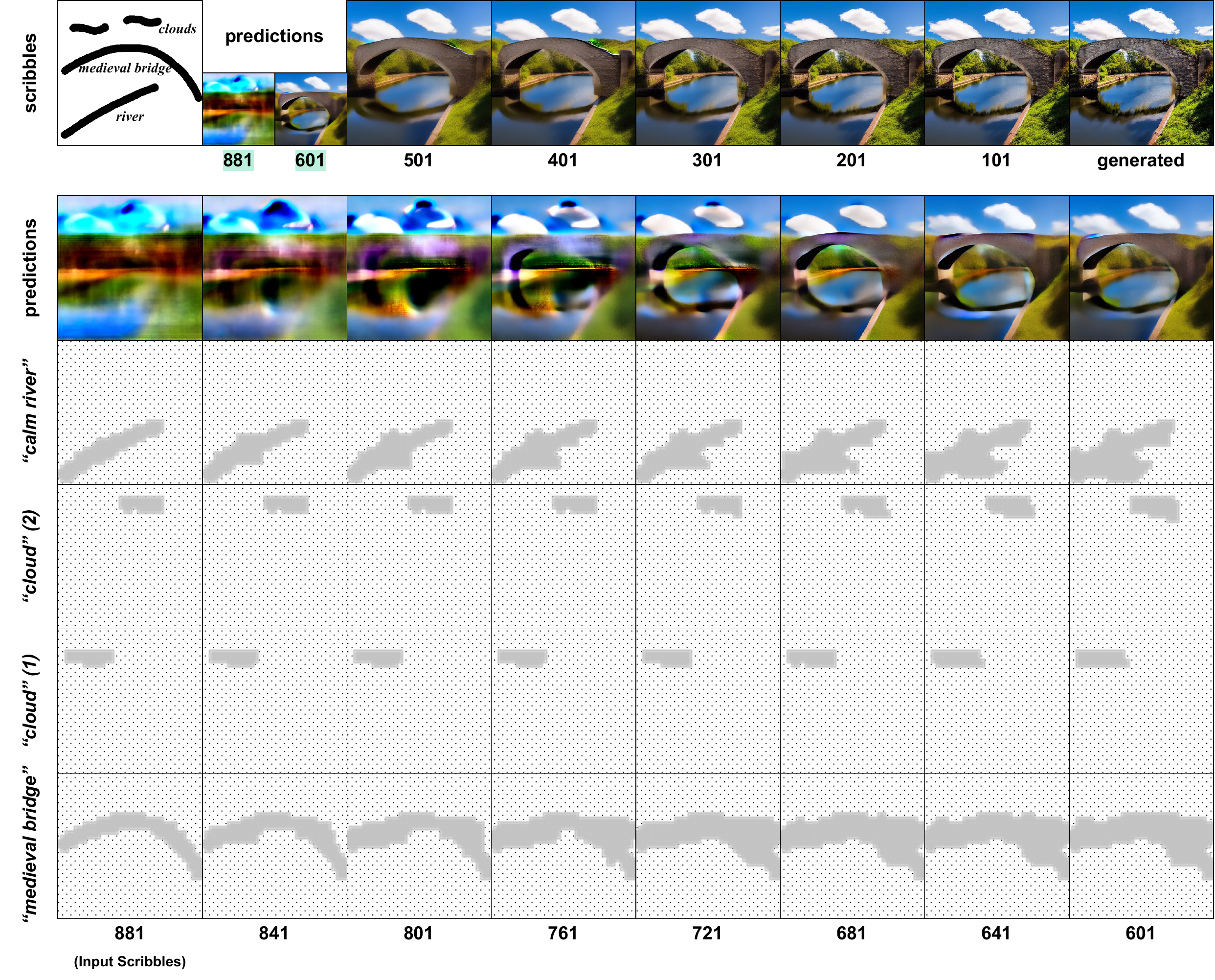}
    \caption{
        \textbf{Scribble Propagation.}
        At specific timesteps, our method extends the input scribble, improving alignment with the generated image.
        }
    \label{fig:appendix_scribble_propagation}
\end{figure}

\noindent
\cref{fig:appendix_scribble_propagation} shows images inferred from the scribble prompt with different timesteps.
As discussed in the P2 weighting~\cite{choi2022perception_p2_weighting}, we extend the scribble prompt at certain timesteps related to content generation, effectively enhancing alignment between the scribble and the image.

\begin{figure}[ht]
    \centering
    \parbox{1.00 \linewidth}{
        \centering
        \small
        {result without scribble propagation}
    }
    \includegraphics[width=1.00\linewidth]{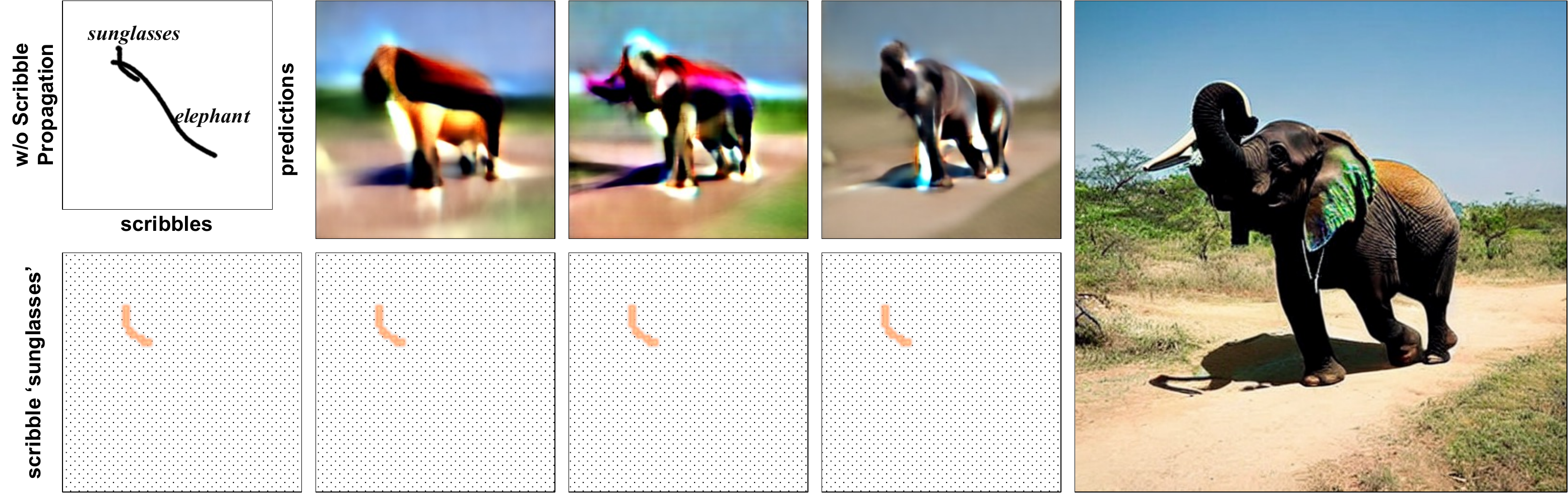}
    \parbox{1.00 \linewidth}{
        \centering
        \small
        {result using \textbf{thick scribbles}, without scribble propagation}
    }
    \includegraphics[width=1.00\linewidth]{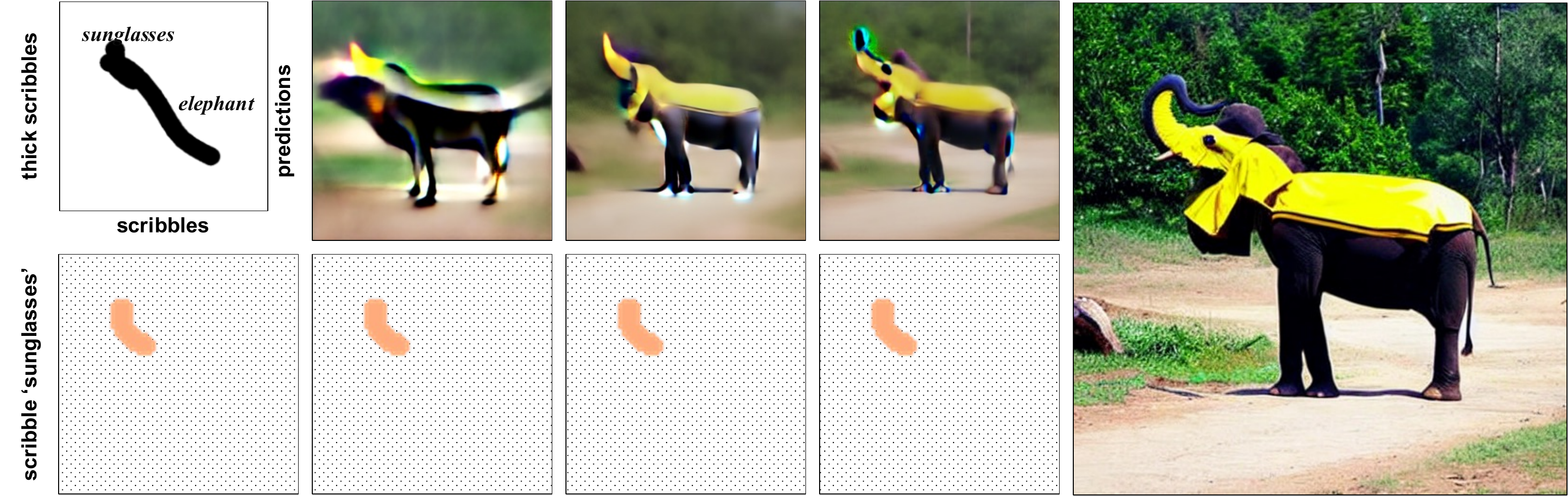}
    \parbox{1.00 \linewidth}{
        \centering
        \small
        {result with scribble propagation}
    }
    \includegraphics[width=1.00\linewidth]{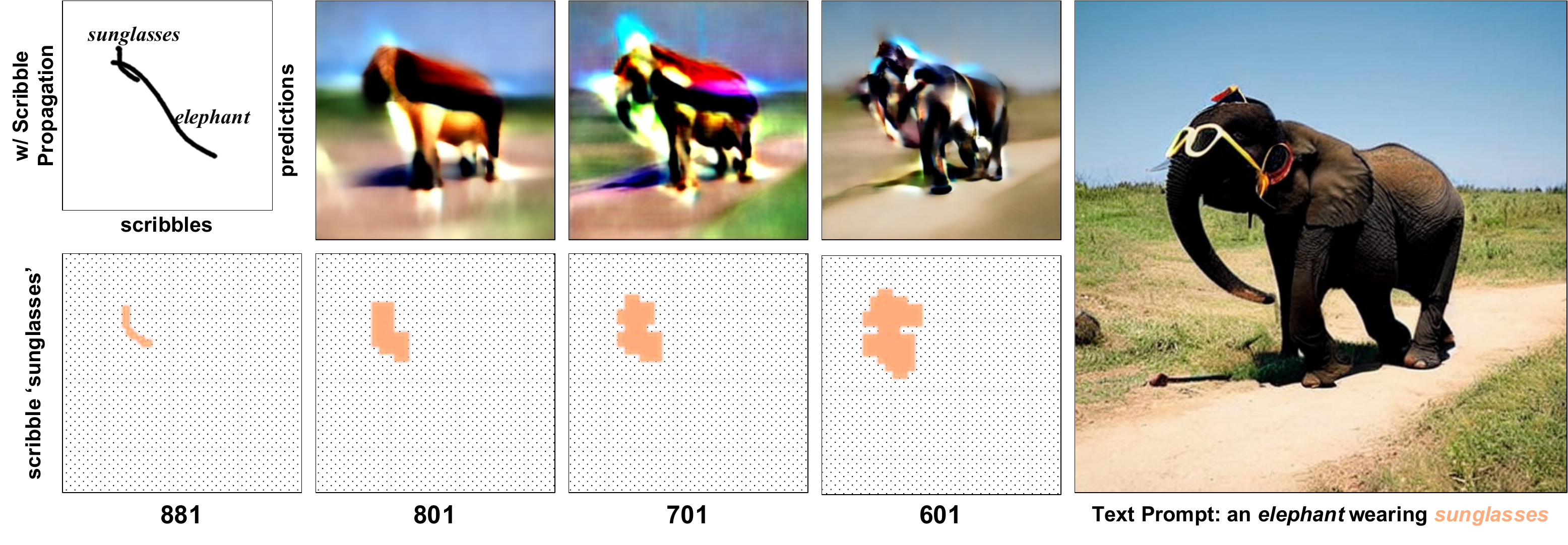}
    \caption{
        \textbf{
            Additional Ablation of Scribble Propagation and the comparison with only using \textit{thick scribbles}.
        }
        Without scribble propagation, the generated object \textit{\textbf{``sunglasses"}} is not properly captured due to the thin nature of the input scribble, leading to incomplete and incorrect object generation. By applying scribble propagation, our method extends the input scribble over time, ensuring that finer details such as the \textit{\textbf{``sunglasses"}} are captured and aligned with the text prompt.
        (Text Prompt: \textit{an \textbf{elephant} wearing \textbf{sunglasses}})
        }
    \label{fig:appendix_ablation_scribble_propagation}
\end{figure}

\noindent
\textbf{Different Propagation Methods.}
Naively applying techniques like Gaussian kernel~\cite{babaud1986uniqueness_Gaussian_kernel} or dilation to intentionally thicken scribbles is suboptimal or ineffective.
Thickening the lines can distort the abstract shape that the user intended to express, as the expanded lines may blur or dilute the original form.
This is particularly problematic for objects with fine details, as certain parts of the object should be expanded while others, such as thin features like an elephant’s trunk, should remain unblurred to preserve accuracy.
An example of this issue is illustrated in \cref{fig:appendix_ablation_scribble_propagation} (second row), where despite thickening the scribble by 16 times from the start, the resulting image lacks key features like sunglasses, leading to an unnatural outcome without proper scribble propagation.

\section{Implementation Details}
\label{sec:supple_imple_detail}


In our implementation, several hyperparameters were chosen to balance the effectiveness and efficiency of the proposed method.
For the scribble propagation, we set the merging threshold $\tau$ to 0.001 to effectively merge anchors near the boundary of a scribble without over-expanding into irrelevant regions.
The number of top-$k$ tokens for token selection was fixed at 20, providing a sufficient range for propagating the scribble to neighboring areas. The scribble propagation starts at timestep $k_1 = 5$ and ends at timestep $k_2 = 15$ within the reverse diffusion process, ensuring that the model has ample time to incorporate the scribble information early in the denoising steps while maintaining computational efficiency.

For self-attention map aggregation, we utilized multiple resolutions, specifically [8, 16, 32, 64], to capture attention from various scales and downsampled the aggregated self-attention maps to a resolution of 64. This multi-resolution approach allowed us to better capture fine-grained spatial information while maintaining computational feasibility. 

The moment alignment process was guided by two terms: $\lambda_1$, which controls the contribution of the centroid moment loss, and $\lambda_2$, which regulates the central moment loss. We empirically set both $\lambda_1$ and $\lambda_2$ to 0.6, which provided a good balance between aligning the position and the orientation of the generated object with the scribble prompt.

Additionally, to ensure balanced optimization, the loss terms were weighted with a ratio of 5:3 for the cross-attention focal loss ($\mathcal{L}_{\texttt{focal}}$) and the moment loss ($\mathcal{L}_{\texttt{moment}}$), respectively. This weighting reflects the relative importance of ensuring precise alignment between the generated image and the scribble in terms of both spatial placement and orientation. Furthermore, we set $\beta$ in \cref{eq:cross_attention_focal_loss} as 2.0. Finally, the anchor grid size was set to $16 \times 16$ with each anchor representing a $2 \times 2$ token cluster, which provided sufficient granularity for the scribble propagation process without causing unnecessary computational overhead.


\section{Overall Algorithm}
\label{sec:supple_overall_algorithm}

The overall workflow of our method, \MODELNAME\., involves iterative guidance during the reverse diffusion process using two main components: \textbf{Cross-Attention Control with Moment Alignment} and \textbf{Scribble Propagation}. 

At each timestep in the reverse diffusion process, the latent code is adjusted based on the focal loss and moment alignment, ensuring that the generated object reflects both the spatial alignment and orientation of the scribble input. The scribble propagation process occurs within a specified interval of timesteps ($k_1$ to $k_2$) and involves iteratively expanding the scribble regions. Notably, the merging of scribble regions is guided by a distance metric similar to Dijkstra's algorithm, where anchors near the boundary of a scribble are evaluated based on Kullback-Leibler divergence. The algorithm selects the $k$ closest anchors, gradually extending the scribble regions. This approach is akin to a shortest-path search, where regions with the smallest divergence are progressively included in the scribble.
For further details on the algorithm, see \cref{alg:overall_algorithm}.

\begin{algorithm}[!h]
\caption{Scribble-Guided Diffusion}\label{alg:overall_algorithm}

\textbf{Input:} A diffusion model $\boldsymbol{\epsilon}_{\theta}$ with parameters $\theta$, a latent code $z_T$ on timestep $T$, a scribble $s \in \left\{0, 1 \right\}^{H \times W}$, and a scribble region $\mathcal{S}$ corresponding to scribble $s$.\\
\textbf{Hyperparameters:} Timestep interval for scribble propagation $[k_1, k_2]$, weights for moment losses $\lambda_1$ and $\lambda_2$, resolution list for self-attention map aggregation \texttt{res}, and aggregation weights $\omega_{i}$ for each resolution level $i$.\\
\textbf{Output:} $z_0$.

\begin{algorithmic}[1]
\For{$t = T, T-1, \dots, 1$}
    \State Calculate $\hat{z}_{t - 1}$ by \cref{eq:ddim_reverse} 

    \State
    \State \textcolor{red}{\# Cross Attention and Moment Loss (\cref{sec:Guidance_for_Moment_Alignment})}

    \State \textcolor{blue}{\# Calculate cross attention loss}
        
    \State Calculate $\mathcal{L}_{\texttt{focal}}$ by \cref{eq:cross_attention_focal_loss} using $\forall c \in \mathcal{C} (s)$
    \State Calculate $\mathcal{L}_{\texttt{centroid}}$ by \cref{eq:first_order_image_moment} using $\forall c \in \mathcal{C} (s)$
    \State Calculate $\mathcal{L}_{\texttt{central}}$ by \cref{eq:second_order_image_moment} using $\forall c \in \mathcal{C} (s)$
    
    \State $\mathcal{L}_{\texttt{moment}} = \lambda_1 \mathcal{L}_{\texttt{centroid}}+ \lambda_2 \mathcal{L}_{\texttt{central}}$
    \State $\mathcal{L}_{\texttt{cross}} = \mathcal{L}_{\texttt{focal}} + \mathcal{L}_{\texttt{moment}}$

    \State \textcolor{blue}{\# Shift latent code}
    \State $z_{t-1} \gets \hat{z}_{t-1} - \nabla_{z_{t}} \mathcal{L}_{\texttt{cross}}$

    \State
    \State \textcolor{red}{\# Scribble Propagation (\cref{sec:Scribble_Propagation})}

    \If{not ($k_1 \le t \le k_2$)}
        \State continue
    \EndIf



    \State \textcolor{blue}{\# Aggregate self-attention maps}  (as DiffSeg~\cite{shuai2024diffseg})

    \For{$i, (H, W)$ in $\texttt{res}$}
        \State $\delta \gets H^{\texttt{agg}} / H$
        \State $\mathcal{A}^{\texttt{new}} \gets \textbf{Resize}\left(\mathcal{A}_{\texttt{self}}^{H \times W}, H^{\texttt{agg}} \times W^{\texttt{agg}} \right)$
        \For{each patch $(h, w)$ in $\mathcal{A}^{\texttt{agg}}$}
            \State $\mathcal{A}^{\texttt{agg}}[h, w] \mathrel{+}= \omega_{i} \cdot \mathcal{A}^{\texttt{new}}\left[h // \delta, w // \delta \right]$
        \EndFor
    \EndFor

    \State $\delta_{\texttt{anc}} \gets H^{\texttt{agg}} / H^{\texttt{anchor}}$
    

    \State \textcolor{blue}{\# Region-avg pool aggregated self-attention maps}

    \State $\mathcal{A}^{\texttt{anc}} \gets \textbf{AvgPool}\left(\mathcal{A}^{\texttt{agg}}, \delta_{\texttt{anc}} \times \delta_{\texttt{anc}}\right)$

    \For{each object $o$}
        \State $\mathcal{A}^{\texttt{scr}}\left[o\right] \gets \frac{1}{\mathcal{S}_{o}} \sum_{(i,j) \in \mathcal{S}_{o}} \mathcal{A}^{\texttt{anc}}[i, j]$
    \EndFor


    \State $\texttt{MergeNeighbors}$($s, \mathcal{S}, \mathcal{B}^{s}$)


    \EndFor
    \end{algorithmic}
\end{algorithm}

\begin{algorithm}[!h]
\caption{\texttt{MergeNeighbors}()}\label{alg:merge_neighbors}
\textbf{Input:} a scribble $s$, a scribble region $\mathcal{S}$ of $s$, boundary anchors $\mathcal{B}^{s}$ of a scribble $s$.\\
\textbf{Hyperparameters:} Distance threshold $\tau_{\texttt{dist}}$ for merging, number of neighbors $k$.\\
\begin{algorithmic}[1]
    \State Initialize $\textbf{dist}_{\texttt{nbr}}$ and $\textbf{obj}_{\texttt{nbr}}$  to $\infty$ and 0 respectively
    \For{each object $o$ and edge $(i,j)$ in $\mathcal{B}^{s}$}
        \State Find neighbors $\mathcal{N}(i, j)$
        \For{each neighbor $(n_i, n_j) \in \mathcal{N}(i, j)$}
            \If{neighbor is visited}
                \State continue
            \EndIf
            \State Calculate distance $d$ using \cref{eq:distance_D}
            \State \textcolor{blue}{\# Select candidates}
            \State \textcolor{blue}{\# which distances are lower than threshold}
            \If{$d < \tau_{\texttt{dist}}$}
                \State $\textbf{dist}_{\texttt{nbr}}[n_i, n_j] \gets d$
                \State $\textbf{obj}_{\texttt{nbr}}[n_i, n_j] \gets o$
            \EndIf
        \EndFor
    \EndFor

    \State \textcolor{blue}{\# Select neighbors with K-highest similarities}
    \State $\textbf{indices}_{\texttt{nbr}} \gets \textbf{TopK}(\textbf{dist}_{\texttt{nbr}}, k)$

    \State \textcolor{blue}{\# Integrate selected neighbors into scribble}
    \For{$(n_i, n_j)$ in $\textbf{indices}_{\texttt{nbr}}$}
        \State $S[\textbf{obj}_{\texttt{nbr}}[ \text{idx}] - 1, n_i, n_j] \gets \textbf{True}$
    \EndFor
\end{algorithmic}
\end{algorithm}


\newpage
\quad
\newpage
\quad
\newpage

\begin{figure*}[ht]
    \centering

    \parbox{1.00 \textwidth}{
        \centering
        \small
        \textit{
            a \textcolor{lightgray}{\textbf{dog}} and a \textcolor{Tan}{\textbf{horse}}
        }
    }
    \begin{subfigure}[t]{0.19\textwidth} \includegraphics[width=\textwidth]{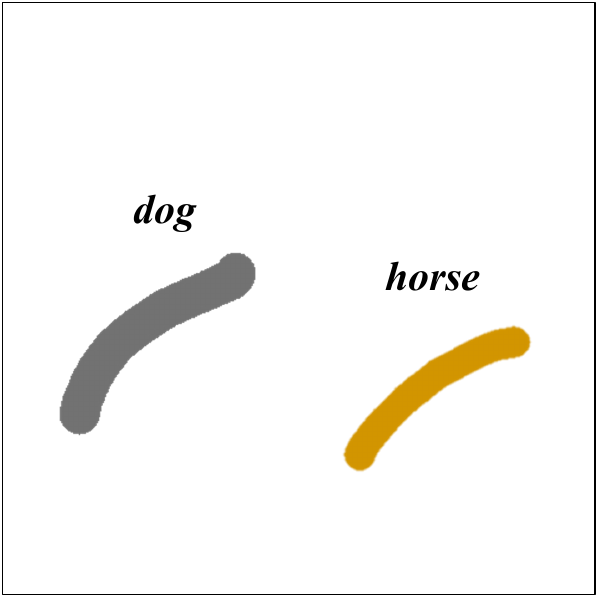} \end{subfigure}
    \begin{subfigure}[t]{0.19\textwidth} \includegraphics[width=\textwidth]{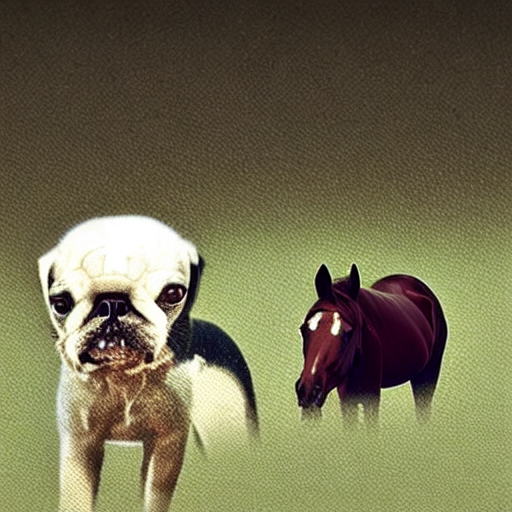} \end{subfigure}
    \begin{subfigure}[t]{0.19\textwidth} \includegraphics[width=\textwidth]{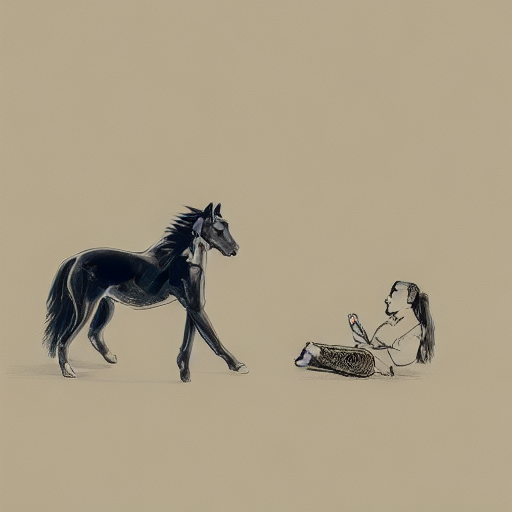} \end{subfigure}
    \begin{subfigure}[t]{0.19\textwidth} \includegraphics[width=\textwidth]{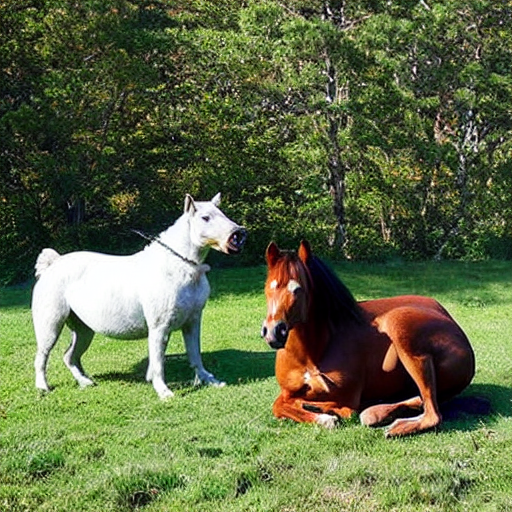} \end{subfigure}
    \begin{subfigure}[t]{0.19\textwidth} \includegraphics[width=\textwidth]{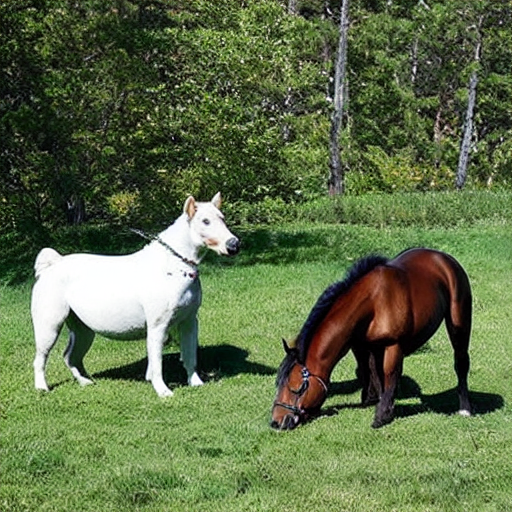} \end{subfigure}

    \vspace{-0.25em}
    
    \parbox{1.00 \textwidth}{
        \centering
        \small
        \textit{
            A \textcolor{Tan}{\textbf{tree}} with a few \textcolor{Bittersweet}{\textbf{birds}} sitting on its \textcolor{ForestGreen}{\textbf{branches}}, while the sun sets in the background.
        }
    }
    \begin{subfigure}[t]{0.19\textwidth} \includegraphics[width=\textwidth]{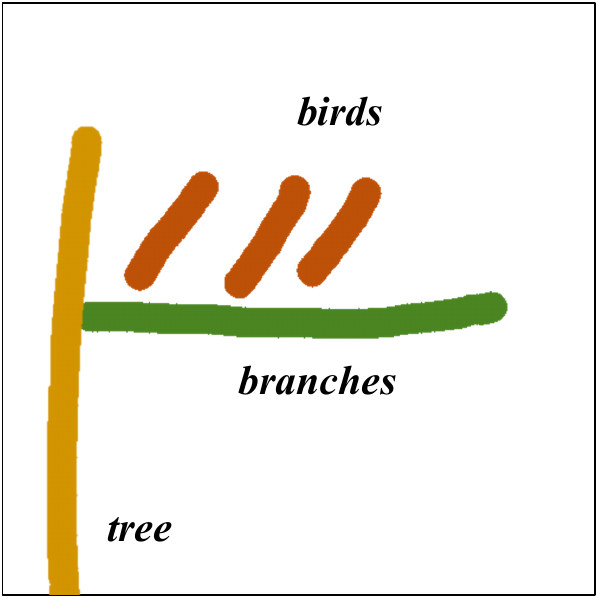} \end{subfigure}
    \begin{subfigure}[t]{0.19\textwidth} \includegraphics[width=\textwidth]{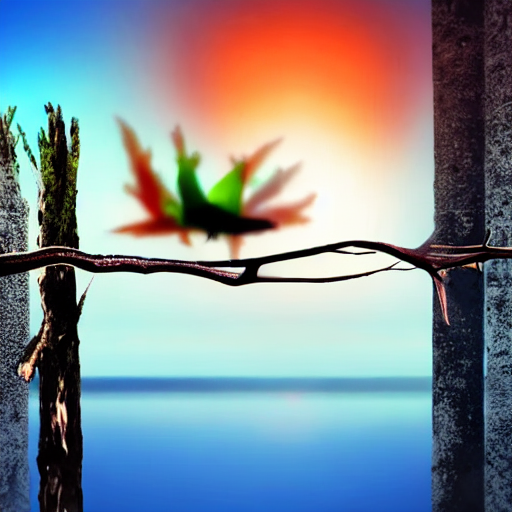} \end{subfigure}
    \begin{subfigure}[t]{0.19\textwidth} \includegraphics[width=\textwidth]{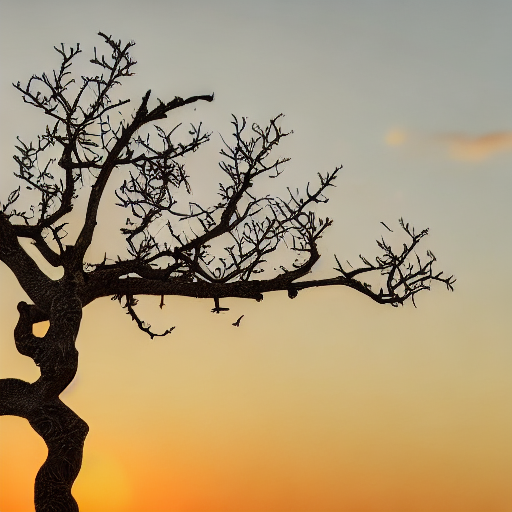} \end{subfigure}
    \begin{subfigure}[t]{0.19\textwidth} \includegraphics[width=\textwidth]{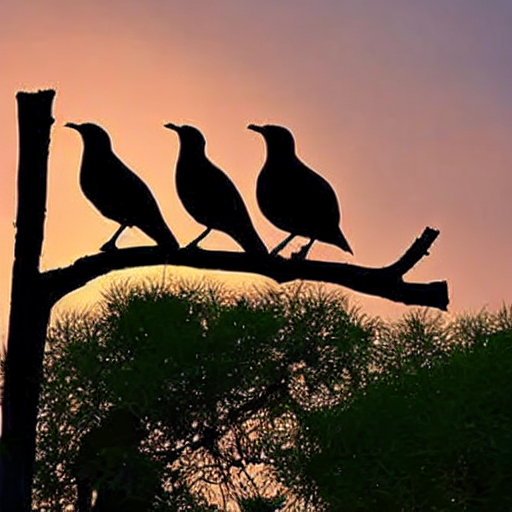} \end{subfigure}
    \begin{subfigure}[t]{0.19\textwidth} \includegraphics[width=\textwidth]{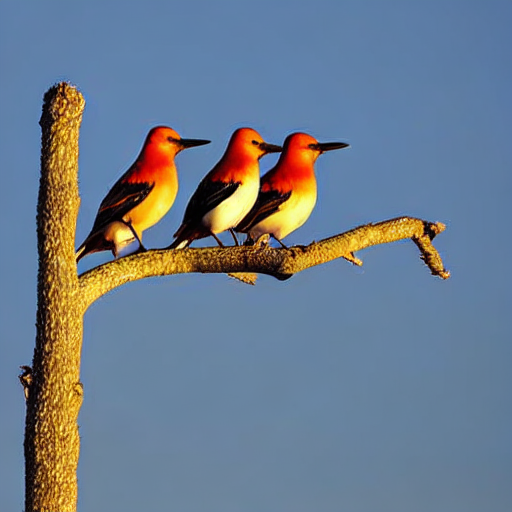} \end{subfigure}

    \vspace{-0.25em}
    
    \parbox{1.00 \textwidth}{
        \centering
        \small
        \textit{
            A \textcolor{teal}{\textbf{mermaid}} sitting on a \textcolor{teal}{\textbf{rock}} by the \textcolor{blue}{\textbf{ocean}}, with a full \textcolor{yellow}{\textbf{moon}} and stars in the sky, gentle waves, ethereal light, peaceful, calming
        }
    }
    \begin{subfigure}[t]{0.19\textwidth} \includegraphics[width=\textwidth]{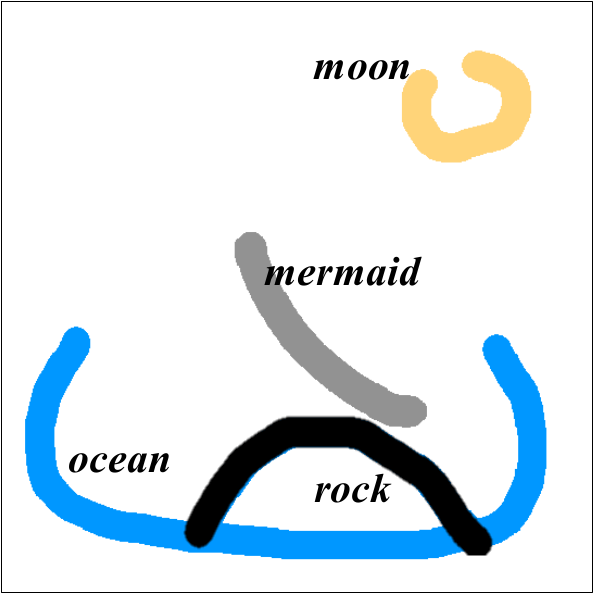} \end{subfigure}
    \begin{subfigure}[t]{0.19\textwidth} \includegraphics[width=\textwidth]{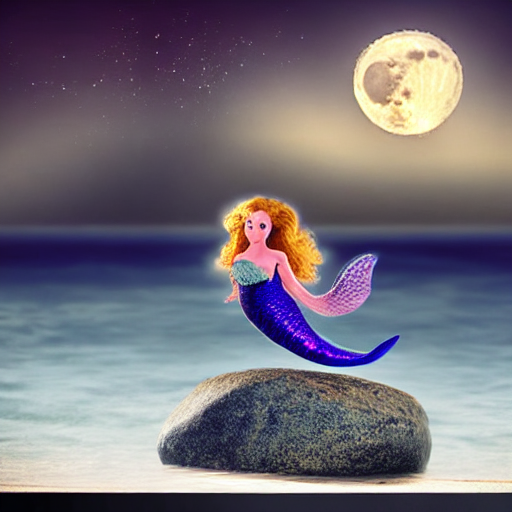} \end{subfigure}
    \begin{subfigure}[t]{0.19\textwidth} \includegraphics[width=\textwidth]{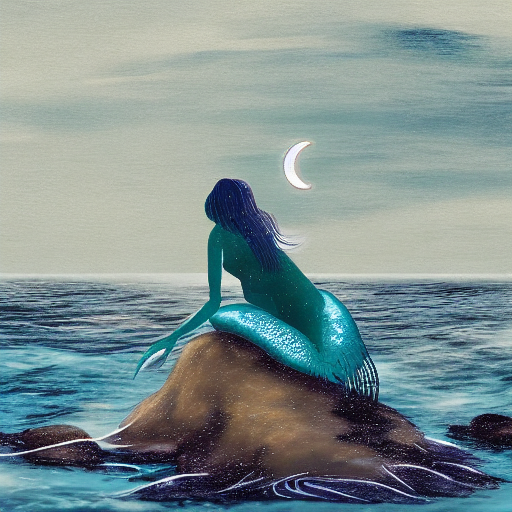} \end{subfigure}
    \begin{subfigure}[t]{0.19\textwidth} \includegraphics[width=\textwidth]{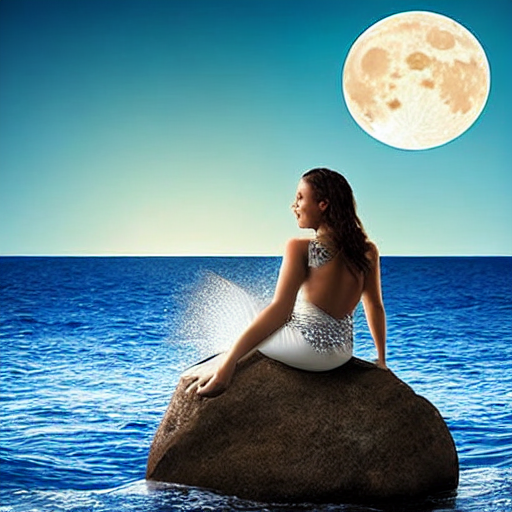} \end{subfigure}
    \begin{subfigure}[t]{0.19\textwidth} \includegraphics[width=\textwidth]{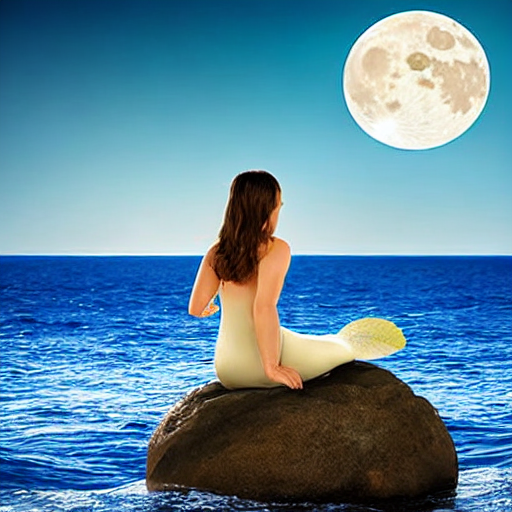} \end{subfigure}

    \vspace{-0.25em}
    \parbox{1.00 \textwidth}{
        \centering
        \small
        \textit{
            A \textcolor{RoyalBlue}{\textbf{car}} driving down a \textcolor{darkgray}{\textbf{winding road}} through the hills, with trees lining the path and \textcolor{lightgray}{\textbf{clouds}} above.
        }
    }
    \begin{subfigure}[t]{0.19\textwidth} \includegraphics[width=\textwidth]{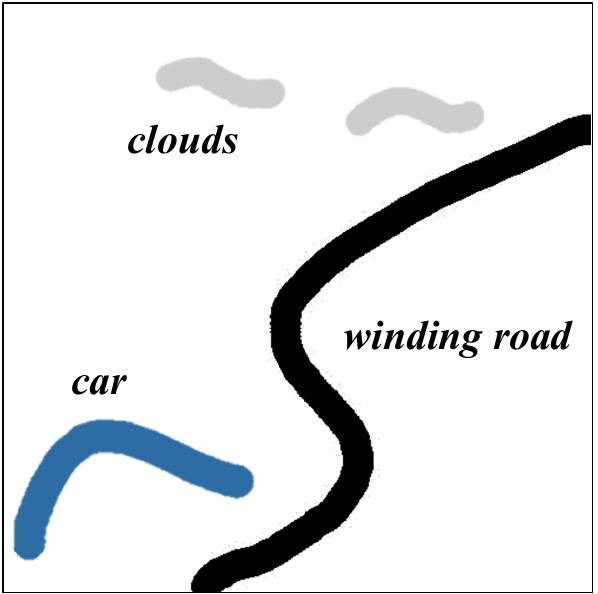} \end{subfigure}
    \begin{subfigure}[t]{0.19\textwidth} \includegraphics[width=\textwidth]{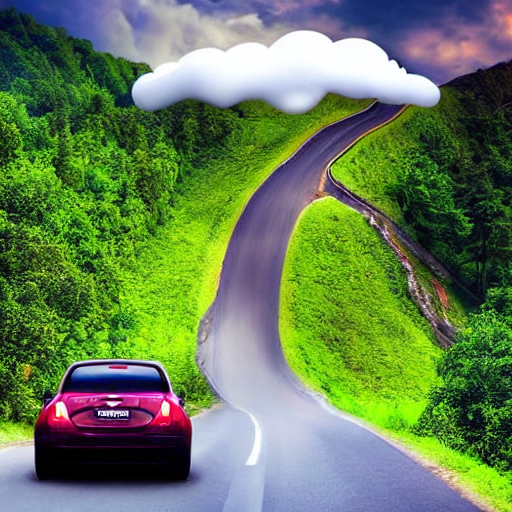} \end{subfigure}
    \begin{subfigure}[t]{0.19\textwidth} \includegraphics[width=\textwidth]{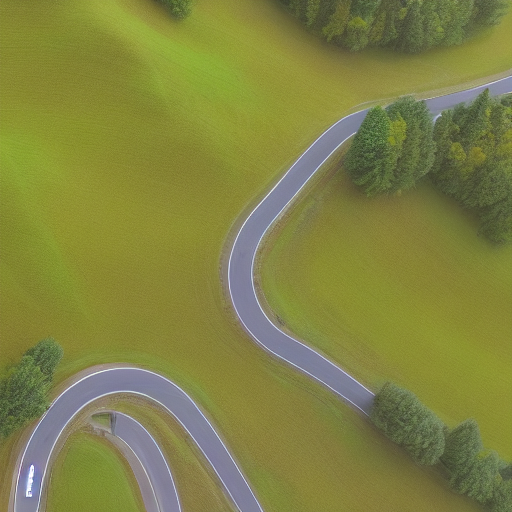} \end{subfigure}
    \begin{subfigure}[t]{0.19\textwidth} \includegraphics[width=\textwidth]{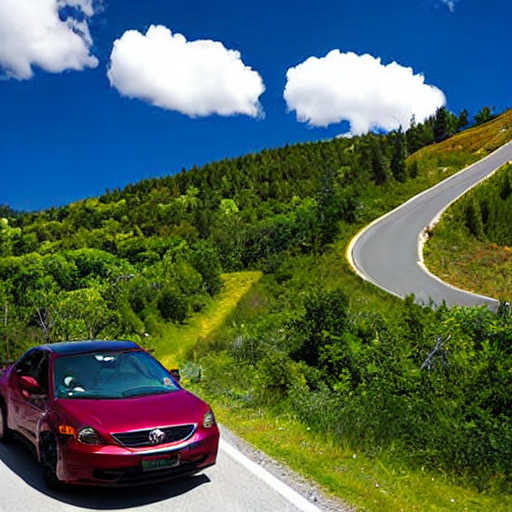} \end{subfigure}
    \begin{subfigure}[t]{0.19\textwidth} \includegraphics[width=\textwidth]{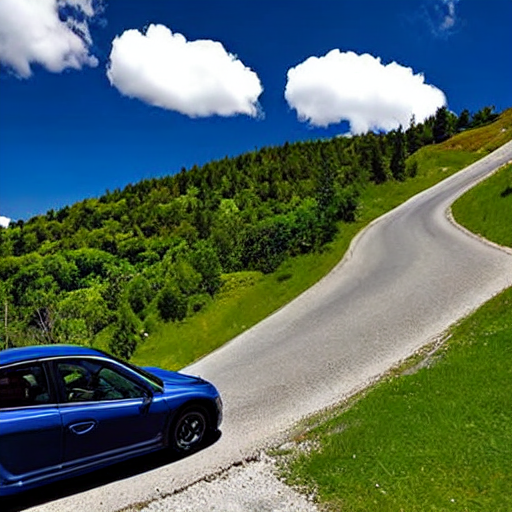} \end{subfigure}

    \vspace{-0.25em}
    \parbox{1.00 \textwidth}{
        \centering
        \small
        \textit{
            \textcolor{darkgray}{\textbf{Cute panda}} peacefully drifting on a \textcolor{brown}{bamboo raft} down a serene \textcolor{RoyalBlue}{\textbf{river}} in a lush \textcolor{ForestGreen}{\textbf{bamboo forest}}, detailed digital painting
        }
    }
    \begin{subfigure}[t]{0.19\textwidth}
        \includegraphics[width=\textwidth]{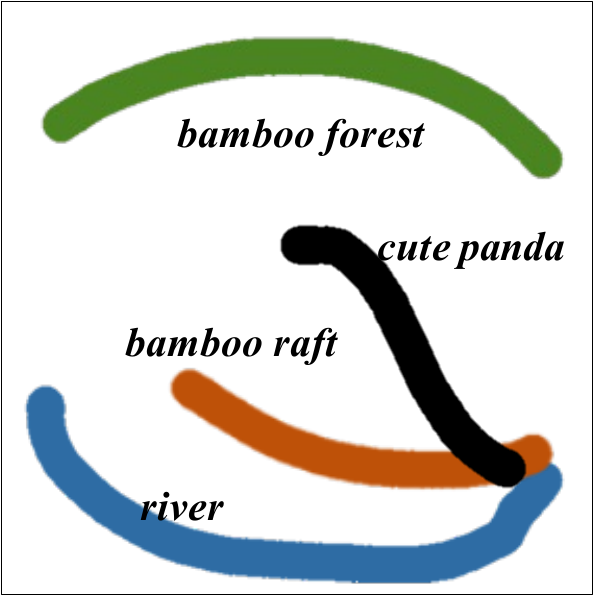}
        \caption{Scribbles}
    \end{subfigure}
    \begin{subfigure}[t]{0.19\textwidth}
        \includegraphics[width=\textwidth]{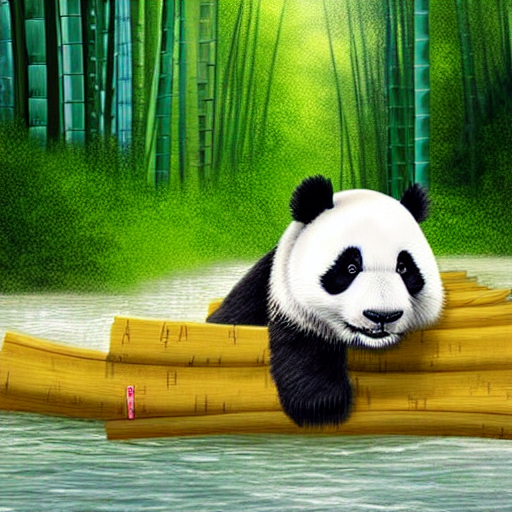}
        \caption{BoxDiff~\cite{xie2023boxdiff}}
    \end{subfigure}
    \begin{subfigure}[t]{0.19\textwidth}
        \includegraphics[width=\textwidth]{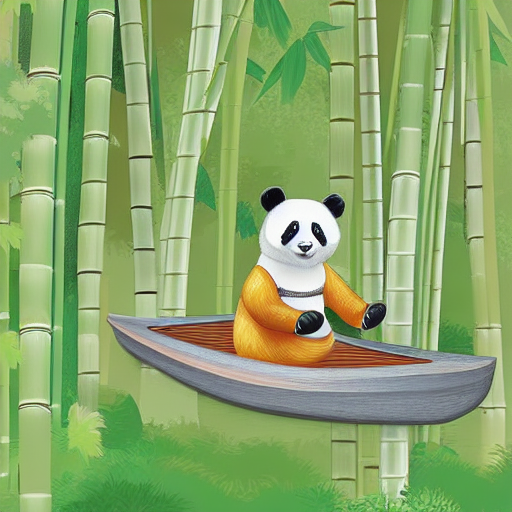}
        \caption{DenseDiff~\cite{densediffusion}}
    \end{subfigure}
    \begin{subfigure}[t]{0.19\textwidth}
        \includegraphics[width=\textwidth]{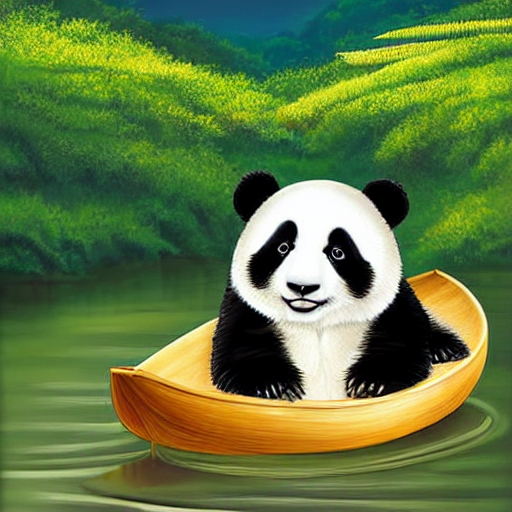}
        \caption{GLIGEN~\cite{li2023gligen}}
    \end{subfigure}
    \begin{subfigure}[t]{0.19\textwidth}
        \includegraphics[width=\textwidth]{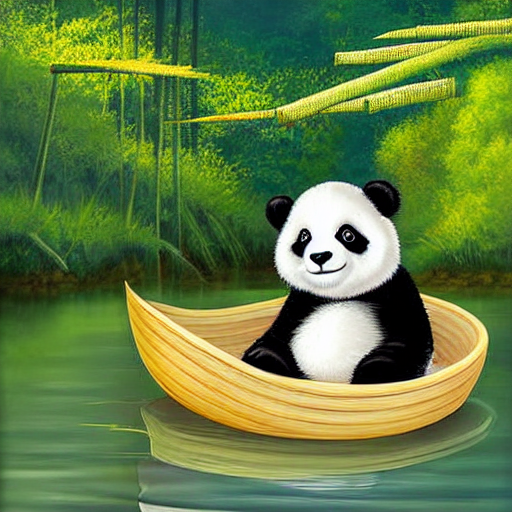}
        \caption{\MODELNAME \. (Ours)}
    \end{subfigure}
    
    \caption{
        \textbf{Additional qualitative comparison of Text-to-Image generation methods using scribble prompts.} \MODELNAME \. yields outcomes that better reflect the scribble inputs, especially concerning the accuracy of object orientations and abstract shape representation.
    }
    \label{fig:qualitative_supplementary}
    \vspace{-1.0em}
\end{figure*}

\begin{figure*}[ht]
    \centering
    \parbox{1.00 \textwidth}{
        \centering
        \small
        \textit{
            A photo of an \textcolor{PineGreen}{\textbf{horse}}
        }
    }
    \begin{subfigure}[t]{0.16\textwidth} \includegraphics[width=\textwidth]{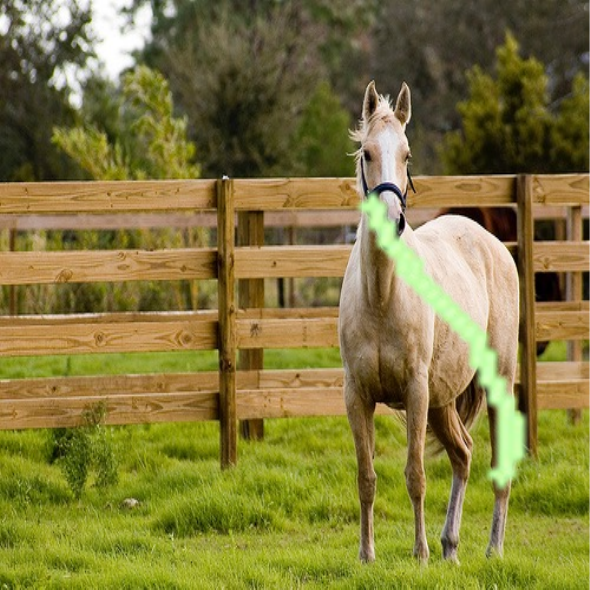} 
    \end{subfigure}
    \begin{subfigure}[t]{0.16\textwidth} \includegraphics[width=\textwidth]{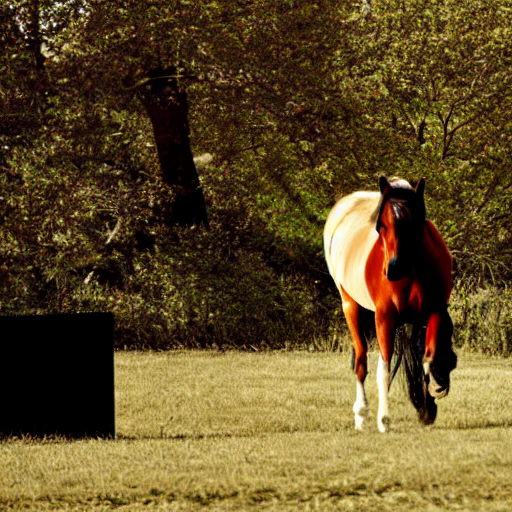} 
    \end{subfigure}
    \begin{subfigure}[t]{0.16\textwidth} \includegraphics[width=\textwidth]{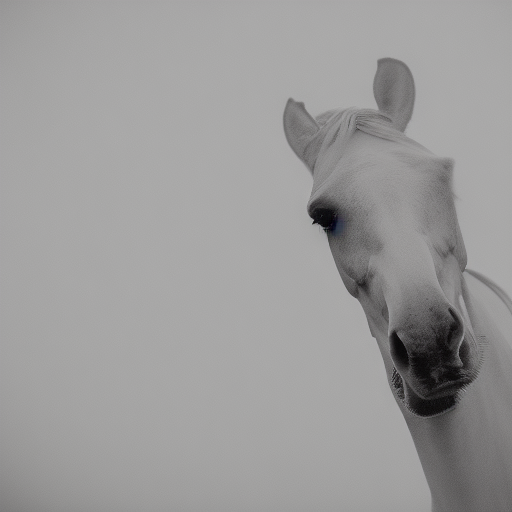}
    \end{subfigure}
    \begin{subfigure}[t]{0.16\textwidth} \includegraphics[width=\textwidth]{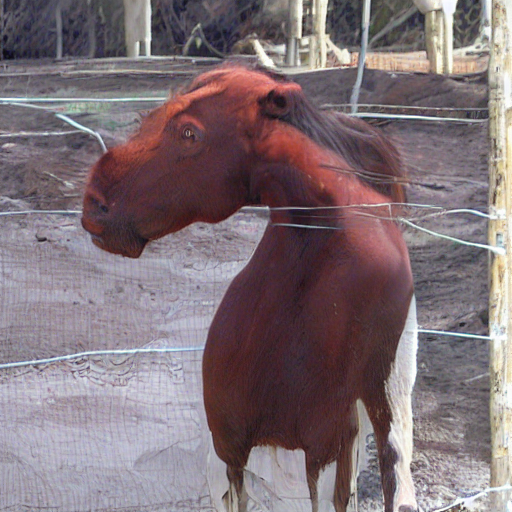} 
    \end{subfigure}
    \begin{subfigure}[t]{0.16\textwidth} \includegraphics[width=\textwidth]{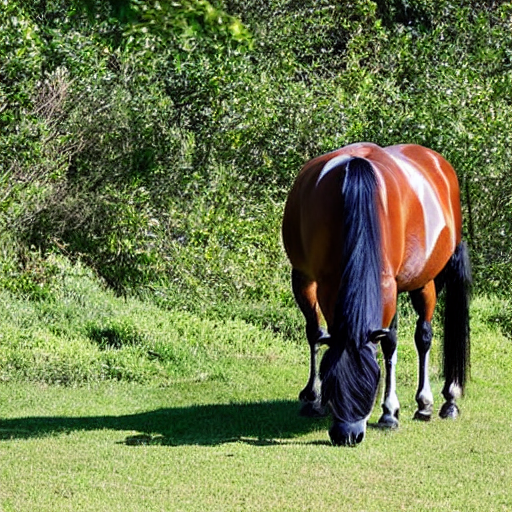} 
    \end{subfigure}
    \begin{subfigure}[t]{0.16\textwidth} \includegraphics[width=\textwidth]{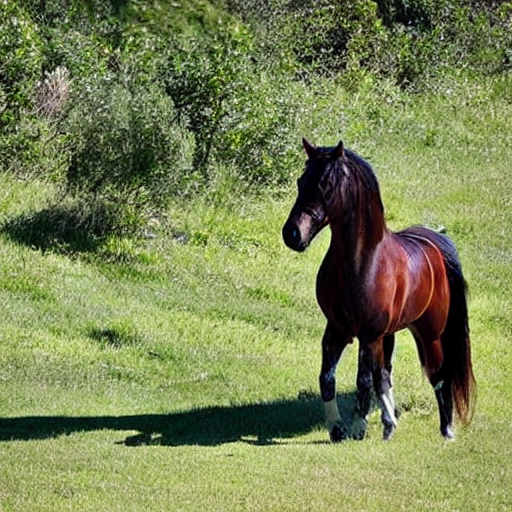} 
    \end{subfigure}

    \parbox{1.00 \textwidth}{
        \centering
        \small
        \textit{
            A photo of a \textcolor{PineGreen}{\textbf{person}}
        }
    }
    \begin{subfigure}[t]{0.16\textwidth} \includegraphics[width=\textwidth]{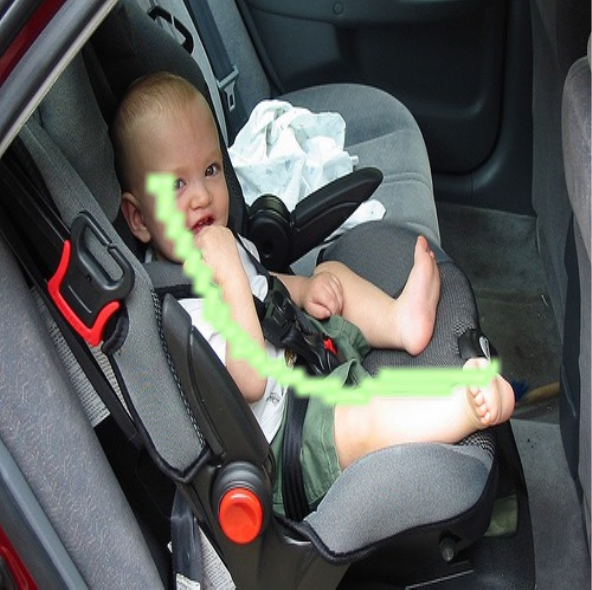} 
    \end{subfigure}
    \begin{subfigure}[t]{0.16\textwidth} \includegraphics[width=\textwidth]{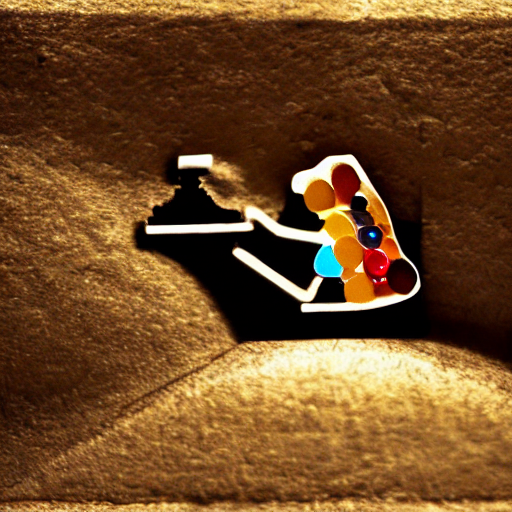} 
    \end{subfigure}
    \begin{subfigure}[t]{0.16\textwidth} \includegraphics[width=\textwidth]{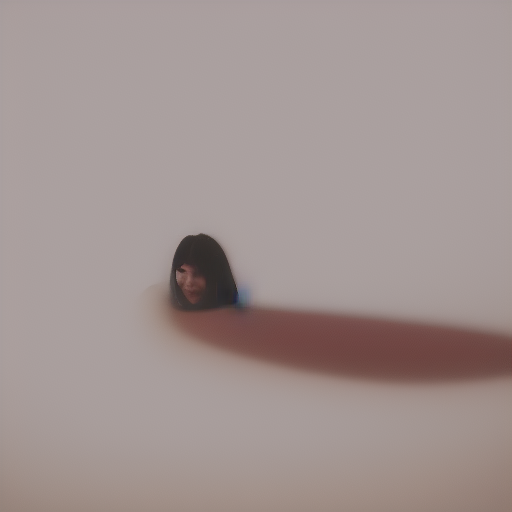}
    \end{subfigure}
    \begin{subfigure}[t]{0.16\textwidth} \includegraphics[width=\textwidth]{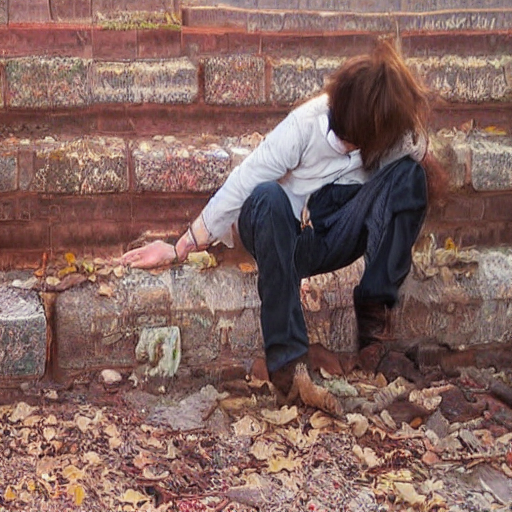} 
    \end{subfigure}
    \begin{subfigure}[t]{0.16\textwidth} \includegraphics[width=\textwidth]{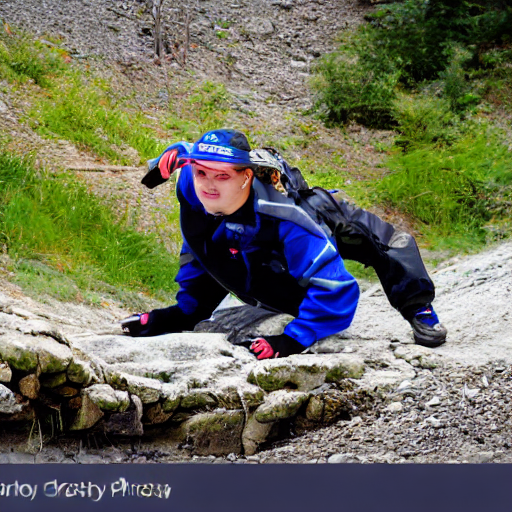} 
    \end{subfigure}
    \begin{subfigure}[t]{0.16\textwidth} \includegraphics[width=\textwidth]{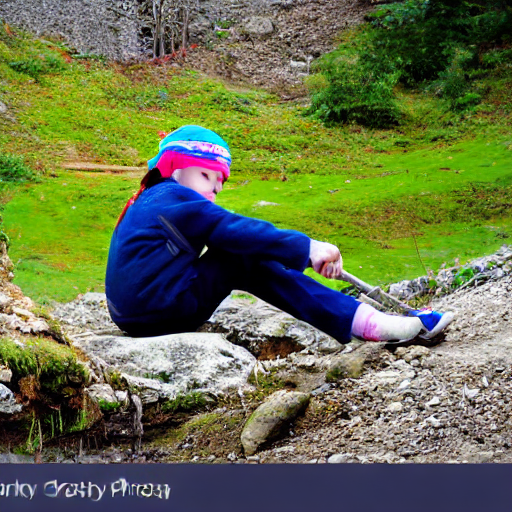} 
    \end{subfigure}

    \parbox{1.00 \textwidth}{
        \centering
        \small
        \textit{
            A photo of a \textcolor{PineGreen}{\textbf{cat}}
        }
    }
    \begin{subfigure}[t]{0.16\textwidth} \includegraphics[width=\textwidth]{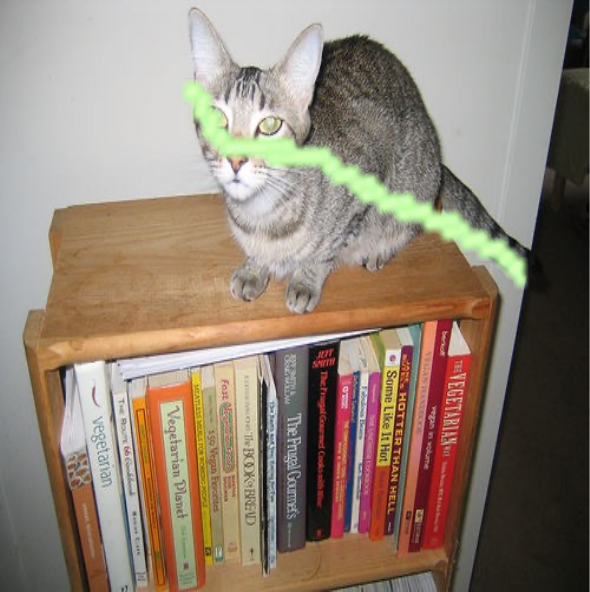} 
    \end{subfigure}
    \begin{subfigure}[t]{0.16\textwidth} \includegraphics[width=\textwidth]{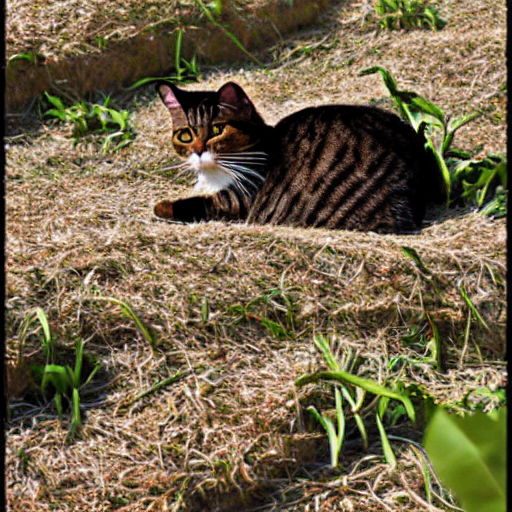} 
    \end{subfigure}
    \begin{subfigure}[t]{0.16\textwidth} \includegraphics[width=\textwidth]{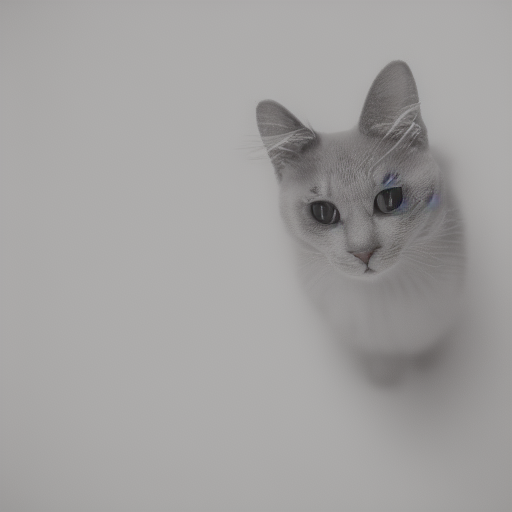}
    \end{subfigure}
    \begin{subfigure}[t]{0.16\textwidth} \includegraphics[width=\textwidth]{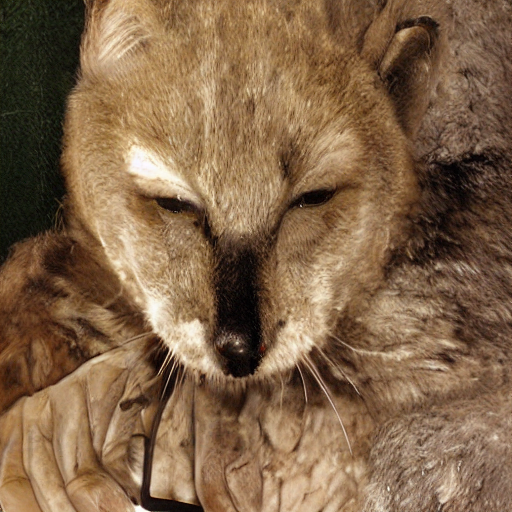} 
    \end{subfigure}
    \begin{subfigure}[t]{0.16\textwidth} \includegraphics[width=\textwidth]{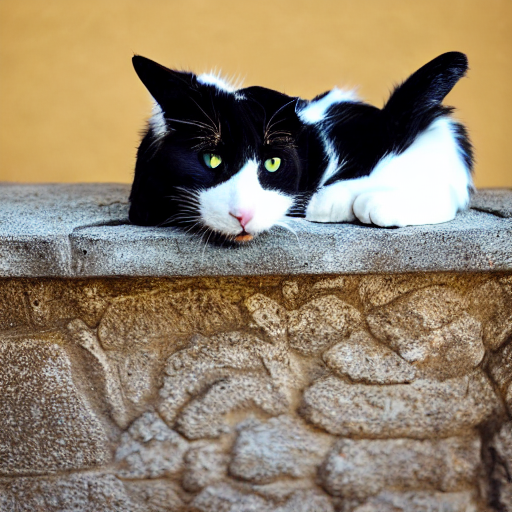} 
    \end{subfigure}
    \begin{subfigure}[t]{0.16\textwidth} \includegraphics[width=\textwidth]{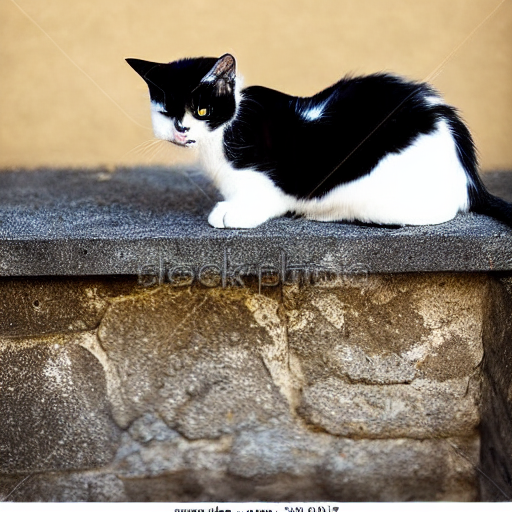} 
    \end{subfigure}

    \parbox{1.00 \textwidth}{
        \centering
        \small
        \textit{
            A photo of a \textcolor{PineGreen}{\textbf{train}}
        }
    }
    \begin{subfigure}[t]{0.16\textwidth} \includegraphics[width=\textwidth]{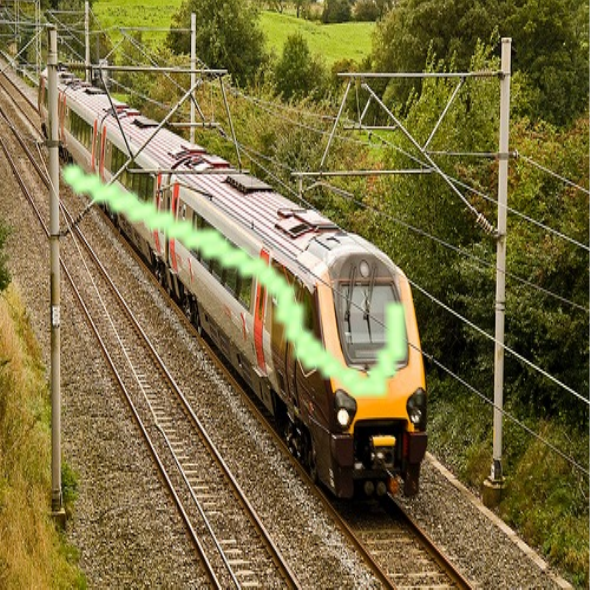} 
    \caption{Scribbles~\cite{lin2016scribblesup_data}}
    \end{subfigure}
    \begin{subfigure}[t]{0.16\textwidth} \includegraphics[width=\textwidth]{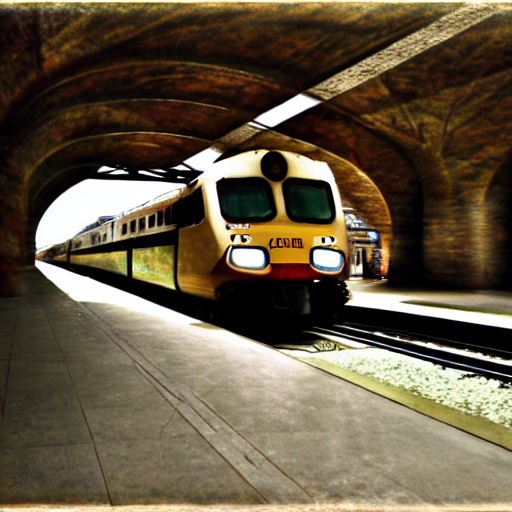} 
    \caption{BoxDiff~\cite{xie2023boxdiff}}
    \end{subfigure}
    \begin{subfigure}[t]{0.16\textwidth} \includegraphics[width=\textwidth]{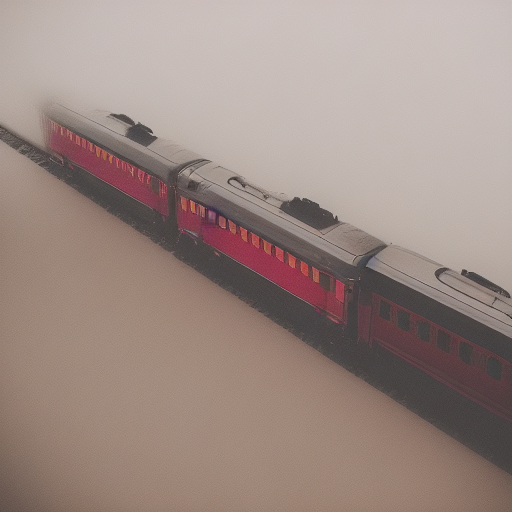}
    \caption{DenseDiff~\cite{densediffusion}}
    \end{subfigure}
    \begin{subfigure}[t]{0.16\textwidth} \includegraphics[width=\textwidth]{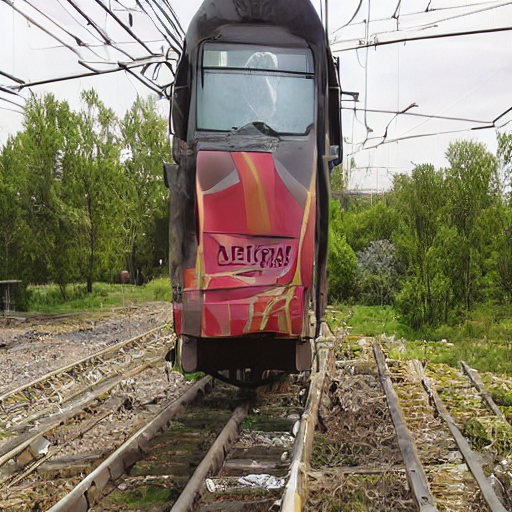} 
    \caption{ControlNet~\cite{mo2024freecontrol}}
    \end{subfigure}
    \begin{subfigure}[t]{0.16\textwidth} \includegraphics[width=\textwidth]{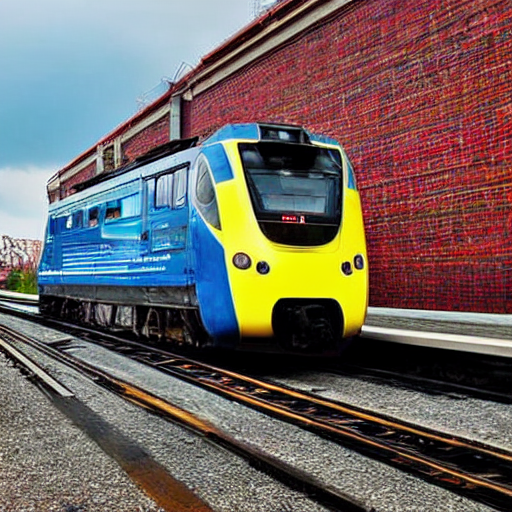} 
    \caption{GLIGEN~\cite{li2023gligen}}
    \end{subfigure}
    \begin{subfigure}[t]{0.16\textwidth} \includegraphics[width=\textwidth]{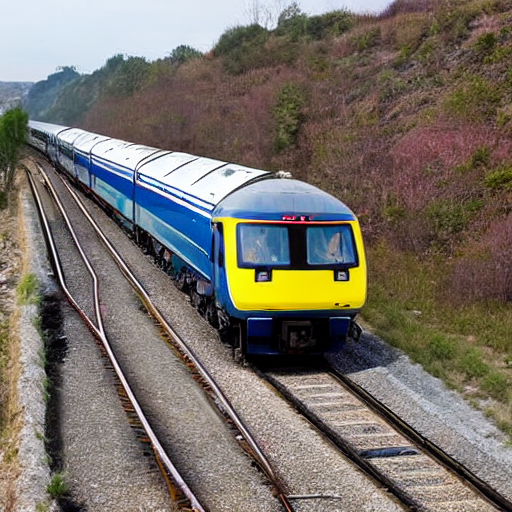} 
    \caption{\MODELNAME \. (Ours)}
    \end{subfigure}
    
    \caption{
        \textbf{Additional qualitative results on the PASCAL-Scribble dataset~\cite{lin2016scribblesup_data}.} Comparison of various Text-to-Image generation methods, including the ControlNet fine-tuned on the training dataset. 
        As shown in (f), \MODELNAME\. provides the closest representation of the original image (a) Scribbles, effectively capturing both the head direction and the standing posture.
    }
    \vspace{-1.em}
    \label{fig:appendix_qualitative_comparison_pascal}
\end{figure*}

\begin{figure}[ht]
    \centering
    \parbox{0.48\linewidth}{
        \centering
        \scriptsize{\textit{A \textbf{lion} is wearing a gold \textbf{crown}}}
    }
    \hfill
    \parbox{0.48\linewidth}{
        \centering
        \scriptsize{\textit{A \textbf{horse} drinking water at a \textbf{pond}}}
    }
    \vspace{0.5em}
    \begin{subfigure}[t]{0.24\linewidth}
        \includegraphics[width=\textwidth]{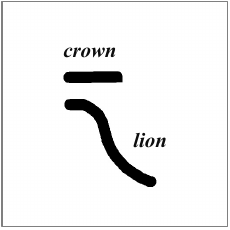}
    \end{subfigure}
    \begin{subfigure}[t]{0.24\linewidth}
        \includegraphics[width=\textwidth]{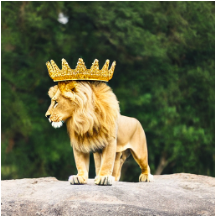}
    \end{subfigure}
    \hfill
    \begin{subfigure}[t]{0.24\linewidth}
        \includegraphics[width=\textwidth]{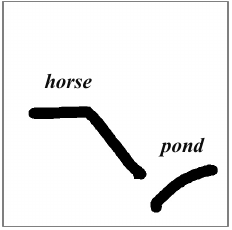}
    \end{subfigure}
    \begin{subfigure}[t]{0.24\linewidth}
        \includegraphics[width=\textwidth]{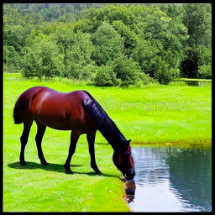}
    \end{subfigure}

    \parbox{0.48\linewidth}{
        \centering
        \scriptsize{\textit{a \textbf{cheetah} and an \textbf{elephant}}}
    }
    \hfill
    \parbox{0.48\linewidth}{
        \centering
        \scriptsize{\textit{\textbf{Three cars} parked next to \\each other in the parking lot}}
    }
    \vspace{0.5em}
    \begin{subfigure}[t]{0.24\linewidth}
        \includegraphics[width=\textwidth]{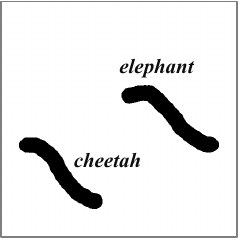}
    \end{subfigure}
    \begin{subfigure}[t]{0.24\linewidth}
        \includegraphics[width=\textwidth]{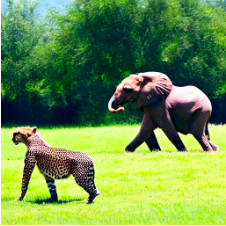}
    \end{subfigure}
    \hfill
    \begin{subfigure}[t]{0.24\linewidth}
        \includegraphics[width=\textwidth]{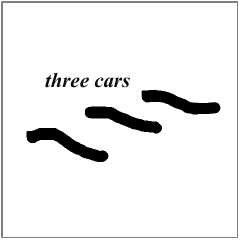}
    \end{subfigure}
    \begin{subfigure}[t]{0.24\linewidth}
        \includegraphics[width=\textwidth]{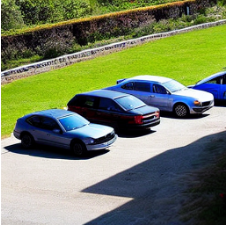}
    \end{subfigure}

    \parbox{0.48\linewidth}{
        \centering
        \scriptsize{\textit{\textbf{Rabbit} and \textbf{turtle} playing \textbf{soccer}\\
        on a \textbf{beach} by the \textbf{ocean}}}
    }
    \hfill
    \parbox{0.48\linewidth}{
        \centering
        \scriptsize{\textit{a \textbf{snake} coiled up in the \textbf{grass}}}
    }
    \vspace{0.5em}
    \begin{subfigure}[t]{0.24\linewidth}
        \includegraphics[width=\textwidth]{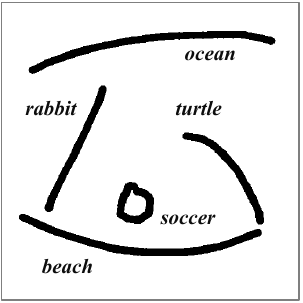}
    \end{subfigure}
    \begin{subfigure}[t]{0.24\linewidth}
        \includegraphics[width=\textwidth]{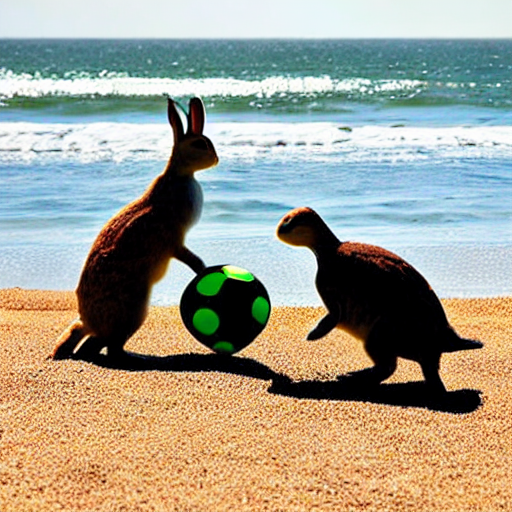}
    \end{subfigure}
    \hfill
    \begin{subfigure}[t]{0.24\linewidth}
        \includegraphics[width=\textwidth]{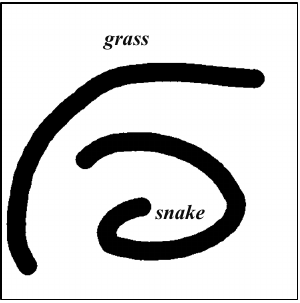}
    \end{subfigure}
    \begin{subfigure}[t]{0.24\linewidth}
        \includegraphics[width=\textwidth]{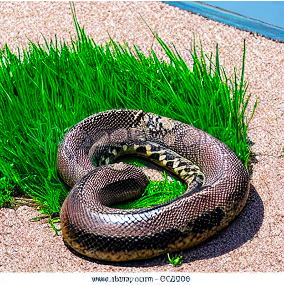}
    \end{subfigure}

    \parbox{0.48\linewidth}{
        \centering
        \scriptsize{\textit{a \textbf{horse} grazing on the \textbf{grass}\\and \textbf{three dogs} sitting on the \textbf{meadow}}}
    }
    \hfill
    \parbox{0.48\linewidth}{
        \centering
        \scriptsize{\textit{a \textbf{grizzly bear} catching \\a \textbf{salmon} in a crystal clear \textbf{river} \\surrounded by a \textbf{forest}}}
    }
    \vspace{0.5em}
    \begin{subfigure}[t]{0.24\linewidth}
        \includegraphics[width=\textwidth]{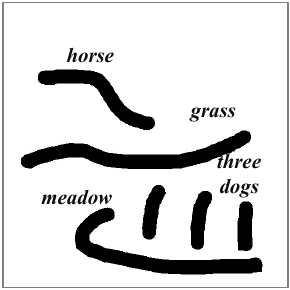}
    \end{subfigure}
    \begin{subfigure}[t]{0.24\linewidth}
        \includegraphics[width=\textwidth]{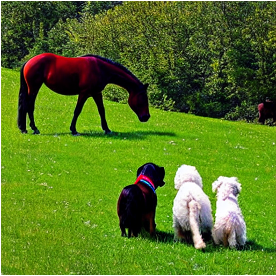}
    \end{subfigure}
    \hfill
    \begin{subfigure}[t]{0.24\linewidth}
        \includegraphics[width=\textwidth]{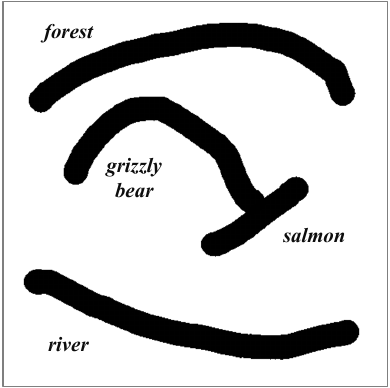}
    \end{subfigure}
    \begin{subfigure}[t]{0.24\linewidth}
        \includegraphics[width=\textwidth]{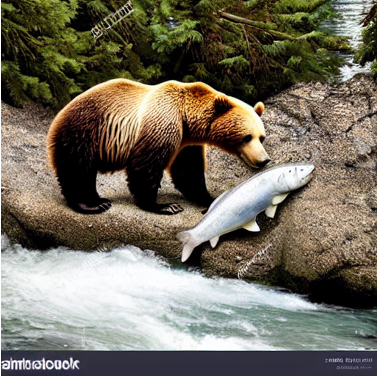}
    \end{subfigure}

    \caption{
        \textbf{Examples of Text-to-Image generation using scribble prompts by \MODELNAME\..} Each row contains two pairs of scribbles and their generated images, with the corresponding prompt placed above each pair. The layout ensures alignment and clarity for each example.
    }
    \vspace{-1em}
    \label{fig:supple_quali_sgdiff}
\end{figure}


\section{More Qualitative Results}
\label{sec:supple_more_qualitative_results}

Additional qualitative comparison results are provided alongside~\cref{fig:qualitative_comparison}.
The additional experimental results ~\cref{fig:qualitative_supplementary} show that the proposed model demonstrates better alignment with scribbles.

In ~\cref{fig:appendix_qualitative_comparison_pascal}, we offer supplementary visual comparisons between our method and other text-to-image generation methods including the fine-tuned ControlNet with scribbles. We observe that our \MODELNAME\. most accurately replicates the original image from the dataset. 

~\cref{fig:supple_quali_sgdiff} presents additional examples generated by \MODELNAME\.. The scribbles serve as a structural guide, providing the layout that the images should follow.



\section{Additional Ablation Studies}
\label{sec:supple_ablation_study}

\begin{table}[ht]
    \small
    \centering
    \resizebox{\linewidth}{!}{\begin{tabular}{@{}l c rrr@{}}
        \toprule
        $\mathcal{L}_{\texttt{moment}}$  & Scribble Prop. & mIoU ($\uparrow$) & Scribble Ratio ($\uparrow$) \\  
        \midrule

        \xmark & \xmark & 0.391 & 0.697 \\ 
        \cmark & \xmark & 0.406 & 0.715 \\ 
        \xmark & \cmark & 0.396 & 0.697 \\ 
        \cmark & \cmark & \textbf{0.410} & \textbf{0.717} \\
        
        \bottomrule
    \end{tabular}}
    \caption{
        \textbf{Ablation study on our proposed components.} 
        With all components activated, our approach achieves the highest mIoU and Scribble Ratio score. This result indicates that each element plays a vital role in enhancing the quality of the final output. 
    }
    \label{tab:quantitative_pascal}
\end{table}

We conduct an ablation study on the PASCAL Scribble dataset to evaluate the effectiveness of our components: moment loss $\mathcal{L}_{\texttt{moment}}$ and scribble propagation.
\cref{tab:quantitative_pascal} shows the performance of different configurations in terms of mIoU and Scribble Ratio. As shown in \cref{tab:quantitative_pascal}, the increase of  $\mathcal{L}_{\texttt{moment}}$ improves both the mIoU and scribble ratio. Moreover, the proposed scribble propagation also contributes to further improvements in mIoU. Comprehensively, employing scribble propagation and $\mathcal{L}_{\texttt{moment}}$ achieves a 0.02 point improvement in the mIoU and 0.02 gain in the scribble ratio.

As demonstrated in \cref{fig:appendix_ablation_scribble_propagation}, omitting scribble propagation results in significant issues during generation, particularly when handling thin and sparse scribbles. For example, without scribble propagation, the thin scribble representing \textit{"sunglasses"} is ignored, and no sunglasses are generated. By contrast, when applying scribble propagation, our method iteratively extends the scribble during the denoising process, ensuring that smaller, detailed elements—such as the sunglasses—are accurately generated and aligned with the input prompt. This effect is particularly beneficial when handling thin scribbles, as they are more prone to being overlooked during generation.

\begin{figure}[ht]
    \centering
    \parbox{1.0\linewidth}{
        \centering
        \scriptsize{\textit{\textcolor{darkgray}{\textbf{Cute panda}} peacefully drifting on a \textcolor{brown}{bamboo raft} down a serene \textcolor{RoyalBlue}{\textbf{river}} in a lush \textcolor{ForestGreen}{\textbf{bamboo forest}}, detailed digital painting}}
    }
    \vspace{0.5em}
    \begin{subfigure}[t]{0.24\linewidth}
        \includegraphics[width=\textwidth]{assets_qualitative/cute_panda/scribbles.pdf}
    \end{subfigure}
    \begin{subfigure}[t]{0.24\linewidth}
        \includegraphics[width=\textwidth]{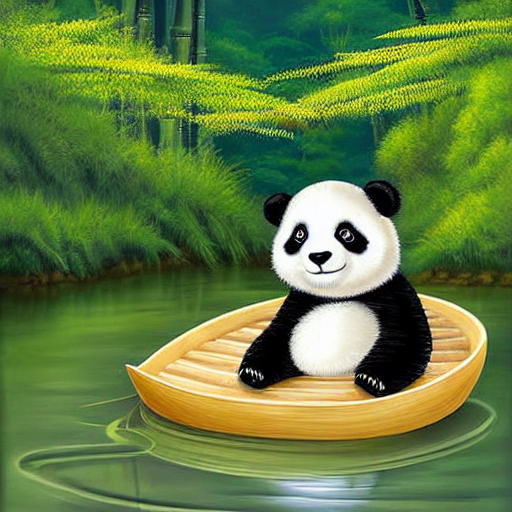}
        \caption{$\lambda_1, \lambda_2 = 0.2$}
    \end{subfigure}
    \hfill
    \begin{subfigure}[t]{0.24\linewidth}
        \includegraphics[width=\textwidth]{assets_qualitative/cute_panda/sgdiff.png}
        \caption{$\lambda_1, \lambda_2 = 0.6$}
    \end{subfigure}
    \begin{subfigure}[t]{0.24\linewidth}
        \includegraphics[width=\textwidth]{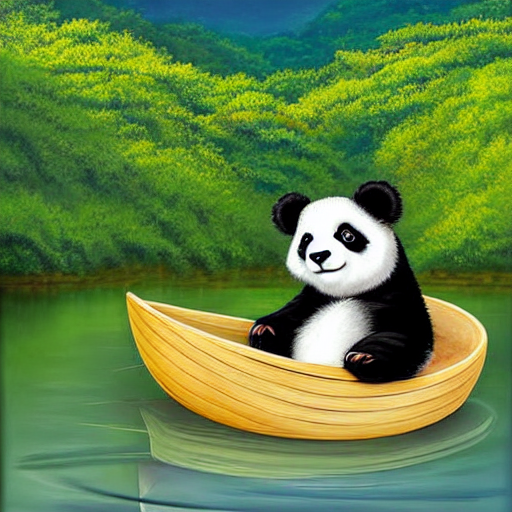}
        \caption{$\lambda_1, \lambda_2 = 1.0$}
    \end{subfigure}

    \caption{
        \textbf{Change in image as the scale $\lambda_1$ and $\lambda_2$ changes.} As the values of $\lambda_1$ and $\lambda_2$ increase, the generated image increasingly aligns with the scribble input. This is evident in the images from left to right, where the shape of the \emph{bamboo raft} progressively conforms to the thin scribble, and the orientation of the \emph{cute panda} shifts from facing forward to the left, as specified by the input scribble.
    }
    \label{fig:supple_moment_ablation}
\end{figure}

We also show the impact of the scales $\lambda_1$ and $\lambda_2$ while fixing other parameters in ~\cref{fig:supple_moment_ablation}. Both $\lambda_1$ and $\lambda_2$ are hyperparameters used to weigh the centroid and central moment losses. 
We observe that as the $\lambda_1$ and $\lambda_2$ scales increase, the image becomes more closely aligned with the thin scribble input. This is particularly noticeable in the \emph{bamboo raft}, whose shape adapts to better reflect the thin scribble structure. In addiotion, the orientation of the \emph{cute panda} moves from facing forward to the left by increasing $\lambda_1$ and $\lambda_2$

\begin{figure}[ht]
    \centering
    \parbox{1.0\linewidth}{
        \centering
        \footnotesize{\textit{a \textbf{pig} is next to a \textbf{cow}}}
    }
    \vspace{0.5em}
    \begin{subfigure}[t]{0.32\linewidth}
        \includegraphics[width=\textwidth]{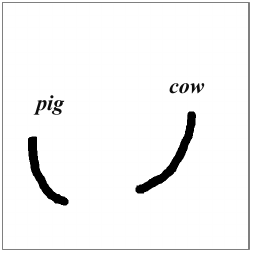}
    \end{subfigure}
    \begin{subfigure}[t]{0.32\linewidth}
        \includegraphics[width=\textwidth]{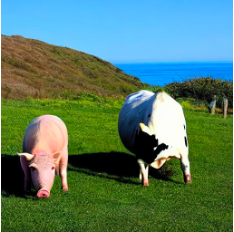}
    \end{subfigure}
    \begin{subfigure}[t]{0.32\linewidth}
        \includegraphics[width=\textwidth]{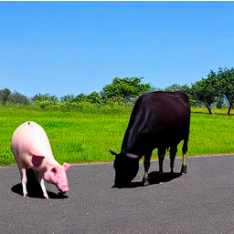}
    \end{subfigure}

    \parbox{1.0\linewidth}{
        \centering
        \footnotesize{\textit{a \textbf{cheetah} and an \textbf{elephant}}}
    }
    \vspace{0em}
    \begin{subfigure}[t]{0.32\linewidth}
        \includegraphics[width=\textwidth]{assets_qualitative/a_cheetah_and_an_elephant/scribbles.pdf}
    \end{subfigure}
    \begin{subfigure}[t]{0.32\linewidth}
        \includegraphics[width=\textwidth]{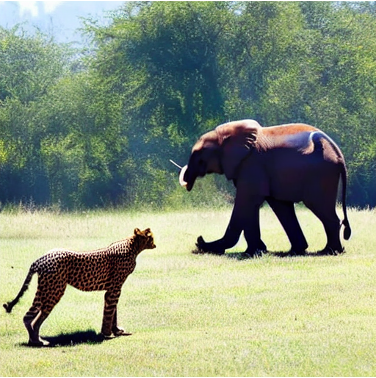}
    \end{subfigure}
    \begin{subfigure}[t]{0.32\linewidth}
        \includegraphics[width=\textwidth]{assets_qualitative/a_cheetah_and_an_elephant/sgdiff.png}
    \end{subfigure}
    
    \parbox{1.0\linewidth}{
        \centering
        \footnotesize{\textit{a \textbf{dog} and a \textbf{cat}.}}
    }
    \vspace{0em}
    \begin{subfigure}[t]{0.32\linewidth}
        \includegraphics[width=\textwidth]{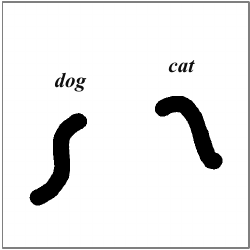}
    \end{subfigure}
    \begin{subfigure}[t]{0.32\linewidth}
        \includegraphics[width=\textwidth]{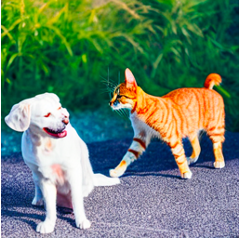}
    \end{subfigure}
    \begin{subfigure}[t]{0.32\linewidth}
        \includegraphics[width=\textwidth]{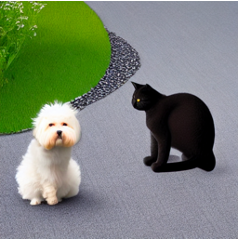}
    \end{subfigure}

    \parbox{1.0\linewidth}{
        \centering
        \footnotesize{\textit{a \textbf{person} surfing in the sea on a big \textbf{wave}.}}
    }
    \vspace{0em}
    \begin{subfigure}[t]{0.32\linewidth}
        \includegraphics[width=\textwidth]{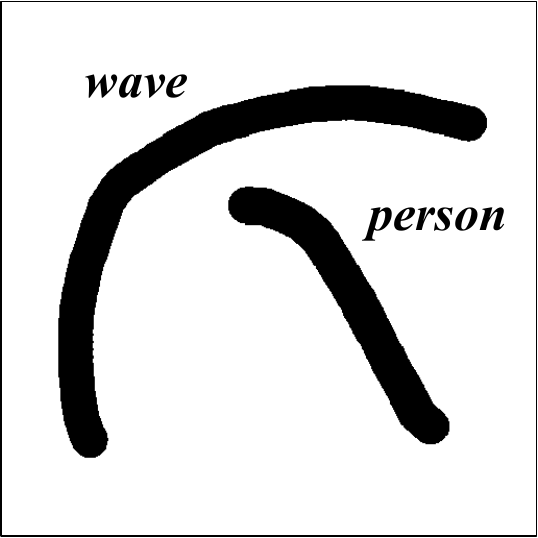}
        \caption{scribbles}
    \end{subfigure}
    \begin{subfigure}[t]{0.32\linewidth}
        \includegraphics[width=\textwidth]{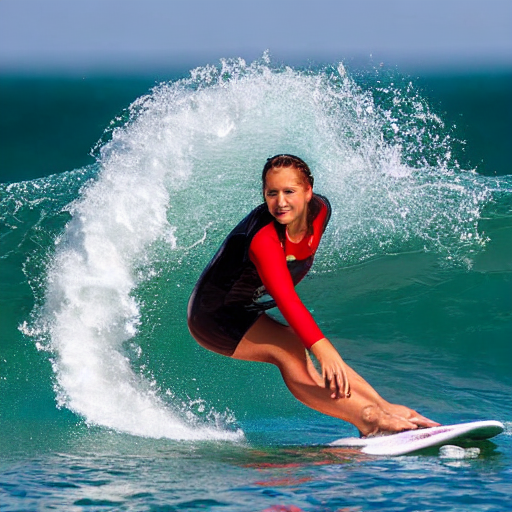}
        \caption{w/o $\mathcal{L}_{\texttt{moment}}$}
    \end{subfigure}
    \begin{subfigure}[t]{0.32\linewidth}
        \includegraphics[width=\textwidth]{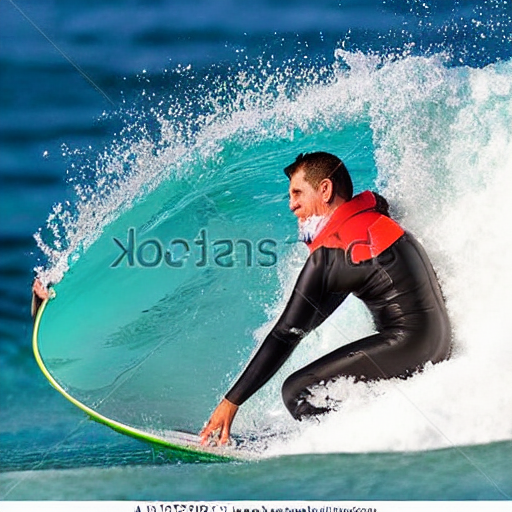}
        \caption{w/ $\mathcal{L}_{\texttt{moment}}$}
    \end{subfigure}

    \caption{
        \textbf{Moment Loss.} 
        We show a visual comparison of our approach both with and without moment loss. Notably, in the images labeled (c), where moment loss is applied, the subjects are oriented toward the target direction. 
        This observation clearly indicates that moment loss effectively contributes to the proper alignment of the object's orientation.
    }
    \label{fig:supple_quali_sgdiff_02}
\end{figure}

\section{User Study Details}
\label{sec:supple_user_study}

\begin{figure}[ht]
    \centering
    \includegraphics[width=0.95\linewidth]{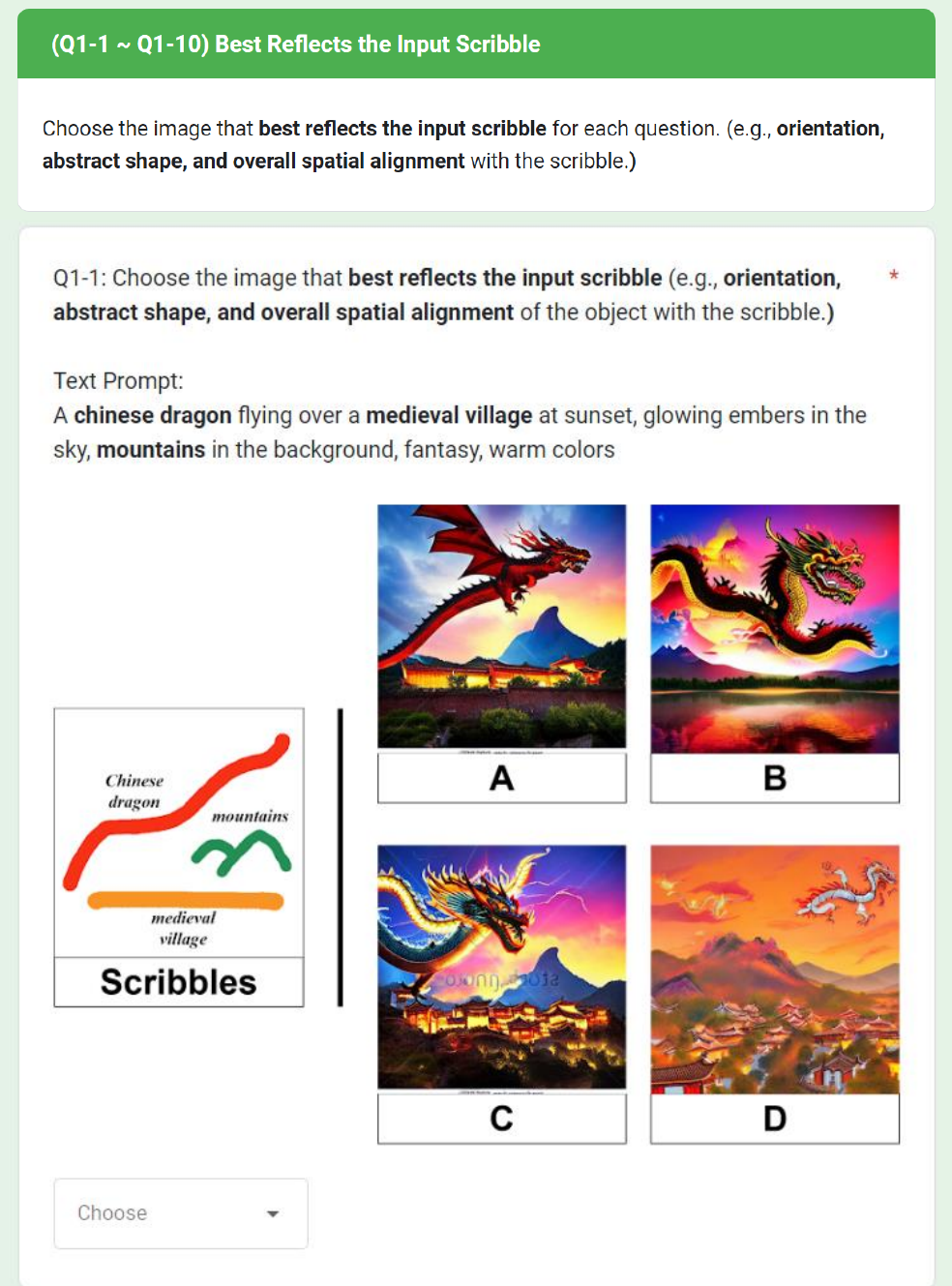}
    \caption{
        \textbf{Screenshot of our user study.}
        Participants were asked to compare images from four methods, including our approach. 
        }
    \label{fig:user_study_example}
\end{figure}

User study focused on evaluating image quality and alignment to determine the human-preferred approach. Human evaluators were presented with a prompt and an input scribble and were asked to select the best result from four different models: BoxDiff, DenseDiff, GLIGEN, and our proposed method. The images were randomly ordered and labeled A through D. Each participant was tasked with completing a total of 30 evaluation questions, as there were three distinct questions associated with each set of 10 samples. An example of the survey is shown in \cref{fig:user_study_example}. 

Below we include the full questions used for our user study. 
\begin{itemize}
    \item Choose the image that \textbf{best reflects the input scribble} (\textit{\textbf{e.g.}}, orientation, abstract shape, and overall spatial alignment of the object with the scribble.)
    \item Choose the image that best represents the content of the text prompt, considering all key elements described in the text. (\textit{\textbf{e.g.}}, \textbf{no key elements in bold are missing} and the generated object is coherent and complete.)
    \item Choose the image that best balances \textbf{reflecting the input scribble} and \textbf{accurately representing the content of the text prompt}. (The best image considering both Set 1 and Set 2 criteria.)
\end{itemize}

The first question aims to assess the generated image's alignment with the input scribble. This measure is crucial for determining how well the model adheres to user-provided visual guides, such as scribbles, which are necessary for customization or specific design constraints. This question evaluates aspects such as orientation, abstract shape, and spatial arrangement.

The second question evaluates how effectively the generated images capture the essence of the text prompt, ensuring that all critical elements highlighted in the prompt are correctly depicted in the generated images. This question is asked to measure the model's capacity not to neglect any necessary key objects, leading to complete representations.

The last question seeks to determine the optimal balanced assessment, which combines the criteria asked in the two previous questions. This is particularly relevant to scenarios where both textual and visual cues must be considered to generate contextually appropriate and visually coherent outputs.

\section{Limitation \& Discussion}
\label{sec:limitation_and_discussion}

This study focuses on improving the incorporation of scribbles as a form of guidance in text-to-image (T2I) generation models, rather than enhancing the overall T2I performance.
Future research can explore methods to boost the performance of T2I models directly while maintaining improvements in scribble-based guidance.

In addition to the Bezier Scribbles~\cite{chen2023scribbleseg} used in this study, future work could investigate developing models that are robust across various types of sketches, such as Axial Scribbles and Boundary Scribbles.
These models should effectively handle different forms of sketch input to improve flexibility in practical applications.



\end{document}